\documentclass[letterpaper,journal]{IEEEtran}
\usepackage{amsmath,amsfonts}
\usepackage{array}
\usepackage[caption=false,font=small]{subfig}
\usepackage{textcomp}
\usepackage{stfloats}
\usepackage{url}
\usepackage{verbatim}
\usepackage{graphicx}
\usepackage{cite}
\usepackage{xspace}

\usepackage{times}

\usepackage{epsfig}
\usepackage{amsmath}
\usepackage{amssymb}
\usepackage{bm}
\usepackage{booktabs}
\usepackage{multirow}
\usepackage{rotating}
\usepackage{xcolor}
\usepackage{colortbl}
\usepackage{capt-of}
\usepackage{setspace}
\usepackage{soul}
\usepackage{mathtools} 
\usepackage{eso-pic} 
\usepackage{pifont} 
\usepackage[ruled,vlined,,linesnumbered]{algorithm2e}
\usepackage{dsfont}
\usepackage{balance}
\usepackage{float}
\usepackage{changepage}

\definecolor{revisionblue}{RGB}{0,0,0}
\newcommand{\rev}[1]{{\color{revisionblue}#1}}

\newif\ifisArXivMode
\isArXivModetrue    

\renewcommand{\figurename}{Fig.} 
\usepackage[pagebackref=true,breaklinks=true,colorlinks,bookmarks=false]{hyperref}
\usepackage{cleveref} 
\Crefname{section}{Sec.}{Secs.} 
\Crefname{figure}{Fig.}{Figs.} 
\Crefname{table}{Table}{Tables} 
\crefname{suppsection}{Sec.}{Secs.}
\Crefname{suppsection}{Sec.}{Secs.}
\crefname{equation}{}{}
\Crefname{equation}{}{}

\newcommand{\bdmath}{\begin{dmath}}
\newcommand{\edmath}{\end{dmath}}
\newcommand{\beq}{\begin{equation}}
\newcommand{\eeq}{\end{equation}}
\newcommand{\bdm}{\begin{displaymath}}
\newcommand{\edm}{\end{displaymath}}
\newcommand{\bea}{\begin{eqnarray}}
\newcommand{\eea}{\end{eqnarray}}
\newcommand{\beal}{\beq \begin{array}{ll}}
\newcommand{\eeal}{\end{array} \eeq}
\newcommand{\beas}{\begin{eqnarray*}}
\newcommand{\eeas}{\end{eqnarray*}}
\newcommand{\ba}{\begin{array}}
\newcommand{\ea}{\end{array}}
\newcommand{\bit}{\begin{itemize}}
\newcommand{\eit}{\end{itemize}}
\newcommand{\ben}{\begin{enumerate}}
\newcommand{\een}{\end{enumerate}}

\newcommand{\calE}{{\cal E}}

\newcommand{\calG}{{\cal G}}

\newcommand{\calI}{{\cal I}}

\newcommand{\calS}{{\cal S}}

\newcommand{\calV}{{\cal V}}

\newcommand{\setal}{~\emph{et~al.}\xspace}
\providecommand{\eg}{\emph{e.g.,}\xspace}
\providecommand{\ie}{\emph{i.e.,}\xspace}

\def\etalcite#1{\setal~\cite{#1}}

\newcommand{\rom}[1]{\uppercase\expandafter{\romannumeral #1\relax}}

\newcommand{\M}[1]{{\bm #1}} 
\renewcommand{\boldsymbol}[1]{{\bm #1}}

\newcommand{\hide}[1]{}

\newcommand{\hiddenText}{{\color{gray} hidden text.}}
\newcommand{\hideWithText}[1]{\hiddenText}

\DeclareMathOperator*{\argmin}{arg\,min}

\newcommand{\twonorm}[1]{\|#1\|_{2}}

\newcommand{\SOthree}{\ensuremath{\mathrm{SO}(3)}\xspace}

\newcommand{\MC}{\M{C}}

\newcommand{\MR}{\M{R}}

\newcommand{\MI}{\M{I}}
\newcommand{\MV}{\M{V}}

\newcommand{\vc}{\boldsymbol{c}}

\newcommand{\vn}{\boldsymbol{n}}

\newcommand{\vp}{\boldsymbol{p}}
\newcommand{\vq}{\boldsymbol{q}}

\newcommand{\vv}{\boldsymbol{v}}
\newcommand{\vt}{\boldsymbol{t}}

\newcommand{\blue}[1]{{\color{blue}#1}}

\newcommand{\linkToPdf}[1]{\href{#1}{\blue{(pdf)}}}
\newcommand{\linkToPpt}[1]{\href{#1}{\blue{(ppt)}}}
\newcommand{\linkToCode}[1]{\href{#1}{\blue{(code)}}}
\newcommand{\linkToWeb}[1]{\href{#1}{\blue{(web)}}}
\newcommand{\linkToVideo}[1]{\href{#1}{\blue{(video)}}}
\newcommand{\linkToMedia}[1]{\href{#1}{\blue{(media)}}}
\newcommand{\award}[1]{\xspace} 

\newcommand{\vz}{\boldsymbol{z}}

\newcommand{\oursname}{{BUFFER-X}\xspace}
\newcommand{\fastversion}{{BUFFER-X-Lite}\xspace}
\newcommand{\kmsolver}{{KISS-Matcher solver}\xspace}

\newcommand{\ModelNet}{\texttt{ModelNet40}\xspace}
\newcommand{\ThreeDMatch}{\texttt{3DMatch}\xspace}
\newcommand{\ThreeDLoMatch}{\texttt{3DLoMatch}\xspace}
\newcommand{\ScanNetppi}{\texttt{ScanNet++i}\xspace}
\newcommand{\ScanNetppF}{\texttt{ScanNet++F}\xspace}
\newcommand{\TIERS}{\texttt{TIERS}\xspace}
\newcommand{\KITTI}{\texttt{KITTI}\xspace}
\newcommand{\ETH}{\texttt{ETH}\xspace}
\newcommand{\MIT}{\texttt{MIT}\xspace}
\newcommand{\KAIST}{\texttt{KAIST}\xspace}
\newcommand{\Oxford}{\texttt{Oxford}\xspace}
\newcommand{\WOD}{\texttt{WOD}\xspace}

\newcommand{\omitted}[1]{}

\newcommand{\bmat}{\left[ \begin{array}}
\newcommand{\emat}{\end{array}\right]}

\newcommand{\subMeas}[1]{\calS} %

\begin{document}

\title{Toward Zero-Shot Point Cloud Registration Across Diverse Scales, Scenes, and Sensor Setups}

\author{Hyungtae Lim$^{*}$~\IEEEmembership{Member,~IEEE}, Minkyun Seo$^{*}$~\IEEEmembership{Graduate Student Member,~IEEE}, \\ Luca Carlone~\IEEEmembership{Senior Member,~IEEE}, and Jaesik Park$^{\dagger}$,~\IEEEmembership{Member,~IEEE}

\thanks{$^{\dagger}$Corresponding author.}
\thanks{$^{*}$These authors contributed equally to this work.}
\thanks{Hyungtae Lim and Luca Carlone are with the Laboratory for Information and Decision Systems (LIDS), Massachusetts Institute of Technology, Cambridge, MA, USA (e-mail: \texttt{\{shapelim, lcarlone\}@mit.edu})}
\thanks{Minkyun Seo and Jaesik Park are with Department of Computer Science and Engineering, Seoul National University, Seoul, Republic of Korea, (e-mail: \texttt{\{funboy0804, jaesik.park\}@snu.ac.kr})}
\ifisArXivMode
    \thanks{\copyright~2026 IEEE. Personal use of this material is permitted. Permission from IEEE must be obtained for all other uses, in any current or future media, including reprinting/republishing this material for advertising or promotional purposes, creating new collective works, for resale or redistribution to servers or lists, or reuse of any copyrighted component of this work in other works.}
    \thanks{Digital Object Identifier: \href{https://doi.org/10.1109/TPAMI.2026.3722524}{\textcolor{black}{10.1109/TPAMI.2026.3722524}}}
    \thanks{IEEE Transactions on Pattern Analysis and Machine Intelligence, vol.~48, no.~11, pp.~15234--15251, Nov.~2026.}
\else
\fi
}

\ifisArXivMode
    \markboth{IEEE Transactions on Pattern Analysis and Machine Intelligence,~Vol.~48, No.~11, Nov.~2026}%
    {Lim \MakeLowercase{\textit{et al.}}: Toward Zero-Shot Point Cloud Registration Across Diverse Scales, Scenes, and Sensor Setups}
\else
    \markboth{Journal of \LaTeX\ Class Files,~Vol.~X, No.~X, Aug~2026}%
    {Lim \MakeLowercase{\textit{et al.}}: Toward Zero-Shot Point Cloud Registration Across Diverse Scales, Scenes, and Sensor Setups}
\fi

\makeatletter
  \let\@oldmaketitle\@maketitle
  \renewcommand{\@maketitle}{\@oldmaketitle
  \bigskip
  \centering
    \setcounter{figure}{0}
    \includegraphics[width=1.0\textwidth]{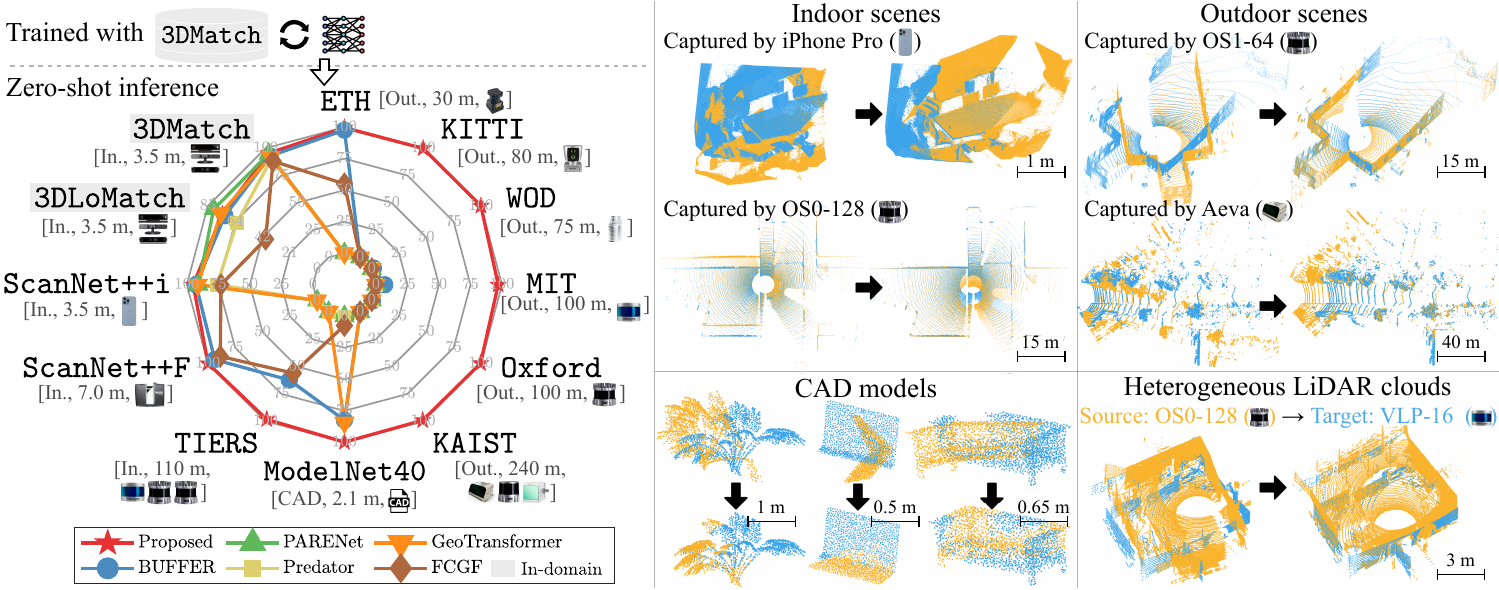}
    \captionof{figure}{Success rate (unit: \%) of \textit{zero-shot point cloud registration} with state-of-the-art approaches on 12 datasets~\cite{Wu15cvpr-ModelNet,Zeng17cvpr-3dmatch,Huang21cvpr-PREDATORRegistration,Geiger13ijrr-KITTI,Yeshwanth23iccv-Scannet++,Qingqing22iros-TIERS,Ramezani20iros-NewerCollege,Sun20cvpr-WaymoDataset,Tian23iros-KimeraMultiExperiments,Pomerleau12ijrr-ETH,Jung23ijrr-HeLiPR}.
    \rev{Using a model trained once on a single source dataset~({\ThreeDMatch}~\cite{Zeng17cvpr-3dmatch} for the results shown here), without target-domain retraining, fine-tuning, or dataset-specific parameter tuning, \textit{BUFFER-X} generalizes across diverse scales, scenes, and sensor setups.}
    In particular, our proposed approach can operate on CAD models~(from left to right, the point clouds represent a plant, a laptop, and a sofa, respectively) as well as on heterogeneous LiDAR point clouds.
    }            
    \vspace{-10mm}
    \label{fig:fig1}
  }
\makeatother

\maketitle

\begin{abstract}
Some deep learning-based point cloud registration methods struggle with zero-shot generalization, often requiring \rev{target-domain retraining, fine-tuning, or dataset-specific metric parameter tuning} for new environments.
We identify three critical limitations: (a)~fixed user-defined parameters~(\eg voxel size, search radius) that fail to generalize across varying scales, (b)~learned keypoint detectors exhibit poor cross-domain transferability, and (c)~absolute coordinates amplify scale mismatches between datasets.
To address these three issues, we present \textit{BUFFER-X}, \rev{a registration framework whose descriptor model is trained once and then applied without parameter tuning or target-domain retraining}, which achieves zero-shot generalization through: (a)~geometric bootstrapping for automatic hyperparameter estimation, (b)~distribution-aware farthest point sampling to replace learned detectors, and (c)~patch-level coordinate normalization to ensure scale consistency.
Our approach employs hierarchical multi-scale matching to extract correspondences across local, middle, and global receptive fields, enabling robust registration in diverse environments.
\rev{For efficiency-critical applications, we present \textit{\fastversion} from a holistic system perspective, integrating the BUFFER-X front-end and adaptive early exit with the fast and robust pose solver to reduce total computation time by 43\% relative to \oursname while preserving comparable accuracy.}
We evaluate on a comprehensive benchmark comprising 12 datasets spanning object-scale, indoor, and outdoor scenes, including cross-sensor registration between heterogeneous LiDAR configurations.
Results demonstrate that our approach \rev{generalizes effectively across diverse scales, scenes, and sensor setups}.
Code: \href{https://github.com/MIT-SPARK/BUFFER-X}{\texttt{https://github.com/MIT-SPARK/BUFFER-X}}.
\end{abstract}


\section{Introduction}
\IEEEPARstart{D}{eep} learning methods for point cloud registration have advanced significantly, with improvements spanning feature learning~\cite{Ao20ietcv-SGHs,Ao20pr-Repeatable,Ao23CVPR-BUFFER,Ao21cvpr-Spinnet,Chen24cpvr-DCATr,Mu24cvpr-ColorPCR, Huang21cvpr-PREDATORRegistration,Chen24icra-Tree-based-Transformer}, correspondence matching~\cite{Huang21cvpr-PREDATORRegistration,Zhang24cvpr-FastMAC, Zhang23cvpr-MaxCliques,Fathian24ral-Clipperplus,Liu24tgrs-DeepSemanticMatching,Liu24tpami-NCMNet,Yuan24cvpr-InlierConfidence,Liu24cvpr-Extend}, and pose solver design~\cite{Shi24ral-RANSAC,Huang24cvpr-Scalable,Yang24tpami-MAC,Zhou16eccv-FastGlobalRegistration}.
These methods demonstrate strong in-domain performance,
accurately aligning partially overlapping point clouds when evaluated on their training distributions~\cite{Yang20tro-teaser, Lim24ijrr-Quatropp, Lim22icra-Quatro, Yang16pami-goicp, Bernreiter21ral-PHASER, Yin23icra-Segregator}.

However, the ability to generalize across diverse real-world environments remains a key challenge~\cite{Ao23CVPR-BUFFER,Ao21cvpr-Spinnet,Poiesi22pami-GeDi,Dosovitskiy20iclr-YOTO,Chen21nips-OnlyTrainOnce}, with recent work focusing on cross-domain transferability.
Despite progress in generalization~\cite{wang2024-ZeroReg,zheng2025-RARE}, practical deployment of some methods still depends on dataset-specific hyperparameters,
particularly voxel size for downsampling and search radius for feature extraction, which must be manually tuned for each new environment.
The need for such intervention limits practical applicability, motivating \rev{registration methods that can be trained once in a source domain and transferred without target-domain retraining, fine-tuning, or dataset-specific parameter adjustment}.

Additionally, evaluation protocols have historically focused on narrow scenarios: outdoor LiDAR scans~\cite{Yew18eccv-3dfeatnet, Geiger13ijrr-KITTI} or indoor RGB-D data~\cite{Zeng17cvpr-3dmatch}.
Cross-domain evaluations involving object-level CAD models~\cite{Wu15cvpr-ModelNet}, indoor LiDAR~\cite{Qingqing22iros-TIERS}, diverse outdoor scanning patterns~\cite{Qingqing22iros-TIERS,Jung23ijrr-HeLiPR,Lim24iros-HeLiMOS}, or heterogeneous sensor configurations where source and target are captured by different devices, remain underexplored.
Therefore, a comprehensive benchmark reflecting real-world sensor diversity is needed to rigorously assess cross-domain generalization.

Based on these insights, in our previous work~\cite{Seo25ICCV-BUFFERX}, we presented \oursname, which employs: (i)~geometric bootstrapping to automatically estimate the voxel size and search radii for each scene, (ii)~distribution-aware sampling to replace learned keypoint detectors, and (iii)~hierarchical multi-scale patch matching to extract robust correspondences across varying receptive fields.
\rev{That work primarily established the feasibility of zero-shot point cloud registration, demonstrating that a model trained on a single source dataset can be transferred across indoor and outdoor scenes without target-domain retraining or parameter tuning.}
\rev{It also helped draw broader attention to zero-shot point cloud registration, a direction subsequently explored by RAP~\cite{Pan26eccv-RAP}.}
\rev{In this paper, we retain the BUFFER-X backbone for a self-contained description while shifting the emphasis from \emph{whether} such transfer is possible to \emph{why} it works and \emph{under which conditions} it remains reliable.}

In summary, this work extends our previous study in three directions.
First, we expand the benchmark to include object-scale scenes and heterogeneous sensor configurations, enabling a broader evaluation of cross-domain generalization (see \Cref{fig:fig1} and \Cref{sec:benchmark,sec:sota_comparison}).
Second, \rev{we provide quantitative analyses of descriptor distributions, complementary correspondence patterns across scales, and parameter sensitivity to explain the domain-generalizable behavior of BUFFER-X and assess its robustness without per-domain parameter tuning} (see \Cref{sec:indepth,sec:param_sensitivity}).
Third, \rev{we present \fastversion from a holistic system perspective by integrating BUFFER-X feature extraction and pair-adaptive early exit with the fast and robust pose solver developed in our previous work, KISS-Matcher~\cite{Lim25icra-KISSMatcher}.}
\rev{While KISS-Matcher focused solely on correspondence pruning and pose estimation without addressing descriptor generation or feature matching, \fastversion integrates its solver with the BUFFER-X front-end to cover the complete registration pipeline, spanning descriptor generation, feature matching, correspondence pruning, and pose estimation. Consequently, \fastversion reduces total computation time while maintaining comparable registration accuracy} (see \Cref{sec:fastbufferx,sec:fast_buffer_x_exp}).

\section{Related Work}
\label{sec:related}

3D point cloud registration, which estimates the relative pose between two partially overlapping point clouds, is a fundamental problem in the fields of robotics and computer vision~\cite{Aoki24icra-3DBBS, Yin24ijcv-LiDARLocSurvey, Chen22ar-Overlapnet, Cattaneo22tro-LCDNet, Lee23cvprw-LearningReg}.
Overall, point cloud registration methods are classified into two categories based on whether their performance relies on the availability of an initial guess for registration: a)~\emph{local} registration~\cite{Besl92pami, Segal09rss-GeneralizedICP,Pomerleau13auro-ICPcomparison,Koide21icra-VGICP,Pomerleau12ijrr-ETH,Vizzo23ral-KISSICP} and b)~\emph{global} registration~\cite{Fischler81,Dong17isprsremotesensing-GHICP,Yang16pami-goicp,Zhou16eccv-FastGlobalRegistration,Bernreiter21ral-PHASER,Yang19rss-teaser,Yang20tro-teaser,Lim22icra-Quatro,Lim24ijrr-Quatropp}.
Global registration methods can be further classified into two types: a)~\emph{correspondence-free}~\cite{Rouhani11iccv-CorrespondenceFreeReg,Brown19pr-AFamiliyofBnB,Bernreiter21ral-PHASER, Papazov12ijrr-Rigid3DGeometryMatching,Chum03jprs-LocallyOptimizedRANSAC,Choi97jcv-RANSAC,Schnabel07cgf-EfficientRANSAC,Olsson09pami-bnbRegistration, Hartley09ijcv-globalRotationRegistration, Pan19robotbiomim-MultiViewBnB,Zhao24cvpr-CorrespondenceFree} and b)~\emph{correspondence-based}~\cite{Fischler81,Yang16pami-goicp,Zhou16eccv-FastGlobalRegistration,Dong17isprsremotesensing-GHICP,Lei17tip-FastDescriptors,Yuan24cvpr-InlierConfidence,Liu24cvpr-Extend,Yu24cvpr-InstanceAware, mei2023-unsuperviseddeepprobabilisticapproach, zheng2025-RARE} approaches.
In this study, we focus on the latter and particularly place more emphasis on deep learning-based registration methods.

Since Qi\etalcite{Qi17cvpr-pointnet} demonstrated that learning-based techniques in 2D images can also be applied to 3D point clouds, a wide range of learning-based point cloud registration approaches have been proposed.  
Building on these advances, novel network architectures with increased capacity have continuously emerged, ranging from MinkUNet~\cite{Choi19iccv-FCGF,Choy20cvpr-deepGlobalRegistration,choy19cvpr-4DSTConv}, cylindrical convolutional network~\cite{Ao21cvpr-Spinnet,Ao23CVPR-BUFFER,Zhu21cvpr-Cylindrical3D}, KPConv~\cite{Huang21cvpr-PREDATORRegistration,Thomas19iccv-kpconv, Bai20cvpr-D3Feat} to Transformers~\cite{Qin23tpami-GeoTransformer,Wu22nips-PointTransformerV2,Wu24cvpr-PointTransformerV3,Chen24cpvr-DCATr, yu2023-RoITr}.

While these advancements have led to improved registration performance, some of these methods often exhibit limited generalization capability,
leading to performance degradation when applied to point clouds captured by different sensor configurations or in unseen environments. 
To tackle the generalization problem, Ao\etalcite{Ao21cvpr-Spinnet, ao22tpami-YOTO} introduce SpinNet, a patch-based method that normalizes the range of local point coordinates within a fixed-radius neighborhood to $[-1, 1]$.
This demonstrates that patch-wise scale normalization is key to achieving a data-agnostic registration pipeline.

Further, Ao\etalcite{Ao23CVPR-BUFFER} proposed BUFFER to enhance efficiency by combining point-wise feature extraction with patch-wise descriptor generation.
However, we found that such learning-based keypoint detectors can hinder robust generalization, as their failure on out-of-domain distributions may trigger cascading failures in subsequent steps; see \Cref{sec:probpt_detector}.
In addition, despite the high generalizability of BUFFER,
we observed that during cross-domain testing, users had to manually specify the optimal voxel size for the test data, which hinders fully zero-shot inference.
On the other hand, Wang\etalcite{wang2024-ZeroReg} and Zheng\etalcite{zheng2025-RARE} proposed zero-shot registration approaches that rely heavily on depth images and RGB images, respectively.
However, this dependency limits their applicability to diverse point cloud registration scenarios, ranging from object-level CAD models to outdoor environments.

Beyond generalizing across diverse scenes captured by a single sensor type,
recent works~\cite{huang2021-3DCSR,Qingqing22iros-TIERS,Jung23ijrr-HeLiPR} have broadened the notion of generalizability to heterogeneous sensor configurations, where source and target clouds are captured by different sensor modalities and scanning patterns.
Accordingly, evaluating generalizability in point cloud registration requires considering scenarios where the source and target point clouds are obtained from different heterogeneous sensors~(\eg sparse omnidirectional to dense omnidirectional, or omnidirectional to face-forward solid-state LiDAR sensors).

Under these circumstances, we revisit the generalization problem in point cloud registration and explore how to achieve zero-shot registration while preserving the key benefits of BUFFER's scale normalization strategy.
In addition, we remove certain modules that hinder robustness and introduce an adaptive mechanism that determines the voxel size and search radii depending on the given point cloud pair. 
To the best of our knowledge, this is the first approach to evaluate the zero-shot generalization across diverse scenes covering various environments, geographic regions, scales, sensor types, and acquisition setups.

\section{Preliminaries}\label{sec:our_preliminaries}

\subsection{Problem statement}\label{sec:problem}
\newcommand{\corr}{\mathcal{A}}
\newcommand{\estoutliers}{\hat{\mathcal{O}}}
\newcommand{\srcpt}{\srcpoint_\srcidx}
\newcommand{\tgtpt}{\tgtpoint_\tgtidx}
\newcommand{\voxel}{v}
\newcommand{\voxelfunc}[1]{f_\voxel({#1})}
\newcommand\srcpoint{\vp}
\newcommand\tgtpoint{\vq}
\newcommand\srcidx{i}
\newcommand\tgtidx{j}
\newcommand{\kth}{k}
\newcommand{\lth}{l}
\newcommand\srccloud{\mathcal{P}}
\newcommand\tgtcloud{\mathcal{Q}}

The goal of point cloud registration is to estimate the relative 3D rotation matrix $\MR \in \SOthree$ and translation vector $\vt \in \mathbb{R}^{3}$ between two unordered 3D point clouds $\mathcal{P}$ and $\mathcal{Q}$.
To this end, most approaches~\cite{Huang21cvpr-PREDATORRegistration, Lim25icra-KISSMatcher} follow three steps:
a)~apply voxel sampling $\voxelfunc{\cdot}$ to the point cloud with voxel size $\voxel$ as preprocessing,
b)~establish associations (or \textit{correspondences})~$\corr$,
and c)~estimate $\MR$ and $\vt$. 

Formally, by denoting corresponding points for a correspondence $(i, j)$ in $\mathcal{A}$ as $\vp_i \in \voxelfunc{\mathcal{P}}$ and  $\vq_j \in \voxelfunc{\mathcal{Q}}$, respectively,
the objective function used for pose estimation can be defined as:
\begin{equation}
  \hat{\MR}, \hat{\vt} =\argmin_{\MR \in \mathrm{SO}(3), \vt \in \mathbb{R}^{3}}  \sum_{(i,j) \in  \corr} \rho\Big( \twonorm{\tgtpoint_\tgtidx -\MR \srcpoint_\srcidx -\vt} \Big),
 \label{eq:final_goal}
\end{equation}
\noindent where $\rho(\cdot)$ represents a robust kernel function that mitigates the effect of spurious correspondences in $\corr$.

\begin{figure}[t!]
    \centering
    \subfloat[]{%
        \includegraphics[width=0.23\textwidth]{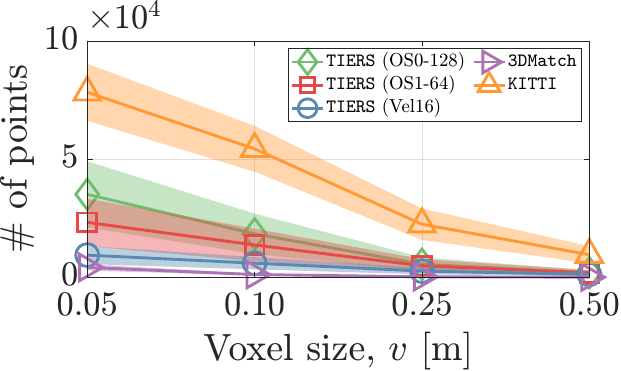}%
        \label{fig:key_elements_a}
    }
    \hfill
    \subfloat[]{%
        \includegraphics[width=0.215\textwidth]{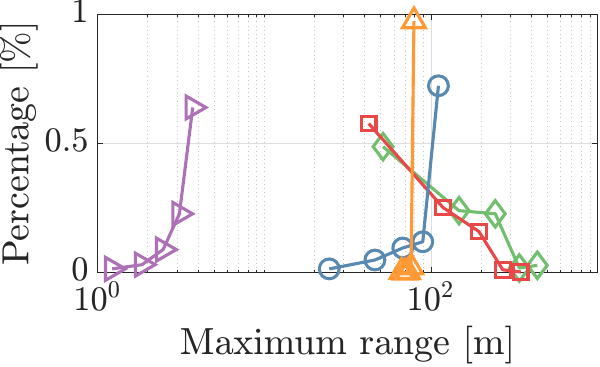}%
        \label{fig:key_elements_b}
    }
    \caption{
    (a) Variation in the number of points after voxelization with different voxel sizes $v$ across datasets.
    Even in indoor scenes, point counts vary significantly depending on the sensor type~(\ie {\TIERS}~\cite{Qingqing22iros-TIERS} vs. {\ThreeDMatch}~\cite{Zeng17cvpr-3dmatch}).
    Notably, {\TIERS} and {\KITTI}~\cite{Geiger13ijrr-KITTI}, both using omnidirectional LiDARs, yield different point densities due to indoor vs. outdoor environments.
    (b) Empirical distribution of the datasets’ maximum range.
    }
    \label{fig:key_elements}
    \vspace{-2mm}
\end{figure}

\subsection{Key observations}\label{sec:factors}

If $\corr$ in \Cref{eq:final_goal} is accurate, solving \cref{eq:final_goal} is a well-studied problem.
However, we have observed that there exist three factors that cause learning-based registration to struggle in estimating $\corr$ when given out-of-domain data.

\subsubsection{Voxel size and search radius}\label{sec:prob_user_defined}

First, the dependencies of optimal search radius $r$ for local descriptors and voxel size $\voxel$ for each dataset are problematic.
The optimal parameters vary significantly across datasets due to differences in scale and point density~(\eg small indoor scenes vs.\ large outdoor spaces~\cite{Ao23CVPR-BUFFER}).
Consequently, improper $r$ or $v$ can severely degrade registration performance by failing to account for specific scale and density characteristics of a given environment or sensor; see \Cref{sec:analyses}.
For instance, in \Cref{fig:key_elements}(a), as $\voxel$ controls the maximum number of points that can be fed into the network, 
a too-small $\voxel$ can trigger out-of-memory errors when outdoor data processed with parameters optimized for indoor environments are taken as input to the network.

In particular, most methods heavily depend on manual tuning, which hinders generalization.
Therefore, we employ \textit{geometric bootstrapping} to adaptively determine $v$ and $r$ at test time based on the scale and point density of the given input clouds; see \Cref{sec:geometric}. %

\subsubsection{Input scale normalization}\label{sec:prob_scale_norm}

Next, directly feeding raw $x$, $y$, and $z$ values into the network leads to strong in-domain dependency~\cite{Huang21cvpr-PREDATORRegistration, Qin23tpami-GeoTransformer}.
That is, when a model fits the training distribution, large-scale discrepancies between training and unseen data can cause catastrophic failure~(see \Cref{fig:key_elements}(b) for an example of maximum range discrepancy).
For this reason, we conclude that normalizing input points within local neighborhoods~(or~\textit{patches)} is necessary to achieve generalizability, ensuring that their coordinates lie within a bounded range~(\eg $[-1, 1]$)~\cite{Ao21cvpr-Spinnet,Ao23CVPR-BUFFER}.

Based on these insights, we adopt patch-based descriptor generation as our pipeline for descriptor matching; see \Cref{sec:tri-scale-patch}.

\subsubsection{Keypoint detection}\label{sec:probpt_detector}

Following \Cref{sec:prob_scale_norm}, we observed that point-wise feature extractor modules in existing methods~\cite{Huang21cvpr-PREDATORRegistration, Qin23tpami-GeoTransformer, Yao24iccv-PARENet} are empirically brittle to out-of-domain data.
Because failed keypoint detection leads to the selection of unreliable and non-repeatable points as keypoints, it results in low-quality descriptors and ultimately degrades the quality of~$\mathcal{A}$~\cite{Harris88avc-HarrisCorner}.

An interesting observation is that replacing the learning-based detector with farthest point sampling~(FPS) preserves registration performance~\cite{Seo25ICCV-BUFFERX}.
For this reason, we adopt FPS over a learning-based module.
Specifically, we apply it separately at local, middle, and global scales to account for multi-scale variations.

\section{BUFFER-X}
\newcommand{\spinnetout}{\mathcal{S}}
\rev{We begin by describing the multi-scale zero-shot registration pipeline of \oursname, originally presented in our previous work~\cite{Seo25ICCV-BUFFERX}; see \Cref{fig:pipeline}.}
First, the appropriate voxel size and radii for each cloud pair are predicted by geometric bootstrapping~(\Cref{sec:geometric}), considering the overall distribution of cloud points and the density of neighboring points, respectively.
Then, we extract Mini-SpinNet-based features~\cite{Ao23CVPR-BUFFER} for the sampled points via FPS at multiple scales~(\Cref{sec:tri-scale-patch}).
Finally, at the intra- and cross-scale levels, refined correspondences are estimated based on consensus maximization~\cite{Sun22ral-TriVoC,Shi24ral-RANSAC,Zhang24tpami-AcceleratingGloballyCM}~(\Cref{sec:hierarchical}) and serve as input for the final relative pose estimation using a solver.

\rev{Note that the learned descriptor is trained once on a source dataset and remains fixed thereafter, requiring neither target-domain retraining nor fine-tuning.
Although BUFFER-X includes several user-defined parameters, it operates out of the box on unseen datasets without modifying their values.}

\begin{figure*}[t!]
    \centering
    \includegraphics[width=1.0\textwidth]{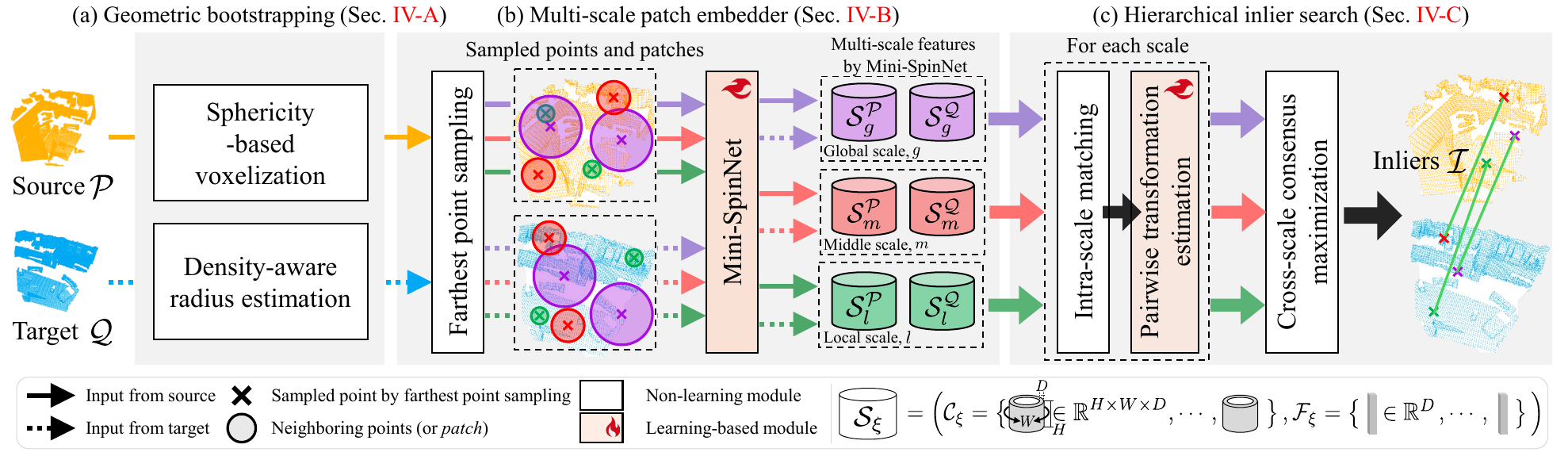}
    \caption{Overview of our \textit{\oursname}, which mainly consists of three steps.
    (a) Geometric bootstrapping~(\Cref{sec:geometric}) to determine the appropriate voxel size and radii for the given source~$\mathcal{P}$ and target~$\mathcal{Q}$ clouds. 
    (b)~Multi-scale patch embedder~(\Cref{sec:tri-scale-patch}) to generate patch-wise descriptor~$\spinnetout_\xi$ for multiple scales~$\xi \in \{l, m, g\}$, where $l$, $m$, and $g$ represent local, middle, and global scales, respectively.
    Specifically, Mini-SpinNet~\cite{Ao23CVPR-BUFFER} outputs cylindrical feature maps~$\mathcal{C}_\xi$ and vector feature sets~$\mathcal{F}_\xi$.
    (c)~Hierarchical inlier search~(\Cref{sec:hierarchical}), which first performs nearest neighbor-based intra-scale matching using~$\mathcal{F}^\mathcal{P}_\xi$ and~$\mathcal{F}^\mathcal{Q}_\xi$ at each scale, followed by pairwise transformation estimation.
    Finally, it identifies globally consistent inliers~$\mathcal{I}$ across all scales to refine correspondences based on consensus maximization~\cite{Sun22ral-TriVoC,Zhang24tpami-AcceleratingGloballyCM}.}
    \label{fig:pipeline}    
\end{figure*}

\subsection{Geometric bootstrapping}\label{sec:geometric}
\newcommand{\covsampledpoint}{\MC}
\newcommand{\Psampled}{\mathcal{P}_\text{sampled}}
\newcommand{\eigenidx}{a}

\noindent \textbf{Sphericity-based voxelization.} 
First, we determine a suitable voxel size $v$ by leveraging sphericity, quantified using eigenvalues~\cite{Hansen21remotesensing-Classification,Alexiou24jivp-PointPCA}, to reflect the spatial dispersion of point clouds.
To this end, we apply principal component analysis (PCA)~\cite{Lim21ral-Patchwork} to the covariance of sampled points, which can efficiently capture point dispersion by analyzing eigenvalues while remaining computationally lightweight.

Formally, let $h(\mathcal{P}, \mathcal{Q})$ be a function that selects the largest point cloud based on cardinality, let $g(\mathcal{P}, \delta)$ be a function that samples $\delta\%$ of points from a given point cloud,
and let $\covsampledpoint \in \mathbb{R}^{3 \times 3}$ be the covariance of $g(h(\mathcal{P}, \mathcal{Q}), \delta_v)$, where $\delta_v$ is a user-defined sampling ratio.
Then, using PCA, three eigenvalues $\lambda_\eigenidx$ and their corresponding eigenvectors $\vv_\eigenidx$ are calculated as follows:
\begin{equation}
\covsampledpoint \vv_\eigenidx = \lambda_\eigenidx \vv_\eigenidx, \quad \eigenidx \in \{1,2,3\},
\label{eq:pca}
\end{equation}
\noindent which are assumed to be $\lambda_1 \geq \lambda_2 \geq \lambda_3$. Then, using these properties, we can compute the \textit{sphericity} $\frac{\lambda_3}{\lambda_1}$~\cite{Alexiou24jivp-PointPCA}, which quantifies how evenly a point cloud is distributed in space.
Since LiDAR points are primarily distributed along the sensor's horizontal plane (\ie forming a disc-like shape), $\frac{\lambda_3}{\lambda_1}$ tends to be low compared to RGB-D point clouds.

\rev{In addition, as observed in \Cref{fig:key_elements}(a), the appropriate voxel size varies across scene scales and spatial distributions; thus, we set $v$ adaptively as follows:}
\newcommand{\coefficient}{\kappa}
\begin{equation}
    v =
    \begin{cases}
        \coefficient_{\text{spheric}} \sqrt{s}, & \text{if} \; \frac{\lambda_3}{\lambda_1} \geq \tau_v, \\
        \coefficient_{\text{disc}} \sqrt{s}, & \text{otherwise},
    \end{cases}
    \label{eq:voxel_size}
\end{equation}
\noindent where $\coefficient_{\text{spheric}}$ and $\coefficient_{\text{disc}}$ are constant user-defined coefficients across all datasets, $\tau_v$ is a user-defined threshold,
and $s$ is the length that represents the spread of points along the eigenvector corresponding to the smallest eigenvalue $\vv_3$~(\ie $s = \max(\Psampled \cdot \vv_3) - \min(\Psampled \cdot \vv_3)$). %
Consequently, as $\frac{\lambda_3}{\lambda_1}$ and $s$ adapt based on the environment (\ie indoor or outdoor) and the field of view of the sensor type (\ie RGB-D or LiDAR point cloud),
\Cref{eq:voxel_size} enables the adaptive setting of $v$.

Hereafter, for brevity, we denote $f_v(\mathcal{P})$ and $f_v(\mathcal{Q})$ simply as $\mathcal{P}$ and $\mathcal{Q}$, respectively, 

\newcommand{\rxi}{r_\xi}
\newcommand{\rxithres}{\tau_{\xi}}
\vspace{2mm}
\noindent \textbf{Density-aware radius estimation.}
Next, in contrast to some state-of-the-art approaches \cite{Ao23CVPR-BUFFER, Ao21cvpr-Spinnet} that use a single fixed user-defined search radius,
we determine $r$ at local, middle, and global scales, respectively, by considering the input point densities. %
Let neighboring search function within the radius $r$ around a query point $\srcpoint_q$ be:
\begin{equation}
\mathcal{N}\big(\srcpoint_q, \mathcal{P}, r\big) = \big\{ \srcpoint \in \mathcal{P} | \; \twonorm{\srcpoint - \srcpoint_q} \leq r \big\}.
\label{eq:search}
\end{equation}

\noindent Then, as presented in \Cref{fig:radius_search}(a), the radius for patch-wise descriptor generation for each scale $\rxi$ is defined as follows:
\begin{equation}
  \rxi = \argmin_{r} \left| \frac{1}{N} \sum_{\srcpoint_q \in \mathcal{P}_r} \text{card}\Big(\mathcal{N}\big(\srcpoint_q, \mathcal{P}_r, r\big) \Big) - \tau_{\xi} \right|,
  \label{eq:density_radius}
\end{equation}
where $\xi \in \{l, m, g\}$ denotes the scale level~(\ie local, middle, and global scale, respectively), 
$\tau_{\xi}$ denotes the user-defined threshold, which represents the desired neighborhood density~(\ie~average proportion of neighboring points relative to the total number of points), satisfying $\tau_l \leq \tau_m \leq \tau_g$ (accordingly, $r_l \leq r_m \leq r_g$ as presented in \Cref{fig:radius_search}(a)),
and $\mathcal{P}_r$ is a set of $N_r$ points sampled from $h(\mathcal{P}, \mathcal{Q})$, where $N_r$ is a user-defined parameter for radius estimation.
To account for cases where the points are too sparse, we set the maximum truncation radius $r_\text{max}$ as 
\rev{$\rxi \leftarrow \min(\rxi, r_\text{max})$}.

\subsection{Multi-scale patch embedder}\label{sec:tri-scale-patch}
\newcommand{\Pxi}{{\mathcal{P}_\xi}}
\newcommand{\Qxi}{{\mathcal{Q}_\xi}}

Next, with the voxelized point clouds $\mathcal{P}$ and $\mathcal{Q}$ and radii estimated by \Cref{eq:density_radius}, patch-wise descriptors are generated at each scale.

\vspace{2mm}
\noindent \textbf{Farthest point sampling.} As discussed in \Cref{sec:probpt_detector}, 
we sample $\Pxi$ from $\mathcal{P}$ at each scale using FPS to be free from a learning-based keypoint detector~(resp. $\Qxi$ from $\mathcal{Q}$).
Note that instead of extracting local, middle, and global-scale descriptors for the same sampled point~\cite{Yu21nips-CofiNet}, we independently sample separate points for each scale, as illustrated in \Cref{fig:pipeline}(b).
This is because we empirically found that different regions may require distinct scales for optimal feature extraction~\cite{Seo25ICCV-BUFFERX}.

\newcommand{\patchidx}{b}
\newcommand{\ppatchidx}{\mathcal{P}_\xi}
\newcommand{\qpatchidx}{\mathcal{Q}_\xi}
\newcommand{\SPxi}{\mathcal{S}^{\mathcal{P}}_\xi}
\newcommand{\SQxi}{\mathcal{S}^{\mathcal{Q}}_\xi}
\newcommand{\FPxi}{\mathcal{F}^{\mathcal{P}}_\xi}
\newcommand{\FQxi}{\mathcal{F}^{\mathcal{Q}}_\xi}
\newcommand{\CPxi}{\mathcal{C}^{\mathcal{P}}_\xi}
\newcommand{\CQxi}{\mathcal{C}^{\mathcal{Q}}_\xi}

\vspace{2mm}
\noindent \textbf{Mini-SpinNet-based descriptor generation.}
Using multiple radii $\rxi$, %
we sample patches at three distinct scales, providing a more comprehensive multi-scale representation.
Then, we use Mini-SpinNet~\cite{Ao23CVPR-BUFFER} for descriptor generation, which is a lightweight version of SpinNet~\cite{Ao21cvpr-Spinnet}.

\begin{figure}[t!]
 	\centering
 	\includegraphics[trim={15mm 0mm 0mm 13mm}, clip, width=1.0\columnwidth]{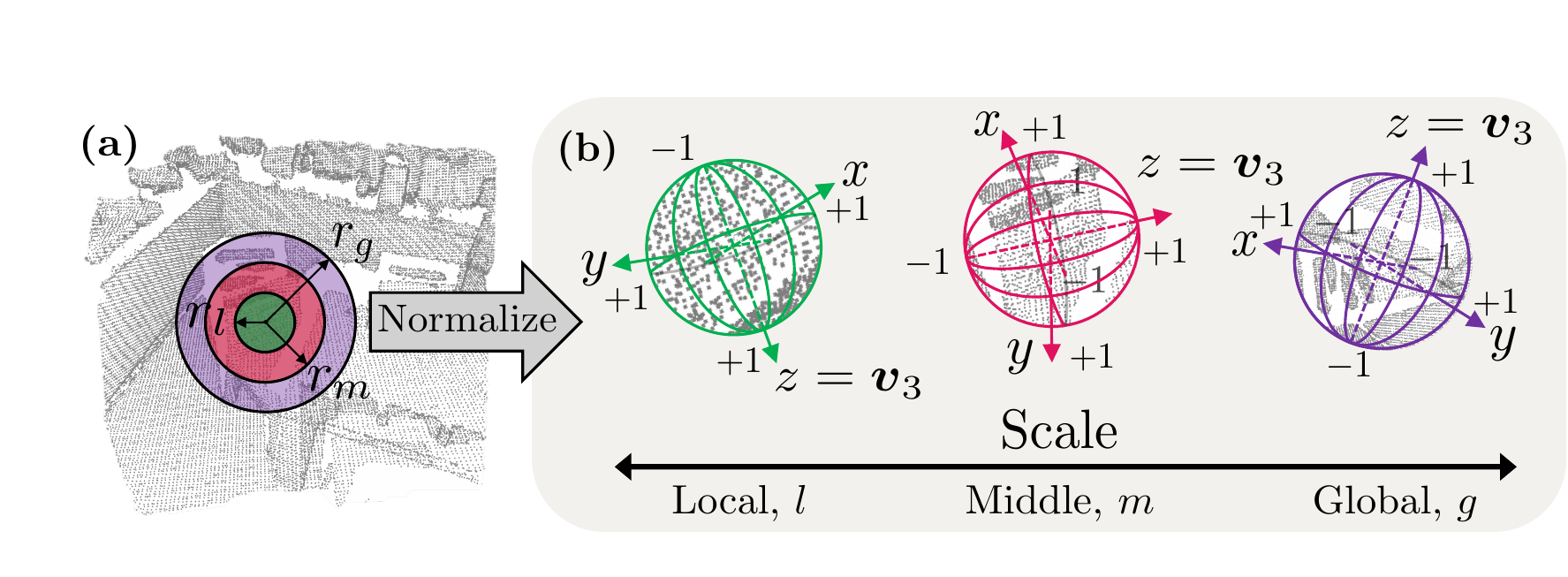}
    \caption{(a)~Visual description of local~($r_l$), middle~($r_m$), and global~($r_g$) radii for the same point to illustrate scale differences and (b) normalized patches ranging from $[-1, 1]$. Note that their reference frames follow the eigenvectors obtained from principal component analysis~(PCA)~\cite{Lim21ral-Patchwork,Alexiou24jivp-PointPCA}. The $z$-axis is assigned to the eigenvector $\vv_3$, which corresponds to the smallest eigenvalue.}
    \label{fig:radius_search}
\end{figure}

In particular, building on the insights from \Cref{sec:prob_scale_norm}, we ensure the scale of points in each patch is normalized to a bounded range of $[-1, 1]$ by dividing by $\rxi$; see \Cref{fig:radius_search}(b). 
By doing so, we can resolve the dependency on the training domain scale.
To maintain consistency across patches at all scales, we fix the patch size to $N_\text{patch}$ and randomly sample when a patch exceeds this size, ensuring a consistent number of points regardless of scale variations.



Finally, taking these normalized patches as inputs, Mini-SpinNet outputs a superset $\SPxi$ consisting of $D$-dimensional feature vectors~$\FPxi$ and cylindrical feature maps~$\CPxi$, which corresponds to $\Pxi$~(resp. $\SQxi$ consisting of $\FQxi$ and $\CQxi$ from $\Qxi$), as described in~\Cref{fig:pipeline}.
Note that while BUFFER~\cite{Ao23CVPR-BUFFER} utilizes learned reference axes to extract cylindrical coordinates, our approach defines the reference axes for each patch by applying PCA to the covariance of points within the patch, 
setting the $z$-direction as $\vv_3$ (as in \Cref{eq:pca} and illustrated by $z=\vv_3$ in \Cref{fig:radius_search}(b)), to eliminate potential dataset-specific inductive biases.

\subsection{Hierarchical inlier search}\label{sec:hierarchical}
\newcommand{\corrinit}{\mathcal{A}_\xi}
\newcommand{\CPximatched}{\widehat{\mathcal{C}}^{\mathcal{P}}_\xi}
\newcommand{\CQximatched}{\widehat{\mathcal{C}}^{\mathcal{Q}}_\xi}
\newcommand{\Pximatched}{\widehat{\mathcal{P}}_\xi}
\newcommand{\Qximatched}{\widehat{\mathcal{Q}}_\xi}

Here, we first perform intra-scale matching to get initial correspondences $\corrinit$ at each scale and then establish cross-scale consistent correspondences in a consensus maximization manner. 

\vspace{2mm}
\noindent \textbf{Intra-scale matching.} First, we perform nearest neighbor-based mutual matching~\cite{Lowe04ijcv} between $\FPxi$ and $\FQxi$, yielding matched correspondences $\corrinit$ at each scale. 
Using $\corrinit$, we extract the corresponding elements from $\CPxi$ and $\CQxi$, denoted as $\CPximatched$ and $\CQximatched$, and the sampled keypoints from $\Pxi$ and $\Qxi$ as $\Pximatched$ and $\Qximatched$, respectively~(\ie~$| \corrinit| = |\CPximatched | = |\CQximatched| = |\Pximatched | = |\Qximatched|$).

\vspace{2mm}
\noindent \textbf{Pairwise transformation estimation.} Next, using each cylindrical feature pair $\vc^{\vp} \in \CPximatched$ and $\vc^{\vq} \in \CQximatched$ at each scale, each of size $\mathbb{R}^{H \times W \times D}$, we calculate pairwise 3D relative transformation between two patches. 
Here, $H$, $W$, and $D$ denote the height, sector size for the yaw direction along the $z$-axis of the reference axes, and feature dimensionality of a cylindrical feature, respectively; see \Cref{fig:pipeline}.

As mentioned in \Cref{fig:radius_search}(b), since the cylindrical feature is aligned with the local reference axes via PCA, 
we estimate the relative 3D rotation by first computing the rotation between the patch-wise principal direction $\mathbf{v}_3^p$ (resp. $\mathbf{v}_3^q$) and the global $z$-axis of the point cloud coordinate frame, $\mathbf{z} = [0\ 0\ 1]^\top$, following Ao\etalcite{Ao23CVPR-BUFFER}. 
Specifically, using Rodrigues' rotation formula~\cite{Mebius07arxiv-Derivation}, the rotation matrix is given by:
\begin{equation}
\MR^{\vp} = \MI + \sin(\theta^{\vp}) [\vn^{\vp}]_{\times} + \big(1 - \cos(\theta^{\vp})\big) [\vn^{\vp}]_{\times}^2,
\end{equation}
\noindent where $\vn^{\vp} = \vv^{\vp}_{3} \times \vz$, $\theta^{\vp} = \cos^{-1} \left( \vv^{\vp}_{3} \cdot \vz \right)$, and $[\cdot]_\times$ denotes the skew operator (resp. $\MR^\vq$).
Thus, once the yaw rotation between the two patches $\MR_{\text{yaw}}$ is determined, the full 3D rotation can be obtained as $\MR = \left(\MR^\vq\right)^\intercal \MR_{\text{yaw}} \MR^\vp$.

As explained by Ao~\etalcite{Ao23CVPR-BUFFER}, $\vc^{\vp}$ and $\vc^{\vq}$ follow discretized SO(2)-equivariant representation; 
thus, by finding the yaw rotation that maximizes circular cross-correlation between $\vc^{\vp}$ and $\vc^{\vq}$, we can estimate the relative SO(2) rotation $\MR_\text{yaw}$. 
To this end, a 4D matching cost volume $\MV \in \mathbb{R}^{H\times W \times W \times D}$ is constructed to represent the sector-wise differences between $\vc^{\vp}$ and $\vc^{\vq}$.
Then, $\MV$ is processed by a 3D cylindrical convolutional network~(3DCCN)~\cite{Ao21cvpr-Spinnet}, mapping $\MV$ to a score vector $\boldsymbol{\beta}$ of size~$W$.  

By applying the softmax operation $\sigma(\cdot)$ to $\boldsymbol{\beta}$, we obtain $\sigma(\boldsymbol{\beta})$,
where the $w$-th element $\sigma_w(\boldsymbol{\beta}) \in [0, 1]$ represents the probability mass assigned to the discrete yaw rotation index~$w$.
Using this distribution, the expected rotation offset~$d$ is computed as follows:
\begin{equation}
d = \sum_{w=1}^{W}  \sigma_w(\boldsymbol{\beta}) \times w.
\label{eq:calc_d}
\end{equation}

Finally, $\MR_\text{yaw}$ is calculated as follows:
\begin{equation}
\MR_\text{yaw} = 
\begin{bmatrix}
\cos \left(\frac{2\pi d}{W} \right) & -\sin \left(\frac{2\pi d}{W} \right) & 0 \\ 
\sin \left(\frac{2\pi d}{W} \right) & \cos \left(\frac{2\pi d}{W} \right) & 0 \\ 
0 & 0 & 1
\end{bmatrix}.
\end{equation}
 
Subsequently, the translation vector is given by $\vt = \vq - \MR \vp$, where $\vp \in \Pximatched$ and $\vq \in \Qximatched$ are a matched point pair.

\vspace{2mm}
\noindent \textbf{Cross-scale consensus maximization.}
Then, using per-pair $(\MR, \vt)$ estimates from all scales, the 3D point pairs with the largest cardinality across scales should be selected as the final inlier correspondences $\mathcal{I}$, ensuring cross-scale consistency.
To achieve this, we formulate the cross-scale inlier selection as \emph{consensus maximization} problem~\cite{Sun22ral-TriVoC,Zhang24tpami-AcceleratingGloballyCM}.

Formally, by denoting $N=\sum_{\xi} |\corrinit|$, let $(\MR, \vt) \in \mathcal{T}$ be a candidate transformation  set of size $N$, 
and let $(\vp_n, \vq_n) \in \mathcal{D}$ be the set of matched point pairs, where $n \in \{1, \dots, N\}$, $\vp_n \in \bigcup_\xi \Pximatched$ and $\vq_n \in \bigcup_\xi \Qximatched$.
Then, $\mathcal{I}$ is estimated as follows:
\begin{gather}
\max _{(\MR, \vt) \in \mathcal{T}, \; \mathcal{I}} |\mathcal{I}|  \label{eq:consensus_max} \\
\text {s.t.} ~~ \twonorm{\MR \vp_n  + \vt - \vq_n} <  \epsilon, ~~ \forall (\vp_n, \vq_n) \in \mathcal{I} \subseteq \mathcal{D},\nonumber
\end{gather}
\noindent where $\epsilon$ is an inlier threshold.
That is, the objective in \Cref{eq:consensus_max} maximizes the number of inlier correspondences aggregated across multiple scales,
and the resulting transformation is estimated using a robust solver such as RANSAC~\cite{Fischler81} or TEASER++~\cite{Yang20tro-teaser}.

Pseudocode of our approach is presented in \Cref{alg:buffer_x}.

\newcommand{\annot}[1]{\textcolor{gray!80}{\% #1}}

\begin{adjustwidth}{-2em}{}
\begin{algorithm}[t!]
{\footnotesize
\SetAlgoLined
\textbf{Input:} \ Source cloud $\srccloud$ and target cloud $\tgtcloud$; User-defined parameters $\tau_v$, $\delta_v$, $N_r$, $[\tau_l, \tau_m, \tau_g]$, and $N_\text{FPS}$ \\
\textbf{Output:} 3D inliers $\mathcal{I}$ \\
$\mathcal{P}_r \leftarrow \texttt{select\_larger\_cloud}(\srccloud, \tgtcloud)$  \\
$\mathcal{P}_\text{sampled} = \texttt{sample}(\mathcal{P}_r, \delta_v)$ \annot{Sample $\delta_v$\% of cloud points} \\
\annot{Step 1. Geometric bootstrapping} \\
$v = \texttt{calc\_voxel\_size}(\mathcal{P}_\text{sampled}, \tau_v)$ \annot{See Eq.~\Cref{eq:voxel_size}} \\
$\mathcal{P} \leftarrow f_v(\mathcal{P}), \mathcal{Q} \leftarrow f_v(\mathcal{Q})$
\annot{Downsample the point clouds} \\
$\mathcal{P}_r \leftarrow \texttt{select\_larger\_cloud}(\srccloud, \tgtcloud)$  \\
$\mathcal{R} = \texttt{estimate\_radii}(\texttt{sample}(\mathcal{P}_r, N_r), [\tau_l, \tau_m, \tau_g]),$ \\
\qquad where $\mathcal{R} = [r_l, r_m, r_g] $ \annot{See Eq.~\Cref{eq:density_radius}} \\
\annot{Step 2. Multi-scale patch embedder} \\
$\mathcal{M}^\mathcal{P} = \varnothing, \mathcal{M}^\mathcal{Q} = \varnothing$ \annot{Containers of embedding output} \\
\For{$r_\xi$ \normalfont{in} $\mathcal{R}$}{
    $\mathcal{P}_\xi = \texttt{farthest\_point\_sampling}(\mathcal{P}, N_\text{FPS})$  \\
    $\mathcal{Q}_\xi = \texttt{farthest\_point\_sampling}(\mathcal{Q}, N_\text{FPS})$ \\
    $\mathcal{F}^\mathcal{P}_\xi, \mathcal{C}^\mathcal{P}_\xi = \texttt{MiniSpinNet}(\mathcal{P}_\xi, \mathcal{P}, r_\xi)$ \\
    $\mathcal{F}^\mathcal{Q}_\xi, \mathcal{C}^\mathcal{Q}_\xi = \texttt{MiniSpinNet}(\mathcal{Q}_\xi, \mathcal{Q}, r_\xi)$ \\
    $\mathcal{M}^\mathcal{P}.\texttt{append}((\mathcal{P}_\xi, \mathcal{F}^\mathcal{P}_\xi, \mathcal{C}^\mathcal{P}_\xi))$ \\
    $\mathcal{M}^\mathcal{Q}.\texttt{append}((\mathcal{Q}_\xi, \mathcal{F}^\mathcal{Q}_\xi, \mathcal{C}^\mathcal{Q}_\xi))$ \\
}
\annot{Step 3. Hierarchical inlier search} \\
$\mathcal{D} = \varnothing, \mathcal{T} = \varnothing$ \\ 
\For{$i$ \normalfont{in} \texttt{range}(\texttt{size}($\mathcal{M}^\mathcal{P}$))}{
    $(\mathcal{P}_\xi, \mathcal{F}^\mathcal{P}_\xi, \mathcal{C}^\mathcal{P}_\xi) = \mathcal{M}^\mathcal{P}[i]$ \\
    $(\mathcal{Q}_\xi, \mathcal{F}^\mathcal{Q}_\xi, \mathcal{C}^\mathcal{Q}_\xi) = \mathcal{M}^\mathcal{Q}[i]$ \\
    \annot{Step 3-1. Nearest neighbor-based intra-scale matching}\\ 
    $\mathcal{A}_\xi = \texttt{mutual\_matching}(\mathcal{F}^\mathcal{P}_\xi, \mathcal{F}^\mathcal{Q}_\xi)$ \\
    $ (\widehat{\mathcal{P}}_\xi, \widehat{\mathcal{Q}}_\xi, \widehat{\mathcal{C}}^{\mathcal{P}}_\xi, \widehat{\mathcal{C}}^{\mathcal{Q}}_\xi) = \texttt{filter}(\mathcal{M}^\mathcal{P}[i], \mathcal{M}^\mathcal{Q}[i], \mathcal{A}_\xi)$ \\
    \annot{Step 3-2. Pairwise transformation estimation}\\
    $\mathcal{T}_\xi = \texttt{calc\_pairwise\_R\_and\_t}(\widehat{\mathcal{C}}^{\mathcal{P}}_\xi, \widehat{\mathcal{C}}^{\mathcal{Q}}_\xi)$ \\
    $\mathcal{D}.\texttt{append}((\widehat{\mathcal{P}}_\xi, \widehat{\mathcal{Q}}_\xi))$, $\mathcal{T}.\texttt{append}(\mathcal{T}_\xi)$\\
}
\annot{Step 3-3. Cross-scale consensus maximization}\\
$\mathcal{I} = \texttt{consensus\_maximization}(\mathcal{D}, \mathcal{T})$ \annot{See Eq.~\Cref{eq:consensus_max}} \\
}
    \caption{BUFFER-X pipeline\label{alg:buffer_x}}
\end{algorithm}
\end{adjustwidth}

\subsection{Loss function and training}\label{sec:loss}
\noindent \textbf{Loss functions.}
Unlike BUFFER, which was trained in four stages, our network follows a relatively simpler two-stage training process thanks to its detector-free nature.
First, we train the feature discriminability of Mini-SpinNet descriptors using contrastive learning~\cite{Yew18eccv-3dfeatnet}, followed by training $d$ in \Cref{eq:calc_d} to improve transformation estimation accuracy.

In particular, we employ the Huber loss~\cite{Zhang97ivc-Parameter} $\rho_\text{Huber}(\cdot)$ for training $d$ while remaining robust to outliers~\cite{Barron19-generaladaptiverobustloss}, which is formulated as follows:

\begin{equation}
\rho_\text{Huber}(r) =
\begin{cases}
\frac{1}{2} r^2, & \text{if } |r| \leq \delta \\
\delta (|r| - \frac{1}{2} \delta), & \text{otherwise},
\end{cases}
\end{equation}

\noindent where $r$ denotes the residual and $\delta$ denotes the user-defined truncation threshold. 
Then, denoting the total number of data pairs by ${N_d}$, the $\gamma$-th predicted offset by $d_\gamma$, and the corresponding ground-truth offset by $d^*_\gamma$, the loss function $\mathcal{L}_{d}$ is defined as follows:

\begin{equation}
  \mathcal{L}_{d} = \frac{1}{N_d} \sum_{\gamma=1}^{N_d} \rho_\text{Huber}(d_\gamma - d^*_\gamma).
\end{equation}

\vspace{2mm}
\noindent \textbf{Patch distribution augmentation.} 
Furthermore, \rev{BUFFER-X employs} an inter-patch point distribution augmentation to allow Mini-SpinNet to experience a wider variety of patch distribution patterns.
Specifically, we empirically sample the radius within $[\frac{2}{3}r, \frac{4}{3}r]$ based on a uniform probability. 
As mentioned in \Cref{sec:tri-scale-patch}, since $N_\text{patch}$ points within the radius are randomly selected as an input, a diverse set of patterns can be provided as $r$ varies.

Notably, training is conducted using only a single scale. This leverages the scale normalization characteristic of BUFFER-X, making it unnecessary to train with multi-scale separately.

\section{\fastversion: Reducing Computation Time}\label{sec:fastbufferx}

While BUFFER-X achieves strong generalization through multi-scale matching,
which is essential for handling diverse registration scenarios, the computational cost can be reduced for certain registration pairs where reliable alignment can be achieved with fewer scales.
The key insight is that not all pairs require processing all three scales (\ie some easier pairs may find sufficient inliers at local or middle scales, while challenging pairs benefit from the full multi-scale hierarchy).
To achieve faster inference while preserving this multi-scale robustness, we \rev{present} \fastversion{} \rev{from a holistic system perspective} by (i)~decreasing the number of scales processed via adaptive early exit when sufficient evidence is found and (ii)~decreasing solver time at each processed scale by replacing RANSAC with \rev{the fast and robust correspondence-pruning and pose-estimation modules developed in KISS-Matcher~\cite{Lim25icra-KISSMatcher}.}
\rev{While KISS-Matcher does not address descriptor generation or feature matching, integrating its modules with the BUFFER-X front-end allows \fastversion to span the complete registration pipeline from raw point clouds to the estimated transformation.}
\rev{The complete procedure is summarized in Algorithm~S1 in the supplementary material.}

\subsection{Strategy 1: Decreasing the number of scales processed}

The first strategy reduces computation by adaptively terminating after processing only one scale when sufficient inliers are detected, thereby reducing the total number of feature descriptors that need to be computed.
Importantly, this does not claim that a single scale is universally sufficient; rather, it adapts computational effort based on the difficulty of each registration pair, processing additional scales only when needed.
After obtaining correspondences at the middle scale, we estimate the initial pose and count the number of inliers.
This adaptive early exit approach is inspired by recent advances in image feature matching~\cite{Lindenberger23ICCV-LightGlue} and robust estimation~\cite{Shi21icra-robin}, where similar strategies have proven effective in balancing computation reduction with matching quality.

Formally, we first process only the middle scale (\ie~$\xi = m$) to obtain inliers $\calI_m$ through the matching procedure described through \Cref{sec:prob_user_defined} and \Cref{sec:hierarchical}.
If $|\calI_m|$ exceeds a user-defined threshold $\tau_N$, we terminate early and skip the local and global scales entirely, as sufficient consensus has been found for reliable registration.
Lower values of $\tau_N$ enable more aggressive early exit (processing fewer scales on average),
while higher values ensure more robust registration by processing additional scales more frequently.
Otherwise, we proceed with the full three-scale pipeline as in standard \oursname with an efficient solver presented in \Cref{sec:km_solver}. 

\subsection{Strategy 2: Decreasing solver time}\label{sec:km_solver}

The second strategy reduces \rev{the time required for correspondence pruning and pose estimation} at each scale by replacing RANSAC with a faster solver~\cite{Lim25icra-KISSMatcher}.
While RANSAC also employs an early exit concept,
in the worst case, traditional RANSAC requires numerous random sampling iterations (\eg 50K iterations in our experiments) to achieve reliable registration.
Empirically, for noisy point cloud pairs, this takes approximately 0.1\,sec on an AMD Ryzen Threadripper 7960X 24-Core processor, which is non-negligible and accumulates with multi-stage pose estimation.
To reduce this solver time, we employ deterministic graph-theoretic $k$-core pruning~\cite{Shi21icra-robin} followed by a GNC solver~\cite{Yang20ral-GNC}, instead of iterative random sampling.
\rev{These correspondence-pruning and pose-estimation modules were developed and evaluated in our previous work, KISS-Matcher~\cite{Lim25icra-KISSMatcher}; we collectively refer to them as \textit{\kmsolver}.}

Specifically, \kmsolver~first constructs a compatibility graph $\calG(\calV, \calE)$ where vertices represent correspondences and edges connect pairs satisfying the geometric consistency constraint:
\begin{equation}
-2\beta \leq \|\M{q}_j - \M{q}_{j'}\| - \|\M{p}_i - \M{p}_{i'}\| \leq 2\beta
\end{equation}
for correspondences $(\M{p}_i, \M{q}_j)$ and $(\M{p}_{i'}, \M{q}_{j'})$, where $\beta = 1.5v$ is the inlier noise bound.
The maximum $k$-core of this graph forms a candidate inlier set in linear time $O(|\calV| + |\calE|)$, which is then refined by a GNC solver.
In contrast, RANSAC requires $O(N_\text{iter} \cdot |\calV|)$ complexity, where $N_\text{iter}$ denotes the number of iterations (typically 50K in our experiments).
This substantial reduction in computational complexity makes \kmsolver~particularly beneficial when combined with early exit, enabling rapid evaluation of whether sufficient inliers have been found for reliable registration.

\subsection{Combined effect: Faster and more reliable inference}

The combination of these two strategies enables both computational efficiency and reliable decision-making.
The early exit mechanism adaptively processes additional scales only when needed, while \kmsolver~ensures efficient pose estimation with reliable inlier detection.

Critically, successful early exit requires accurately determining when sufficient evidence has been found for successful registration versus when additional scales are necessary.
Interestingly, \kmsolver~produces well-separated inlier distributions compared to RANSAC, making the final inliers easily discernible.
This clear separation enables confident early termination decisions, maximizing correct early exits while minimizing premature termination; see \Cref{sec:fast_buffer_x_exp}.

\newcommand{\oracle}{\raisebox{-0.6ex}{\includegraphics[width=0.30cm]{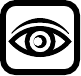}}}
\newcommand{\scalealign}{\raisebox{-0.6ex}{\includegraphics[width=0.30cm]{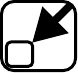}}}

\begin{figure*}[t!]
  \centering
  \begin{minipage}{\textwidth}
  \subfloat[]{
    \includegraphics[width=0.312\textwidth]{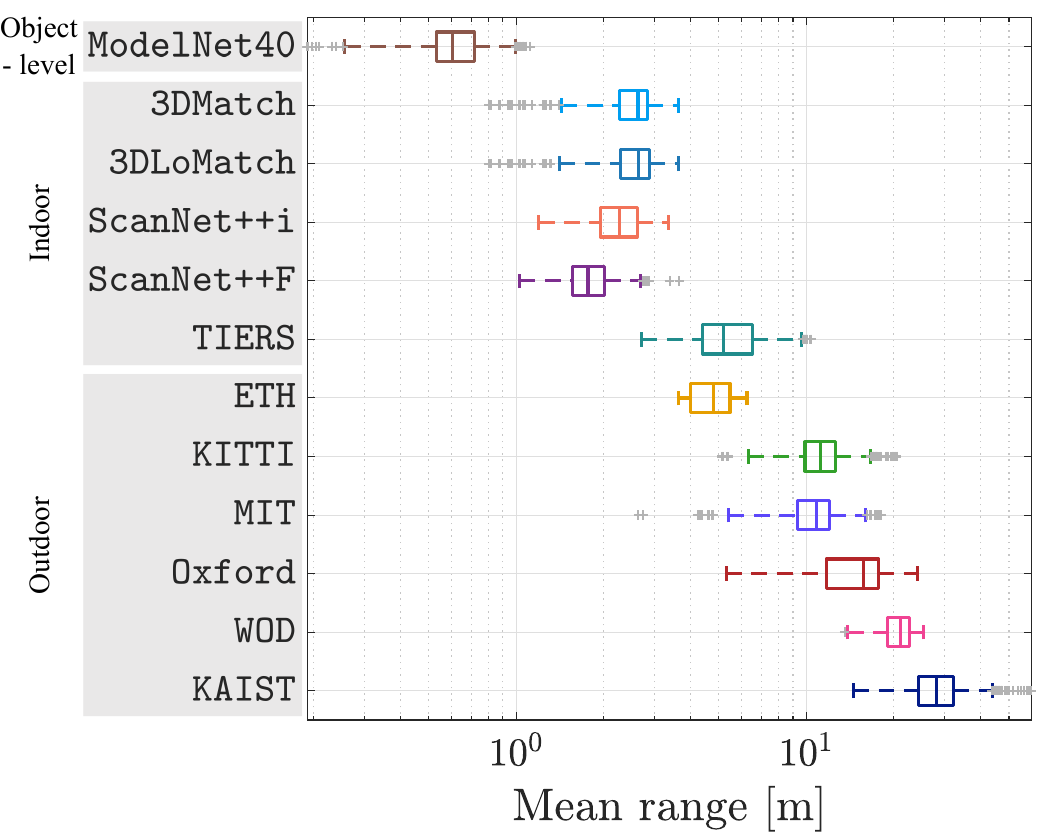}
  }
  \subfloat[]{
    \includegraphics[width=0.22\textwidth]{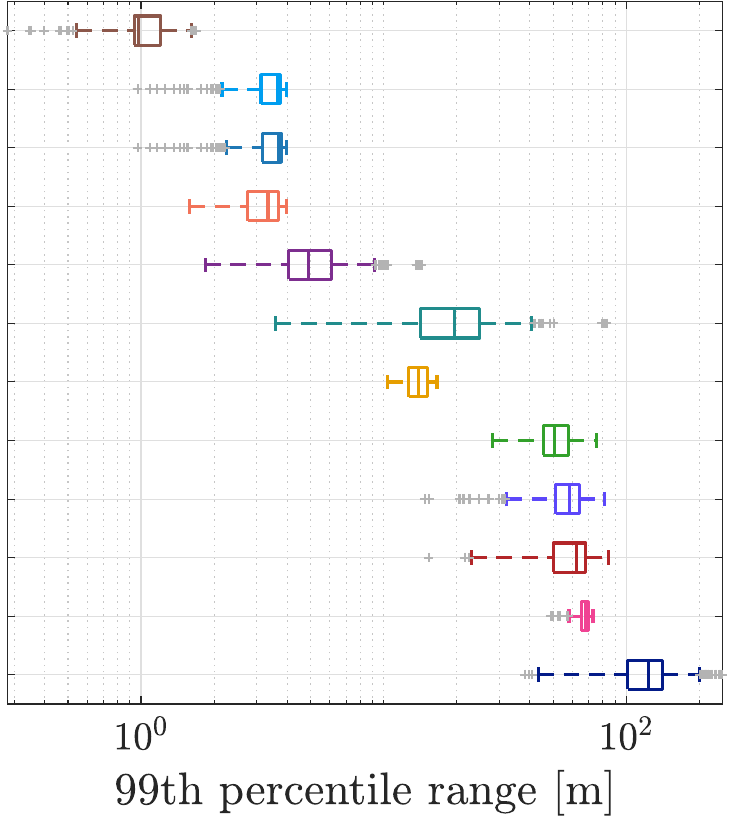}
  }
  \subfloat[]{
    \includegraphics[width=0.42\textwidth]{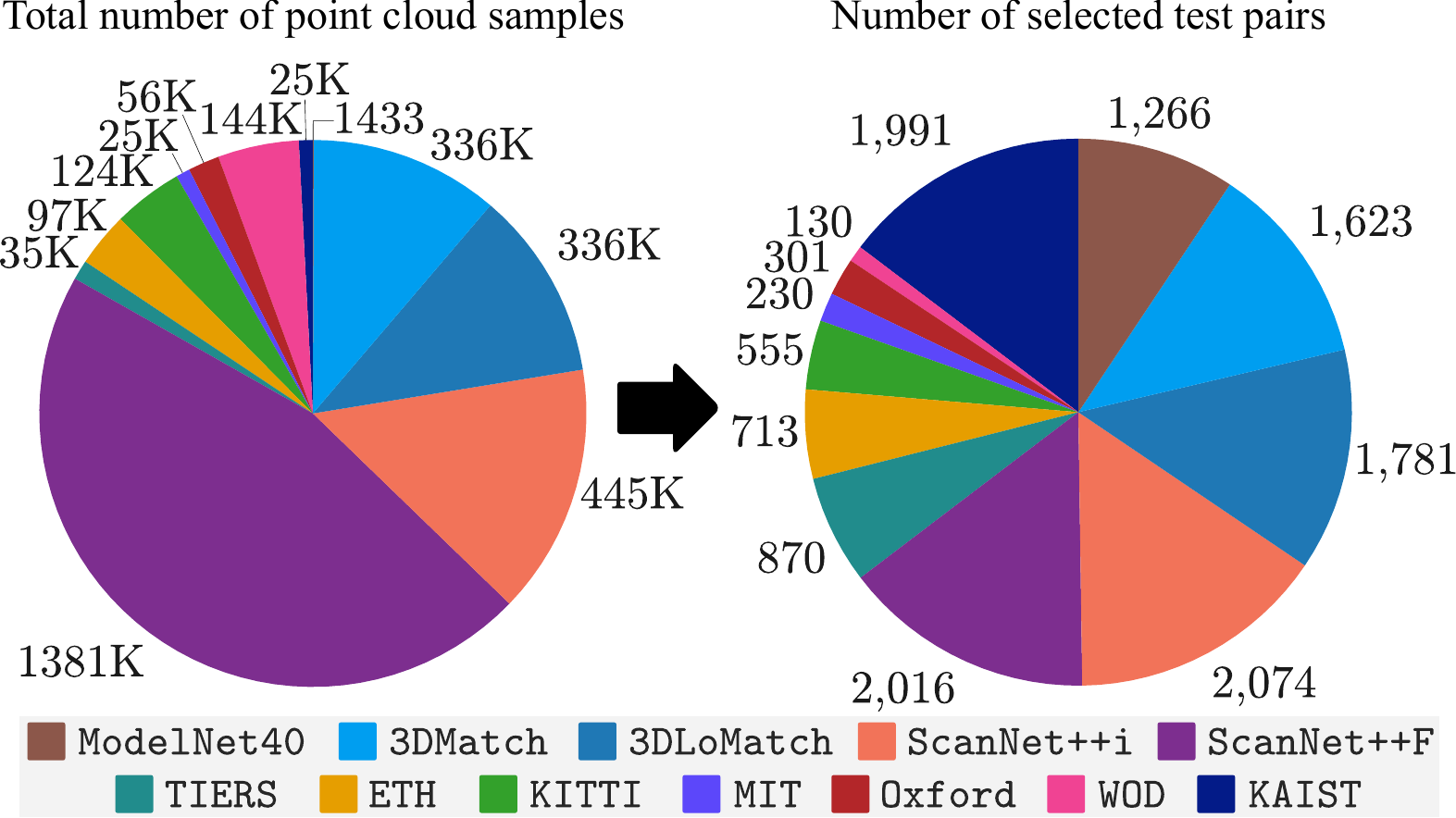}
  }  
  \end{minipage}
  \begin{minipage}{\textwidth}
  \subfloat[]{
    \includegraphics[width=0.24\textwidth]{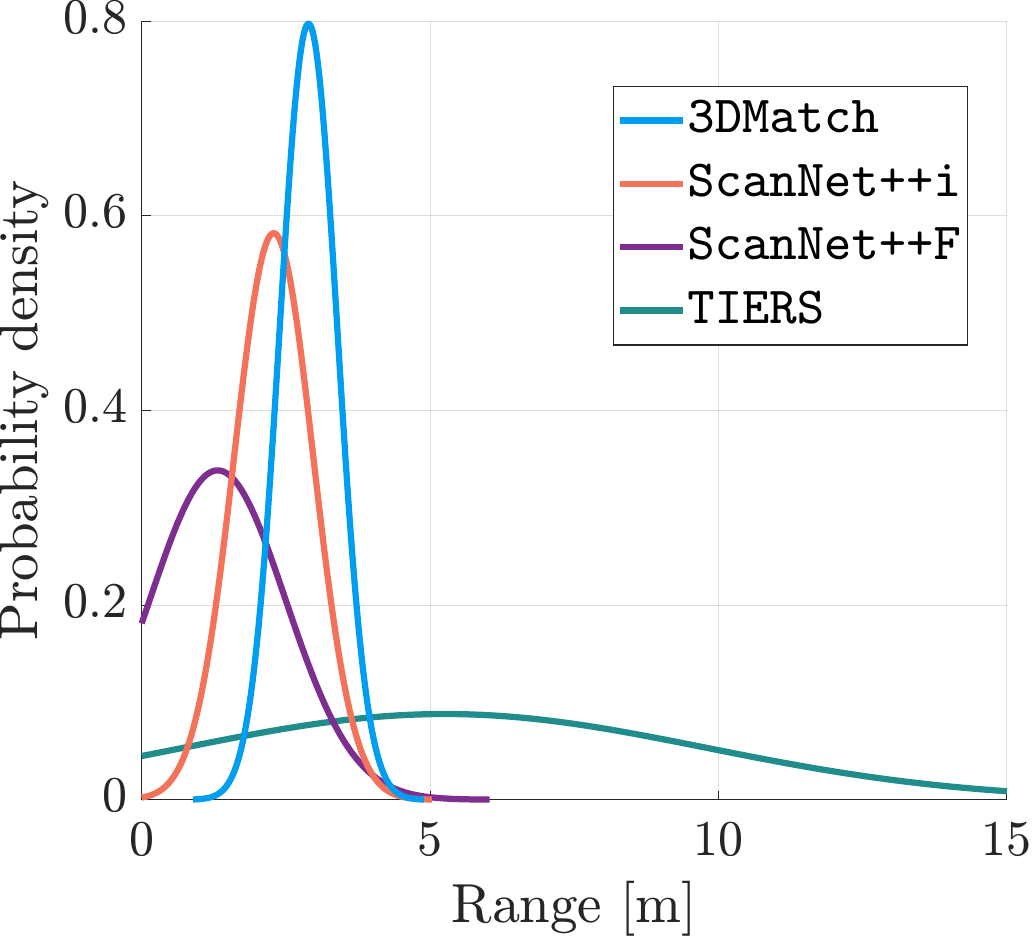}
  }
  \subfloat[]{
    \includegraphics[width=0.24\textwidth]{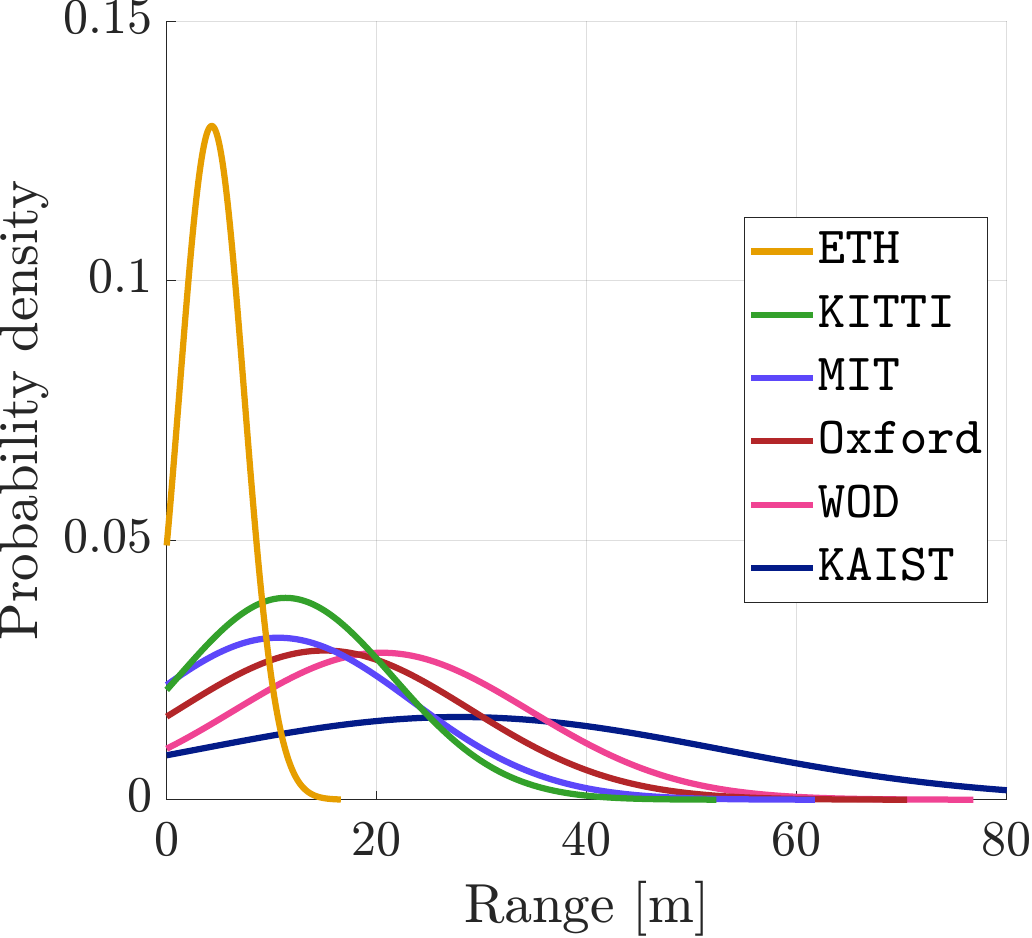}
  }      
  \subfloat[]{
    \includegraphics[width=0.24\textwidth]{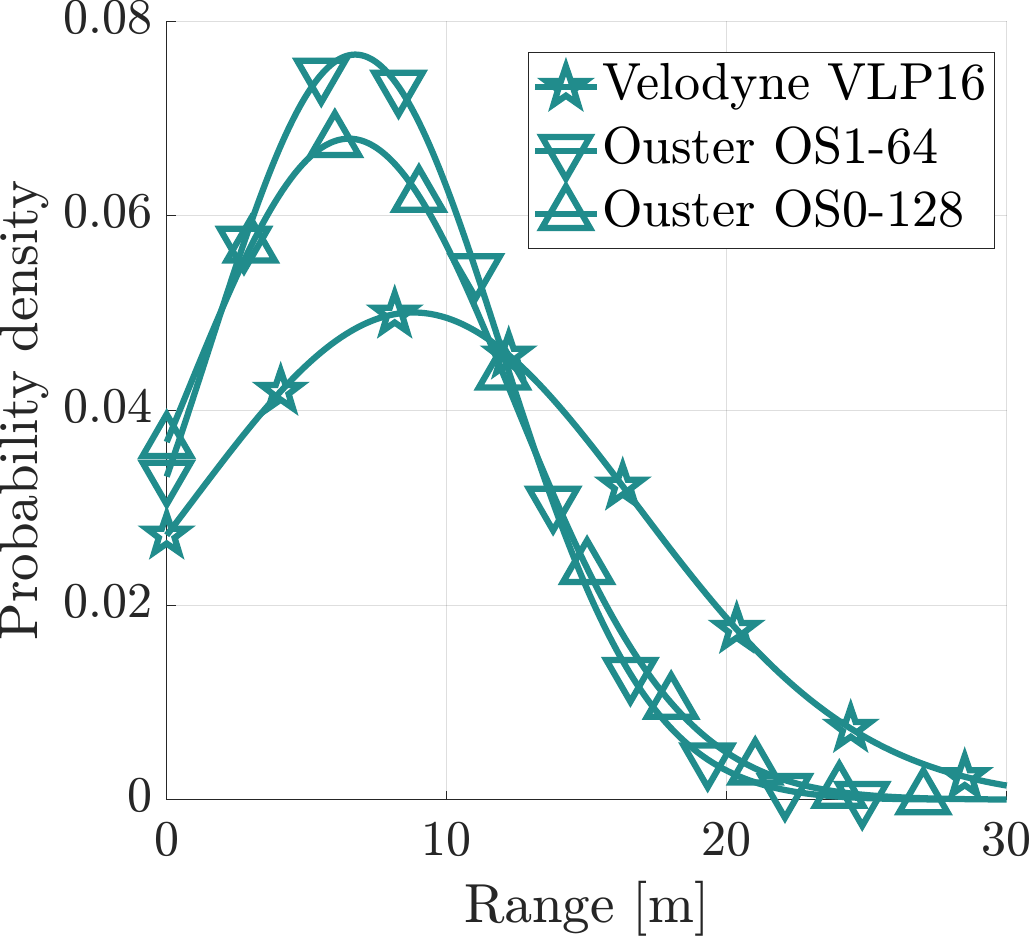}
  }
  \subfloat[]{
    \includegraphics[width=0.24\textwidth]{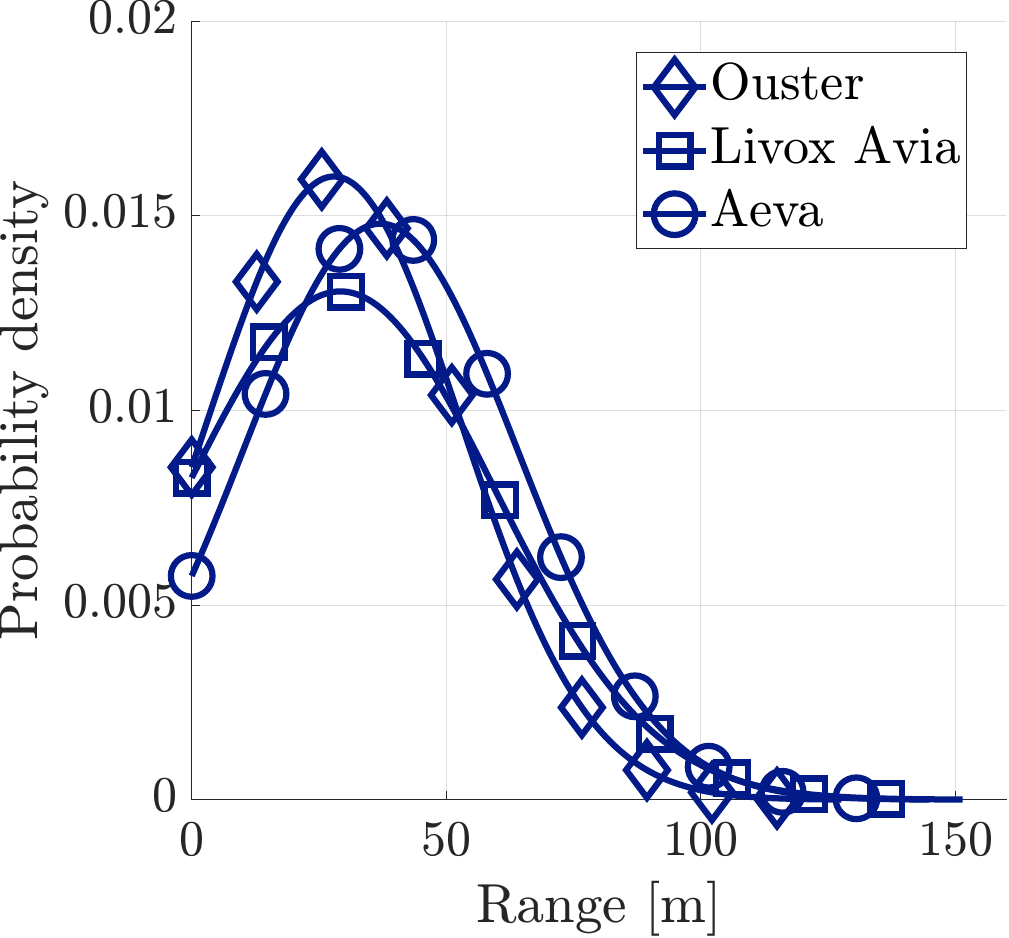}
  }
  \end{minipage}
  \vspace{-1.0mm}
  \caption{Dataset statistics and characteristics analysis, which demonstrates the diversity of point cloud characteristics across different environments, sensor types, and acquisition scenarios in our generalizability benchmark.
    (a)-(b) Box plots of mean and 99th percentile range per frame across different datasets, respectively. 
    (c) Distribution of the number of point clouds and the number of selected test pairs across indoor and outdoor environments.
    (d)-(e)~Gaussian distribution of ranges for indoor and outdoor scenes, respectively, and (f)-(g) those for different sensor types in the \TIERS and \KAIST sequences.}
  \label{fig:data_graphs}
  \vspace{-3mm}
\end{figure*}

\begin{figure*}[t!]
  \centering
  \subfloat[]{
    \includegraphics[width=0.26\textwidth]{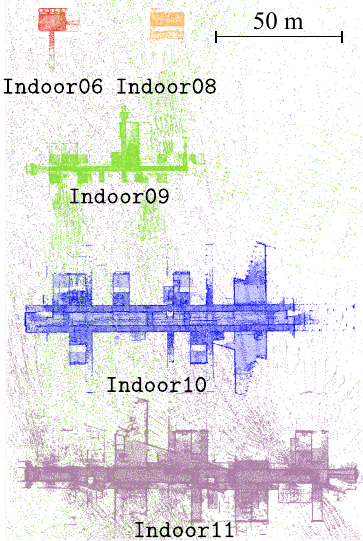}
  }
  \subfloat[]{
    \includegraphics[width=0.35\textwidth]{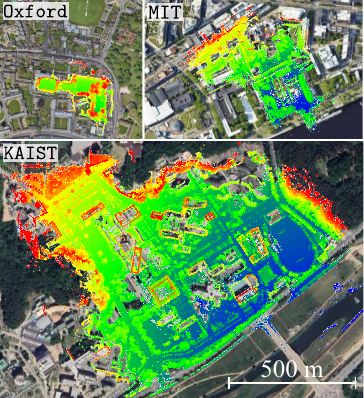}
  }
  \subfloat[]{
    \begin{minipage}[b]{0.31\textwidth}
        \includegraphics[width=\linewidth]{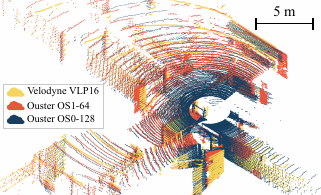}
        \includegraphics[width=\linewidth]{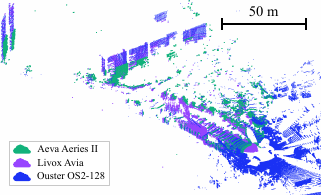}
    \end{minipage}
  }
  \vspace{-1.0mm}
  \caption{(a)~Different scales of sequences from the {\TIERS} dataset~\cite{Qingqing22iros-TIERS} in our experiments. While each scan is utilized for our evaluation, we build and then visualize map clouds using LiDAR point cloud scans and corresponding poses to illustrate the different scales of the surroundings. (b) Scale comparison of three sequences: {\Oxford} from the NewerCollege dataset~\cite{Ramezani20iros-NewerCollege}, {\MIT} from Kimera-Multi~\cite{Tian23iros-KimeraMultiExperiments}, and {\KAIST} from the HeLiPR dataset~\cite{Jung23ijrr-HeLiPR} at the same scale~(\ie~500\,m).
  Note that although these sequences fall under the same campus category, their scales differ. For clarity, the map clouds are visualized with respect to their $z$ values. (c) (T-B) Examples of visualized LiDAR scans from different LiDAR sensors in the \texttt{Indoor10} of {\TIERS} dataset~\cite{Qingqing22iros-TIERS} and \texttt{KAIST05} sequence of the HeLiPR dataset~\cite{Jung23ijrr-HeLiPR}. Note that even in the same environment, differences in the number of LiDAR rays and the field of view result in point clouds with different patterns.}
  \label{fig:campus_scale}
  \vspace{-2.5mm}
\end{figure*}

\section{Experiment Setups}

\subsection{Datasets for generalizability benchmark}\label{sec:benchmark}

We designed our generalizability benchmark using twelve datasets~\cite{Wu15cvpr-ModelNet,Zeng17cvpr-3dmatch,Huang21cvpr-PREDATORRegistration, Yeshwanth23iccv-Scannet++,Qingqing22iros-TIERS, Sun20cvpr-WaymoDataset, Geiger13ijrr-KITTI, Pomerleau12ijrr-ETH, Jung23ijrr-HeLiPR, Tian23iros-KimeraMultiExperiments, Ramezani20iros-NewerCollege} as follows.
Quantitative analyses can be found in \Cref{fig:data_graphs}.

\begin{itemize}
  \item{\ModelNet} follows conventional protocol used in 3D CAD registration benchmark~\cite{Wu15cvpr-ModelNet}. Note that the scale of \ModelNet is up to 15$\times$ smaller than those in outdoor sequences (see Figs.~\ref{fig:data_graphs}(a) and (b)).

\item{\ThreeDMatch} follows the conventional protocol proposed by Zeng~\textit{et al.}~\cite{Zeng17cvpr-3dmatch}.

\item{\ThreeDLoMatch} follows the conventional protocol proposed by Huang~\textit{et al.}~\cite{Huang21cvpr-PREDATORRegistration}.
This dataset is derived from {\ThreeDMatch} by selectively extracting pairs with low overlap (\ie~10--30\%), allowing for the evaluation of robustness to low-overlap scenarios.

\item{\ScanNetppi} is from \texttt{ScanNet++}~\cite{Yeshwanth23iccv-Scannet++}.
  There are depth images captured using a LiDAR sensor attached to an iPhone 13 Pro.
  These depth images were converted into point clouds using the toolbox provided by 3DMatch~\cite{Zeng17cvpr-3dmatch}.
  To generate dense point cloud fragments, 50 consecutive frames were accumulated.
  Finally, pairs with an overlap ratio of at least 0.4 (\ie 40\% overlap between two fragments) were selected as the final test pairs.

\item{\ScanNetppF} 
  is also from \texttt{ScanNet++}~\cite{Yeshwanth23iccv-Scannet++}. 
  Since the dataset provides only a merged PLY map and stationary FARO LiDAR poses, but no individual scans, we generate virtual scans by raycasting from each pose into the merged point cloud.
The ray sampling follows the scanner's horizontal and vertical angular resolutions~\cite{Ryu23cvpr-InstantDomainAugmentation}, and the nearest intersection along each ray is selected to simulate realistic scans.

\item{\TIERS} consists of \texttt{Indoor06}, \texttt{Indoor08}, \texttt{Indoor09}, \texttt{Indoor10}, and \texttt{Indoor11} sequences in the TIERS dataset~\cite{Qingqing22iros-TIERS}, as presented in \Cref{fig:campus_scale}(a).
In particular, we used data obtained from the Velodyne VLP-16, Ouster OS1-64, and Ouster OS0-128. 
Although additional sensors are available, point clouds from the Livox Horizon and Livox Avia can contain too few points or capture only partial wall surfaces, which are not suitable for registration evaluation.

\item{\ETH} is from the \texttt{gazebo\_summer}, \texttt{gazebo\_winter}, \texttt{wood\_autumn}, and \texttt{wood\_summer} sequences of the dataset proposed by Pomerleau~\textit{et al.}~\cite{Pomerleau12ijrr-ETH}.
The original dataset contains a wider variety of scenes; however, following the existing protocol proposed by Ao~\textit{et al.}~\cite{Ao23CVPR-BUFFER}, we used four sequences.

\item{\KITTI} follows the conventional protocol proposed by Yew~\textit{et al.}~\cite{Yew18eccv-3dfeatnet}. In test scenes, \texttt{08}, \texttt{09}, and \texttt{10} sequences are employed. %

\item{\MIT} is from \texttt{10\_14\_acl\_jackal} sequence of the Kimera-Multi dataset~\cite{Tian23iros-KimeraMultiExperiments}, which is acquired using a Velodyne VLP 16 sensor on the MIT campus. 
We refer to a subset of it as {\MIT} in our paper to highlight that our dataset was curated with consideration for geographic and cultural environments.
Similarly, we use {\Oxford}~(from the NewerCollege dataset~\cite{Ramezani20iros-NewerCollege}) and {\KAIST} (from the HeLiPR dataset~\cite{Jung23ijrr-HeLiPR}) to emphasize the institutions for the same reason.

\item{\Oxford} is from the \texttt{01}, \texttt{05}, and \texttt{07} sequences of the NewerCollege dataset~\cite{Ramezani20iros-NewerCollege}. 
  An interesting aspect is that \texttt{01} and \texttt{07} were acquired by a handheld setup, while the \texttt{05} sequence was acquired using a quadruped robot.
  As shown in \Cref{fig:campus_scale}(b), the campus scale was relatively small, so using only a single sequence generates only a few test pairs.
  For that reason, to ensure at least a similar number of test pairs as \MIT, we used three sequences. %

\item{\WOD} follows the protocol proposed by Liu~\textit{et al.}~\cite{Liu24cvpr-Extend}. This dataset is from the Waymo Open Dataset~\cite{Sun20cvpr-WaymoDataset} Perception dataset by extracting LiDAR sequences and corresponding pose files. %

\item{\KAIST} is from the \texttt{KAIST05} sequence of the HeLiPR dataset~\cite{Jung23ijrr-HeLiPR}. 
  Originally, HeLiPR contains multiple sequences, but each sequence in the HeLiPR is much longer than those in {\MIT} and {\Oxford}, resulting in many more test pairs compared to other campus scenes~(see \Cref{fig:data_graphs}(c)).
  For this reason,  we balanced the datasets by using only one sequence. %

\end{itemize}

By using these datasets, we evaluate the registration approaches on four aspects.
First, we assess \textit{variation in environmental scales}, ranging from small objects to large indoor and outdoor environments.
Unlike our previous work~\cite{Seo25ICCV-BUFFERX} that primarily used indoor and outdoor scenes, we include \ModelNet, which is an object-scale CAD-based point cloud dataset,~to challenge the domain shift.
Although all \TIERS~sequences were collected within the same indoor building, they were captured in diverse environments such as rooms, classrooms, and hallways, with significantly varying scales of surroundings.
Note that even within indoor or outdoor scenes, variations in scale~(or range) are significantly different, as presented in Figs.~\ref{fig:data_graphs}(d), \ref{fig:data_graphs}(e), \ref{fig:campus_scale}(a), and \ref{fig:campus_scale}(b). 

Second, we evaluate robustness to \textit{different scanning patterns with different sensor types}.
Using {\KAIST} and \TIERS, we assess whether the same space remains robust to different scanning patterns, as variations in the number of laser rays and sensor patterns result in different representations~(\Cref{fig:campus_scale}(c)).

Third, we examine \textit{acquisition setups} to evaluate beyond the typical setting where indoor scanning is performed using handheld devices while outdoor scanning is conducted using vehicles.
We include \TIERS~(acquired using a sensor cart), {\Oxford}~(acquired using both a handheld device and a quadruped robot), and {\MIT}~(captured using a mobile robot with planar motion but significantly more roll and pitch motion compared to a vehicle).

Fourth, we assess \textit{diversity of geographic and cultural environments}.
That is, we leveraged datasets collected by different teams from campuses in Asia, Europe, and the USA to evaluate whether such geographic and cultural variations have an impact on performance.

\vspace{2mm}
\noindent{\bf Training settings.} To evaluate generalization capability, all methods, including state-of-the-art approaches~\cite{Choi19iccv-FCGF,Huang21cvpr-PREDATORRegistration,Qin23tpami-GeoTransformer,Ao23CVPR-BUFFER,Yao24iccv-PARENet} and our \oursname, are trained on a single dataset, such as \ThreeDMatch~\cite{Zeng17cvpr-3dmatch} or \KITTI~\cite{Geiger13ijrr-KITTI}.
Using the same hyperparameters as BUFFER~\cite{Ao23CVPR-BUFFER}, we conducted a two-stage optimization~(\ie~Mini-SpinNet is first trained, followed by training the 3DCCN) using the Adam optimizer~\cite{Kingma15iclr-Adam} with a learning rate of 0.001, a weight decay of 1e-6, and a learning rate decay of 0.5.
We used NVIDIA GeForce RTX 3090 with AMD EPYC 7763 64-Core.

\vspace{2mm}
\noindent \textbf{Testing settings.} For existing datasets~\cite{Zeng17cvpr-3dmatch,Huang21cvpr-PREDATORRegistration,Sun20cvpr-WaymoDataset, Geiger13ijrr-KITTI, Pomerleau12ijrr-ETH},
we follow the conventional given pairs and experimental protocols.
The parameters of our approach are summarized in \Cref{table:bufferx_params}.
\rev{Throughout the generalization tests, we use a single parameter configuration, as listed in \Cref{table:bufferx_params}, for all target datasets without target-domain retraining, fine-tuning, or parameter tuning.}

\vspace{2mm}
\noindent \textbf{Evaluation Metrics.} As a key metric, we use the success rate, which directly assesses the robustness of global registration~\cite{Lim24ijrr-Quatropp}.
Specifically, a registration is deemed successful if the translation and rotation errors are within $\tau_\text{trans}$ and $\tau_\text{rot}$, respectively~\cite{Yew18eccv-3dfeatnet}. 
For successful cases, we evaluated the performance using relative translation error (RTE) and relative rotation error (RRE), which are defined as follows:

\newcommand{\numsuc}{N_\text{success}}
{\footnotesize
\begin{itemize}
	\item $\text{RTE}= \sum_{n=1}^{\numsuc} (\vt_{n, \text{GT}}-{\hat{\vt}}_{n})^{2} / \numsuc$,
	\item $\text{RRE}= \frac{180}{\pi} \sum_{n=1}^{\numsuc} | \cos^{-1} (\frac{\operatorname{Tr}\left({\hat{\MR}}_{n}^{\intercal} \MR_{n, \text{GT}}\right)-1}{2}) | / \numsuc $
\end{itemize}
}

\noindent where $\vt_{n, \text{GT}}$ and $\MR_{n, \text{GT}}$ denote the $n$-th ground truth translation and rotation, respectively; $\numsuc$ represents the number of successful registration.

\begingroup
\begin{table}[t!]
	\setlength{\tabcolsep}{4pt}
  \centering
	\caption{\rev{Global configuration used by BUFFER-X. The same values are fixed for all 12 datasets without any per-dataset tuning.}}
	{\scriptsize
		\begin{tabular}{ccc}
			\toprule \midrule
			Param. & Description & Value  \\ \midrule
      $\kappa_\text{spheric}$ & \rev{Coefficient for voxel size when sphericity is high} & \rev{0.15} \\
      $\kappa_\text{disc} $   & \rev{Coefficient for voxel size when sphericity is low}  & \rev{0.10} \\
      $\tau_v $               & \rev{Threshold in Eq.~\Cref{eq:voxel_size}}  & 0.05 \\
      $\tau_l $         & \rev{Threshold for the local ($l$) search radius}  & 0.005 \\
      $\tau_m $         & \rev{Threshold for the middle ($m$) search radius}  & 0.02 \\
      $\tau_g $         & \rev{Threshold for the global ($g$) search radius}  & 0.05 \\
      $\delta_v $         & \rev{Sampling ratio for sphericity-based voxelization} & 10\% \\
      $N_r $         & \rev{Number of sampling points for radius estimation} & 2,000 \\
      $r_\text{max}$          & Maximum radius  & 5.0\,m \\ \midrule
      $N_\text{FPS} $         & \rev{Number of sampled points by FPS} & 1,500 \\
      $N_\text{patch} $         & \rev{Maximum number of points in each patch}  & 512 \\
      $\delta$  & Truncation threshold for Huber loss & 1.0 \\ 
      $H$  & Height of the cylindrical map in Mini-SpinNet & 7 \\ 
      $W$  & Sector size of the cylindrical map in Mini-SpinNet & 20 \\ 
      $D$  & Feature dimension of the cylindrical map in Mini-SpinNet & 32 \\ \midrule \bottomrule
	\end{tabular}
	}
	\label{table:bufferx_params}
\end{table}
\endgroup

\newcommand{\subline}{ \cmidrule(lr){1-2} \cmidrule(lr){3-5} \cmidrule(lr){6-8} \cmidrule(lr){9-11}}
\newcommand{\errormetrics}{ RTE [cm] \; $\downarrow$ & RRE [$^\circ$] \; $\downarrow$ & Succ. [\%] \; $\uparrow$ }
\definecolor{myemerald}{rgb}{0.753, 0.898, 0.804}
\definecolor{mylightgreen}{rgb}{0.894, 0.933, 0.745}
\definecolor{myyellow}{rgb}{0.996, 0.972, 0.780}
\newcommand{\cg}{\cellcolor{gray!15}} 
\newcommand{\firstc}{\cellcolor{myemerald!100}}
\newcommand{\secondc}{\cellcolor{mylightgreen!100}}
\newcommand{\thirdc}{\cellcolor{myyellow!100}}
\newcommand{\scaleup}{\raisebox{-0.6ex}{\includegraphics[width=0.30cm]{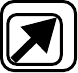}}}
\newcommand{\ora}{+ \oracle}
\newcommand{\osa}{+ {\oracle} + {\scalealign}}
\newcommand{\osu}{+ {\oracle} + {\scaleup}}
\begingroup
\begin{table*}[t!]
        \vspace{-2mm}
        \caption{Quantitative performance comparison of registration success rates to evaluate generalization capability.
          Deep learning-based models were trained only on {\ThreeDMatch}~\cite{Zeng17cvpr-3dmatch} and \KITTI~\cite{Geiger13ijrr-KITTI}, respectively, and to ensure a fair comparison, RANSAC was employed for pose estimation across all learning-based approaches, with a maximum of 50K iterations. 
        For conventional methods, FPFH~\cite{Rusu09icra-fast3Dkeypoints} was used for feature extraction.}
        \setlength{\tabcolsep}{2pt}
        \centering
	{\scriptsize
		\begin{tabular}{l|l|cccccccccccc}
			\toprule \midrule
			& Env. & Object & \multicolumn{5}{c}{Indoor} & \multicolumn{6}{c}{Outdoor} \\  \cmidrule(lr){3-3} \cmidrule(lr){4-8} \cmidrule(lr){9-14}
			& Dataset & \ModelNet &  \ThreeDMatch &\ThreeDLoMatch & \ScanNetppi & \ScanNetppF & \TIERS & \KITTI & \WOD & \KAIST &  \MIT & \ETH & \Oxford \\ \midrule
            \parbox[t]{5mm}{\multirow{3}{*}
            {\rotatebox[origin=c]{90}{\begin{tabular}{@{}c@{}}Conven- \\tional \end{tabular}}}}
      & FGR~\cite{Zhou16eccv-FastGlobalRegistration} \ora & \color{gray} 84.04  & \color{gray} 62.53 & \color{gray} 15.42 & \color{gray} 77.68 & \color{gray} 92.31 & \color{gray} 80.60 & \color{gray} 98.74 & \color{gray} \textbf{100.00} & \color{gray} 89.80 & \color{gray} 74.78 & \color{gray} 91.87 & \color{gray} 99.00  \\
       & Quatro~\cite{Lim22icra-Quatro} \ora & \color{gray} 5.45 & \color{gray} 8.22 & \color{gray} 1.74 & \color{gray} 9.88 & \color{gray} 97.27 & \color{gray} 86.57 & \color{gray} 99.10 & \color{gray} \textbf{100.00} &\color{gray} 91.46 & \color{gray} 79.57 & \color{gray} 51.05 & \color{gray} 91.03  \\
       & TEASER++~\cite{Yang20tro-teaser}  \ora & \color{gray} 86.10 & \color{gray} 52.00 & \color{gray} 13.25 & \color{gray} 66.15 & \color{gray}97.22 & \color{gray}73.13 & \color{gray}98.92 & \color{gray}\textbf{100.00} & \color{gray} 89.20 & \color{gray} 71.30 & \color{gray} 93.69 & \color{gray} 99.34 \\ \midrule
       \color{black}
       \parbox[t]{2mm}{\multirow{17}{*}{\rotatebox[origin=c]{90}{\begin{tabular}{@{}c@{}}Trained solely on \ThreeDMatch\\(acquired with a RGB-D sensor) \end{tabular}}}}
       & FCGF~\cite{Choi19iccv-FCGF} & 7.35 & \cg 88.18 & \cg 40.09 & 72.90 &  88.69 & 55.96 & 0.00 & 0.00 & 0.00 & 0.00 & 54.98 & 0.00 \\
       & \ora & 16.51 & \cg 88.18 & \cg 40.09 & 85.87 & 88.69 & 78.62 & 90.27 & 97.69 & 92.91 & 92.61 & 54.98 & 93.68 \\
       & \osa & 16.51 & \cg 88.18 & \cg 40.09 & 85.87 & 88.69 & 80.11 & 94.41 & 97.69 & 93.55 & 93.04 & 55.53 & 95.68 \\
       & Predator~\cite{Huang21cvpr-PREDATORRegistration} & N/A & \cg 90.60 & \cg 62.40 & 75.94 & OOM & OOM & OOM & OOM & OOM & OOM &OOM & OOM \\
       &  \ora & 84.28 & \cg 90.60 & \cg 62.40 & 75.94 & 29.81 & 56.44 & 0.00 & 0.00 & 0.95 & 0.00 & 0.14 & 0.33 \\
       &  \osa & 84.28 & \cg 90.60 & \cg 62.40 & 75.94 & 86.01 & 75.74 & 77.29 & 86.92 & 87.09 & 79.56 & 54.42 & 93.68 \\
       & GeoTransformer~\cite{Qin23tpami-GeoTransformer} & 84.36& \cg 92.00 & \cg 75.00 & 91.18 & OOM & OOM & OOM & OOM & OOM & OOM & OOM & OOM  \\
       &  \ora & 86.26 & \cg 92.00 & \cg 75.00 & 91.18 & 7.54 & 5.06 & 0.36 & 0.77 & 0.25 & 0.87 & 0.00 & 0.33 \\
       &  \osa & 86.26 & \cg 92.00 & \cg 75.00 & 92.72  & \thirdc 97.02 & \secondc 92.99 & 92.43 & 89.23 & 91.86 & \secondc 95.65 & 71.53 & 97.01\\
       & BUFFER~\cite{Ao23CVPR-BUFFER} & 82.39 & \cg 92.90 & \cg 71.80 & 92.72 & 93.75 & 62.30 & 0.00 & 1.54 & 0.50 & 6.96 & 97.62 & 0.66\\
	&  \ora & \thirdc 92.42 & \cg 92.90 & \cg 71.80 & \thirdc 93.01 & 94.69 & 88.96 & \thirdc 99.46 & \firstc \textbf{100.00} & \thirdc 97.24 & \secondc 95.65 & \secondc 99.30 & 99.00  \\
       &  \osa & \thirdc 92.42 & \cg 92.90 & \cg 71.80 & \thirdc 93.01 & 94.69 & 88.96 & \thirdc  99.46 & \firstc \textbf{100.00} & \thirdc 97.24 & \secondc 95.65 & \secondc 99.30 & 99.00 \\
	& PARENet~\cite{Yao24iccv-PARENet} & N/A & \cg 95.00 & \cg \textbf{80.50} & 90.84 & OOM & OOM & OOM & OOM & OOM & OOM & OOM & OOM \\
      & \ora & 60.43 & \cg 95.00 & \cg \textbf{80.50} & 90.84 & 43.75 & 6.21 & 0.18 & 0.77 & 0.75 & 1.30 & 1.40 & 1.66 \\
      & \osa & 66.14 & \cg 95.00 & \cg \textbf{80.50} & 90.84 & 87.95 & 75.06  & 84.86 & 92.31 & 86.44 & 84.78 & 69.42 & 93.36 \\ \cmidrule(lr){2-14}
          & Our \oursname with only $r_m$ & \firstc \textbf{99.84} & \cg 93.38 & \cg 71.69 & \secondc 93.10 &\secondc 99.60 & \thirdc 90.80 & \firstc \textbf{99.82} & \firstc \textbf{100.00} & \secondc 99.05 & \secondc 95.65 & \secondc 99.30 & \secondc 99.34 \\ 
          & Our \oursname & \firstc \textbf{99.84} & \cg \textbf{95.58} & \cg 74.18 & \firstc \firstc \textbf{94.99} & \firstc \textbf{99.90} & \firstc \textbf{93.45} & \firstc \textbf{99.82} & \firstc \textbf{100.00} & \firstc \textbf{99.15} & \firstc \textbf{97.39} & \firstc \textbf{99.72} & \firstc \textbf{99.67} \\ \midrule 
    \parbox[t]{2mm}{\multirow{17}{*}{\rotatebox[origin=c]{90}{\begin{tabular}{@{}c@{}}Trained solely on \KITTI\\(acquired with a Velodyne HDL-64E LiDAR) \end{tabular}}}}
    & FCGF~\cite{Choi19iccv-FCGF}
       & 7.50 & 8.04 & 0.17 & 19.96 & 23.07 & 77.82 & \cg 98.92 & 95.38 & 88.34& 82.17 & 6.59 & 75.08 \\
       & \ora & 15.32 & 34.97 & 4.12 & 31.00 & 25.10 & 77.93 & \cg  98.92& 96.92 & 94.22 & 89.13 & 39.97 & 86.05 \\
       & \osu & 19.19 & 34.97 & 4.12 & 33.37 &25.10 & 77.93 & \cg 98.92 & 99.23 &94.22 & 90.43 & 39.97 & 90.03 \\
       & Predator~\cite{Huang21cvpr-PREDATORRegistration} & N/A & N/A & N/A & N/A & N/A & 69.43 & \cg 99.82 & \firstc \textbf{100.00} & OOM & 54.78 & 55.68 & 89.04 \\
       &  \ora & 8.60 & 16.47 & 0.00 & 9.40 & 3.72 &  69.77 & \cg \textbf{99.82} & \firstc \textbf{100.00}  & 71.3 & 76.52 & 56.67 & 89.04 \\
       &  \osu & 8.60 & 23.2 & 3.31 & 9.40 & 3.72 & 69.77 & \cg \textbf{99.82} & \firstc \textbf{100.00} & 94.02 & 86.08 & 71.95 & 95.02  \\
       & GeoTransformer~\cite{Qin23tpami-GeoTransformer}  & N/A & N/A & N/A & N/A & N/A & N/A & \cg \textbf{99.82} & \firstc \textbf{100.00} & 63.84 & 93.91 & 77.00 & 73.42 \\
       &  \ora & 74.25 & 5.94 & 0.30 & 15.91 & 34.18 & 20.57 & \cg \textbf{99.82} & \firstc \textbf{100.00} & 63.84 & 93.91 & 77.56 & 73.42 \\
       &  \osu & 74.25 & 62.17 & 14.38 & 76.52 & 90.63 & 87.36 & \cg \textbf{99.82} & \firstc \textbf{100.00} & 96.84 & 96.52 & 81.77 & 98.01 \\
       & BUFFER~\cite{Ao23CVPR-BUFFER} & N/A & N/A & N/A & 17.60 & 88.84 & 93.34& \cg 99.64 & \firstc \textbf{100.00} & 99.50 & 95.22 & 98.18 & \thirdc 99.34  \\
       &  \ora & \thirdc 95.41 & \thirdc 91.19 & \thirdc 64.51 & \thirdc 93.15 & \thirdc 97.81 & \thirdc 93.57 & \cg 99.64 & \firstc \textbf{100.00} & \firstc \textbf{99.55} & \thirdc 97.39 & \thirdc 99.86 & \thirdc 99.34 \\
       &  \osu & \thirdc 95.41 & \thirdc 91.19 & \thirdc 64.51 & \thirdc 93.15 & \thirdc 97.81 & \thirdc 93.57 & \cg 99.64 & \firstc \textbf{100.00} & \firstc \textbf{99.55} & \thirdc 97.39 & \thirdc 99.86 & \thirdc 99.34 \\
       & PARENet~\cite{Yao24iccv-PARENet} & N/A & N/A & N/A & N/A & N/A & N/A & \cg \textbf{99.82} & 97.69 & 57.51 & 75.22 & 68.30 & 66.11 \\
  & \ora & 34.91 & 0.77 & 0.10 & 3.04 & 12.00 & 19.20 & \cg \textbf{99.82} & 98.46 & 57.51 & 75.22 & 68.44 & 66.11 \\
  & \osu & 83.18 & 22.09 & 4.98 & 29.99 & 42.91 & 52.99 & \cg \textbf{99.82} & \firstc \textbf{100.00} & 89.50 & 87.39 & 72.65 & 94.02 \\ 
      \cmidrule(lr){2-14}
  & Our \oursname with only $r_m$ & \firstc \textbf{99.84} & \secondc 91.96 & 63.59 & 92.38 & \secondc 99.45 & \secondc 94.37 & \cg \textbf{99.82} & \firstc \textbf{100.00} &\firstc \textbf{99.55}&\firstc \textbf{99.13}&\firstc \textbf{100.00} & \firstc \textbf{99.67} \\ 
  & Our \oursname & \firstc \textbf{99.84} & \firstc \textbf{93.79} & \firstc \textbf{65.89} & \firstc \textbf{95.13} & \firstc \textbf{99.65} & \firstc \textbf{94.83} & \cg \textbf{99.82} & \firstc \textbf{100.00} & \firstc \textbf{99.55} & \firstc \textbf{99.13} & \firstc \textbf{100.00} & \firstc \textbf{99.67} \\ \midrule \bottomrule
		\end{tabular}
    }
    \label{table:success_rates}
   	\begin{flushleft}
      {
      \scriptsize
      \oracle: Oracle tuning with manually optimized voxel size and search radius for each dataset. \\
      \scalealign, \scaleup: Scale down and up alignment, respectively, to normalize dataset scales (\eg the scale of \KITTI, which typically uses a voxel size of 0.3\,m, is adjusted to match the scale of \ThreeDMatch, where 0.025\,m is commonly used, by dividing by $\frac{0.3}{0.025}$). \\
      N/A: Failure due to too few points remaining after voxelization with the voxel size typically used for larger scale scenes. \\ %
      \vspace{-1.5mm}
      OOM: Out-of-memory error caused by excessive memory usage.
      }
	  \end{flushleft}
	  \label{fig:kitti_odom}
      \vspace{-3mm}
\end{table*}
\endgroup


\newcommand{\qualW}{0.165}
\begin{figure*}[t!]
  \centering
  \subfloat[Source \& Target]{
    \begin{minipage}[b]{\qualW\textwidth}
        \includegraphics[width=\linewidth]{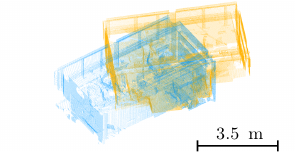}
        \includegraphics[width=\linewidth]{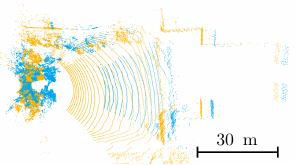}
        \includegraphics[width=\linewidth]{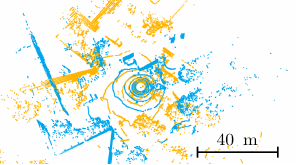}
        \includegraphics[width=\linewidth]{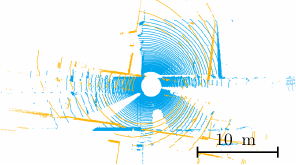}
        \includegraphics[width=\linewidth]{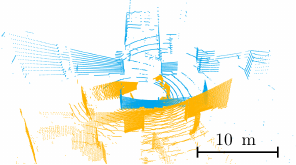}
        \includegraphics[width=\linewidth]{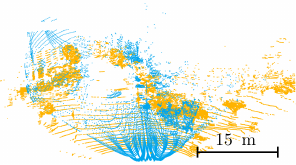}
        \includegraphics[width=\linewidth]{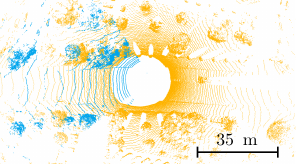}
    \end{minipage}
   }
  \subfloat[GeoTransformer]{
    \begin{minipage}[b]{\qualW\textwidth}
        \includegraphics[width=\linewidth]{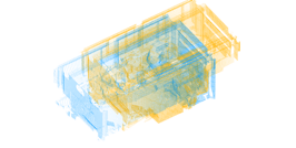}
        \includegraphics[width=\linewidth]{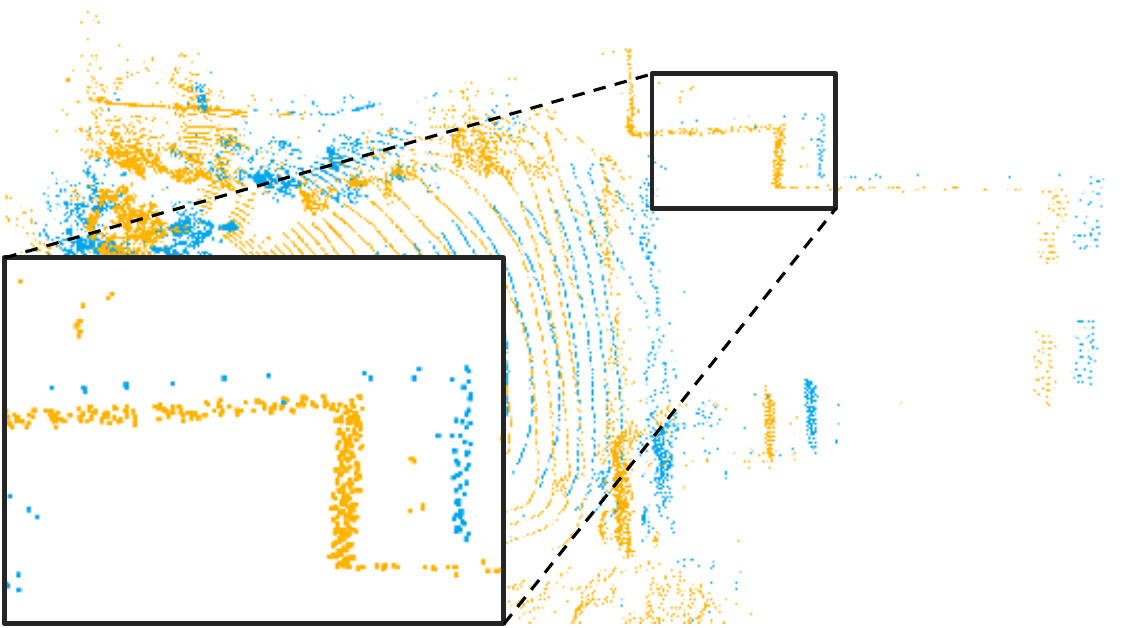}
        \includegraphics[width=\linewidth]{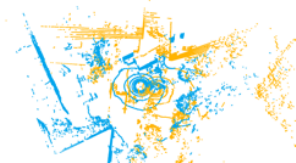}
        \includegraphics[width=\linewidth]{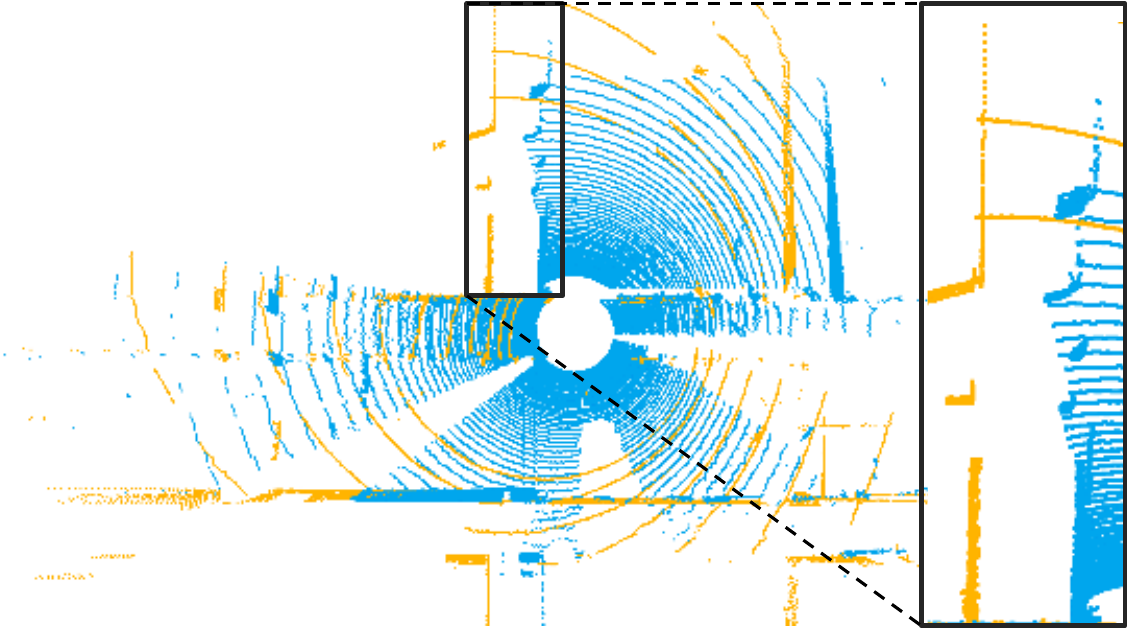}
        \includegraphics[width=\linewidth]{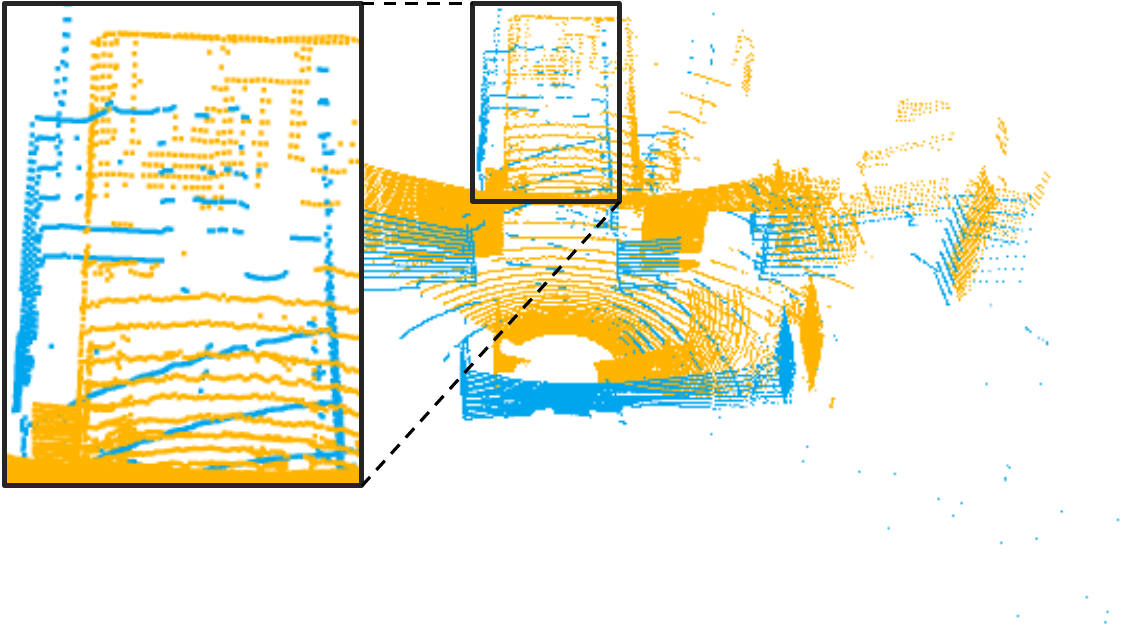}
        \includegraphics[width=\linewidth]{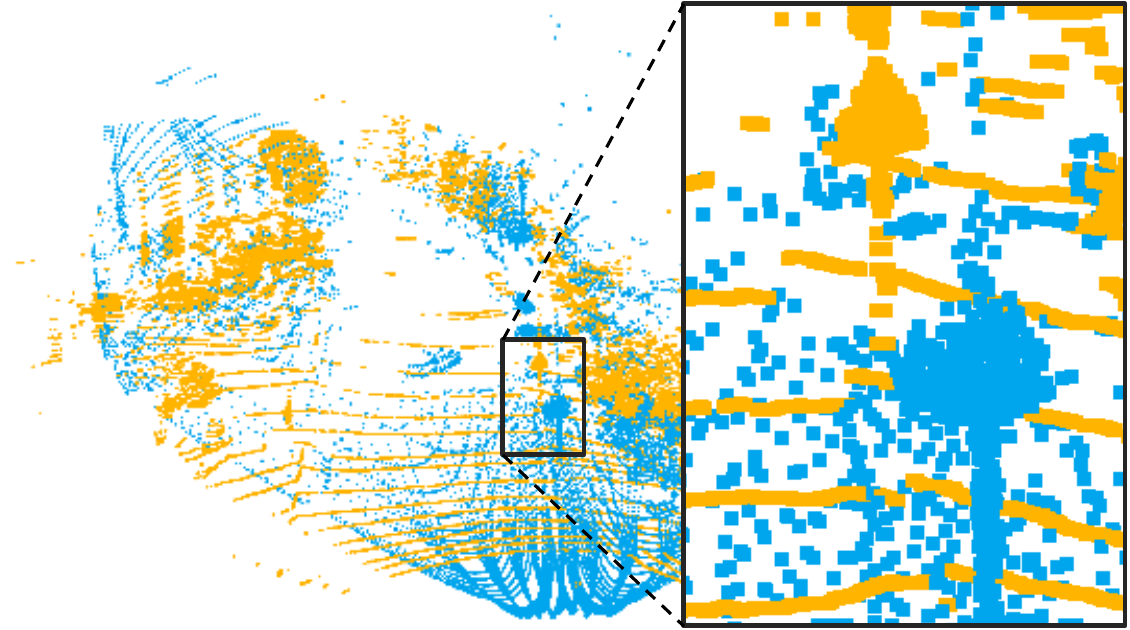}
        \includegraphics[width=\linewidth]{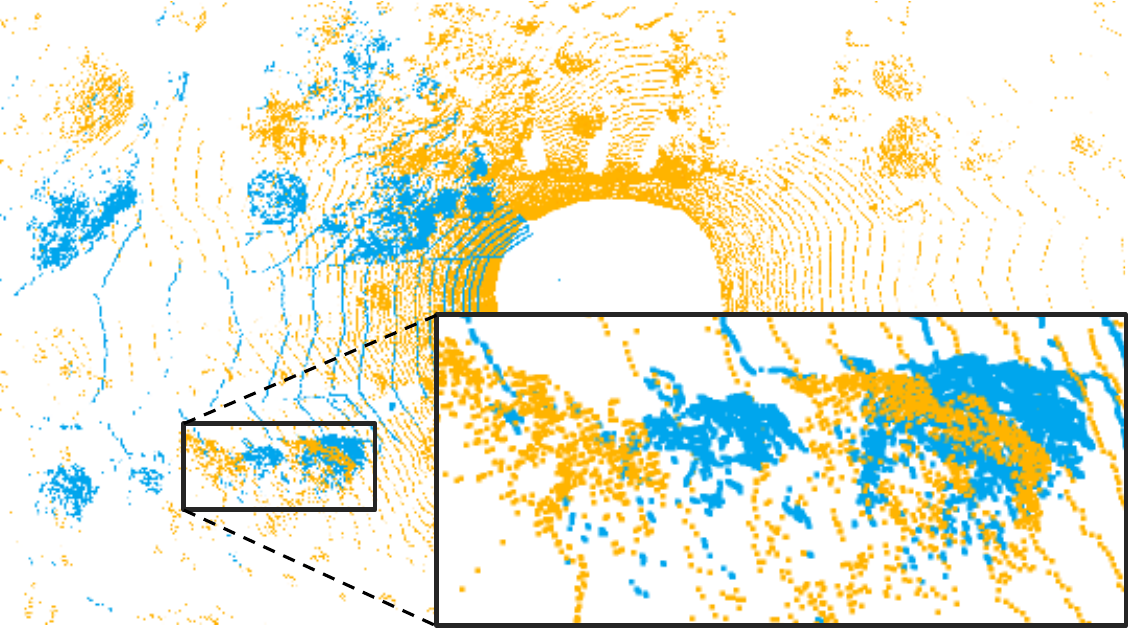}
    \end{minipage}
  }
  \subfloat[BUFFER]{
    \begin{minipage}[b]{\qualW\textwidth}
        \includegraphics[width=\linewidth]{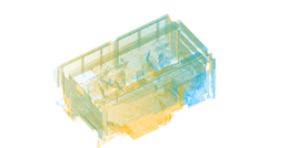}
        \includegraphics[width=\linewidth]{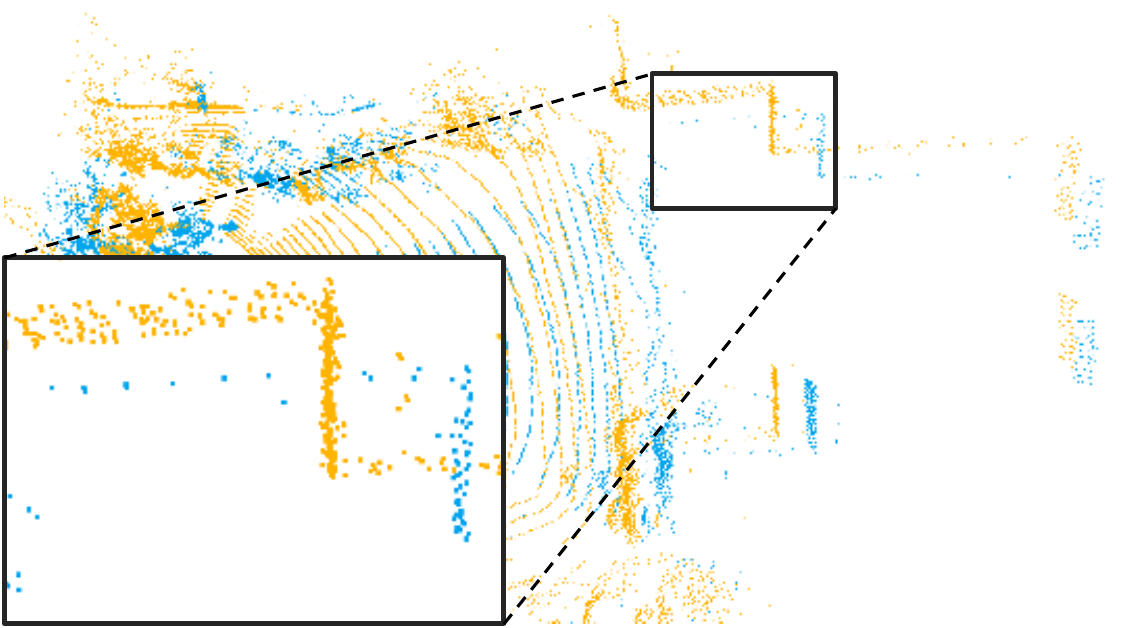}
        \includegraphics[width=\linewidth]{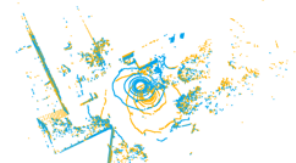}
        \includegraphics[width=\linewidth]{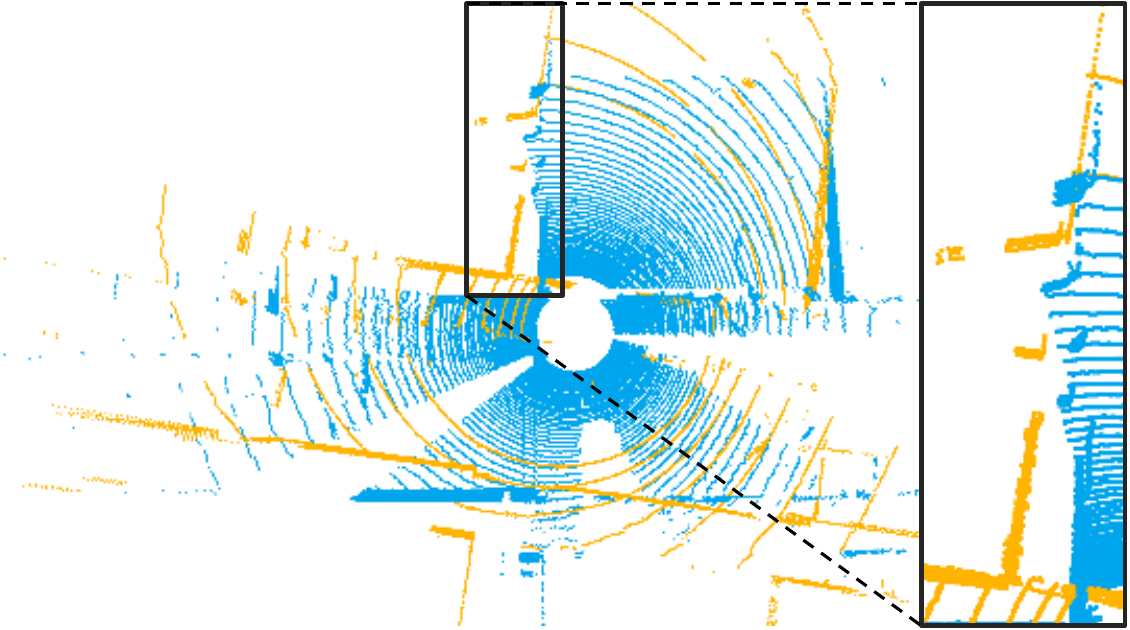}
        \includegraphics[width=\linewidth]{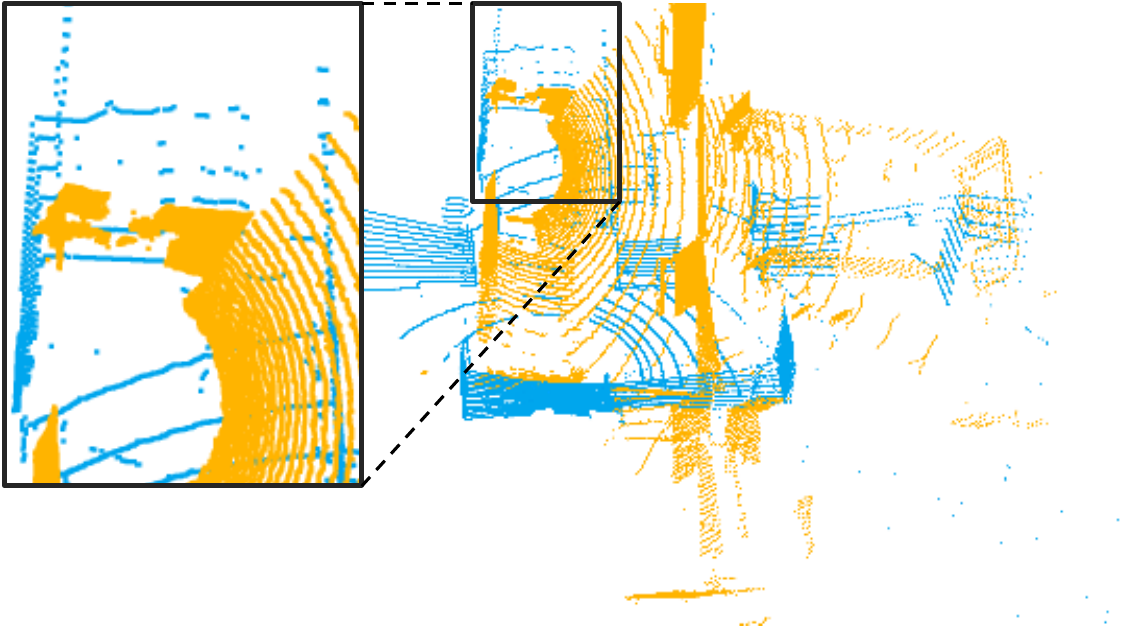}
        \includegraphics[width=\linewidth]{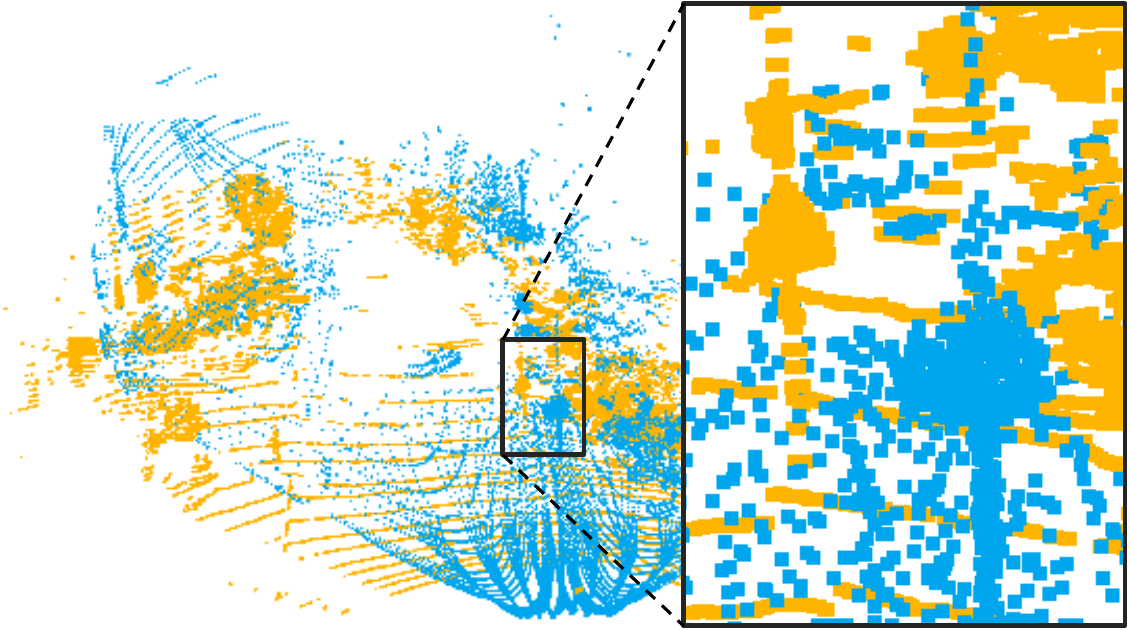}
        \includegraphics[width=\linewidth]{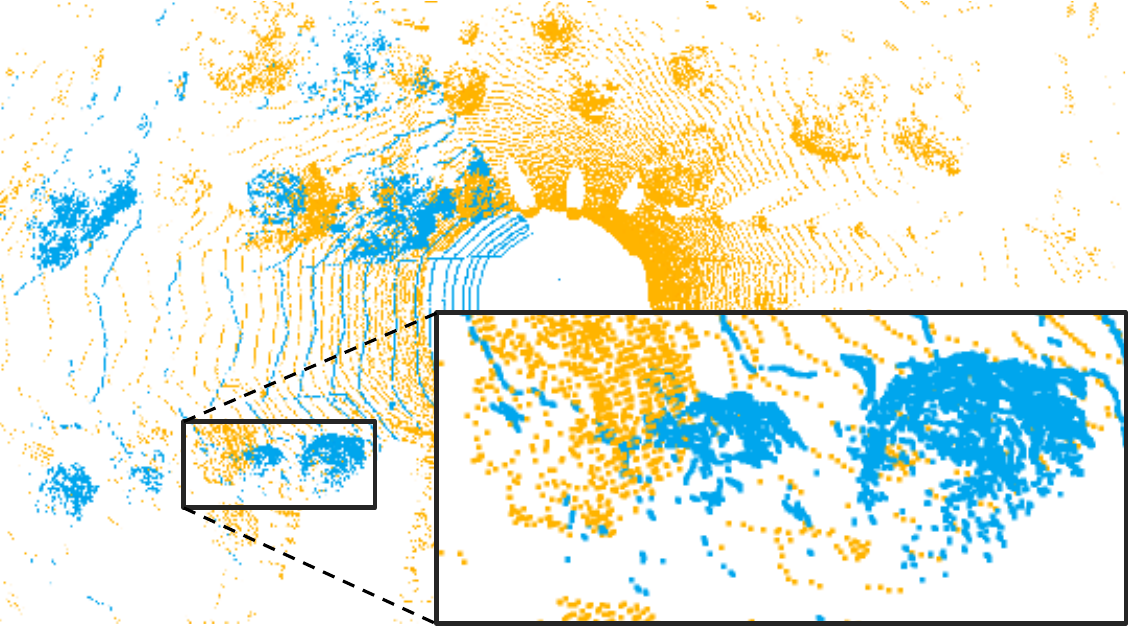}
    \end{minipage}
  }
  \subfloat[Our \oursname]{
    \begin{minipage}[b]{\qualW\textwidth}
        \includegraphics[width=\linewidth]{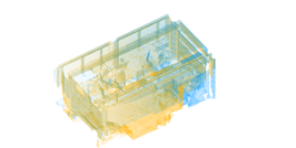}
        \includegraphics[width=\linewidth]{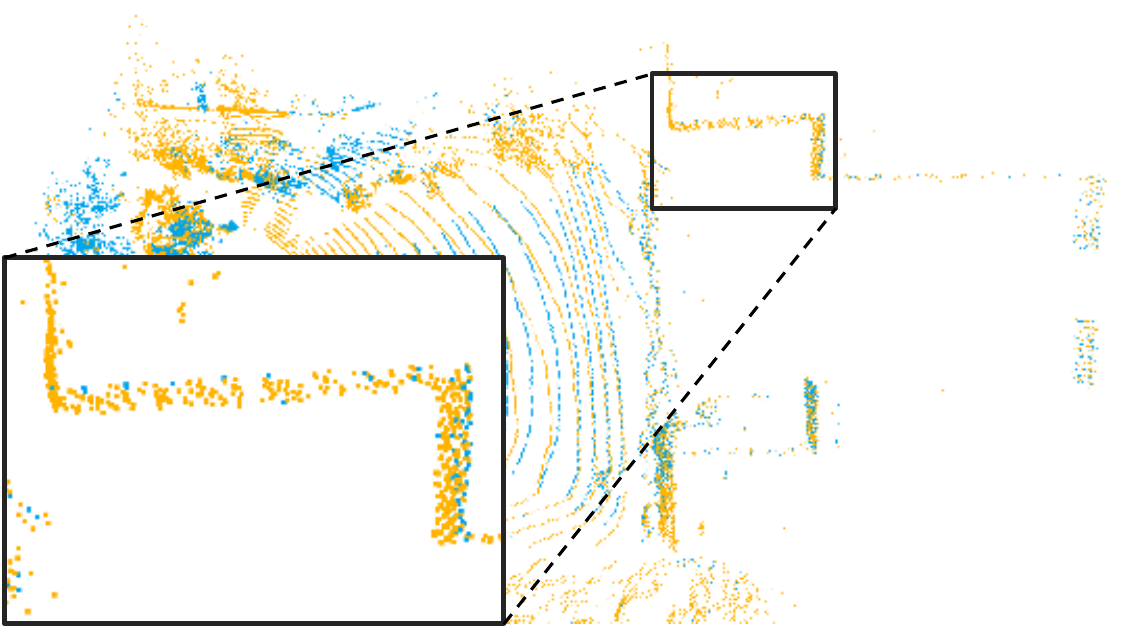}
        \includegraphics[width=\linewidth]{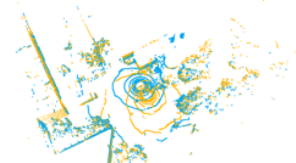}
        \includegraphics[width=\linewidth]{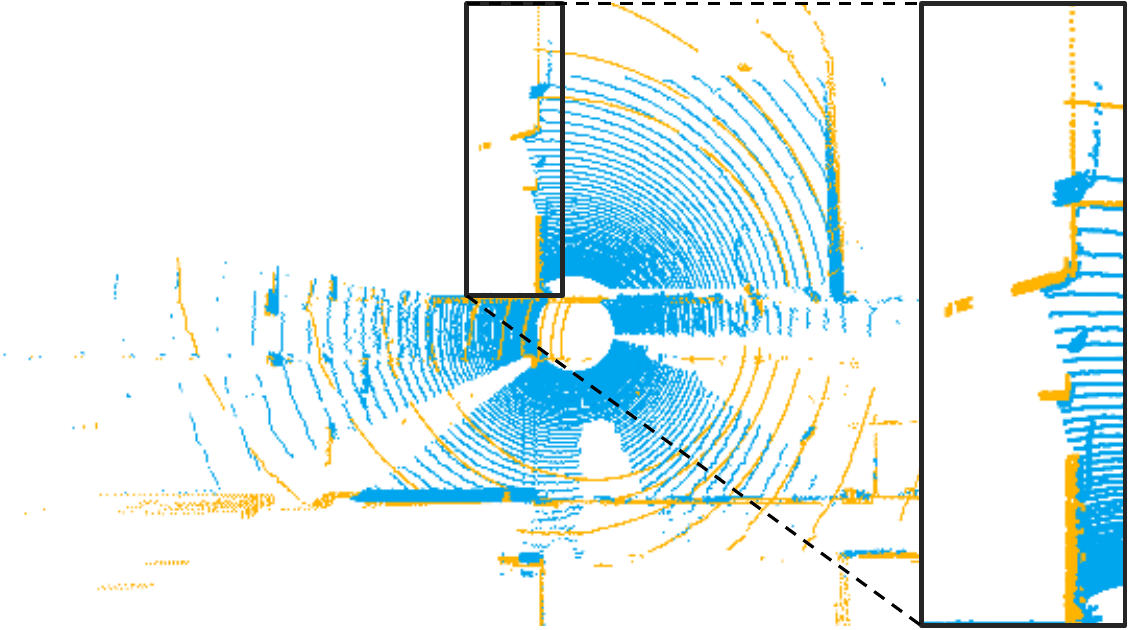}
        \includegraphics[width=\linewidth]{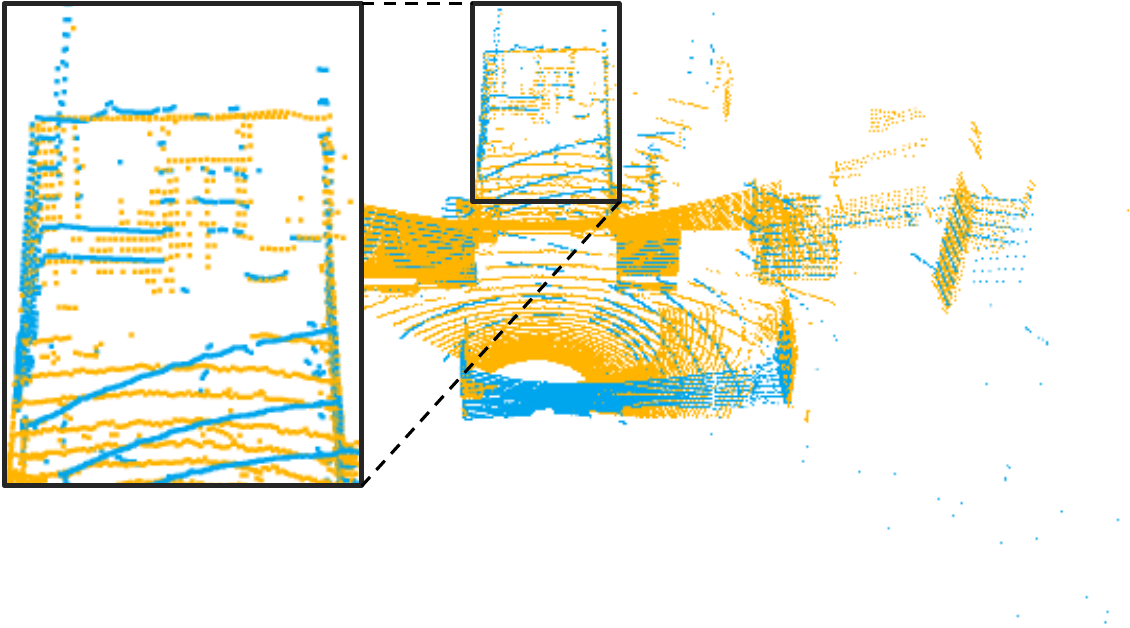}
        \includegraphics[width=\linewidth]{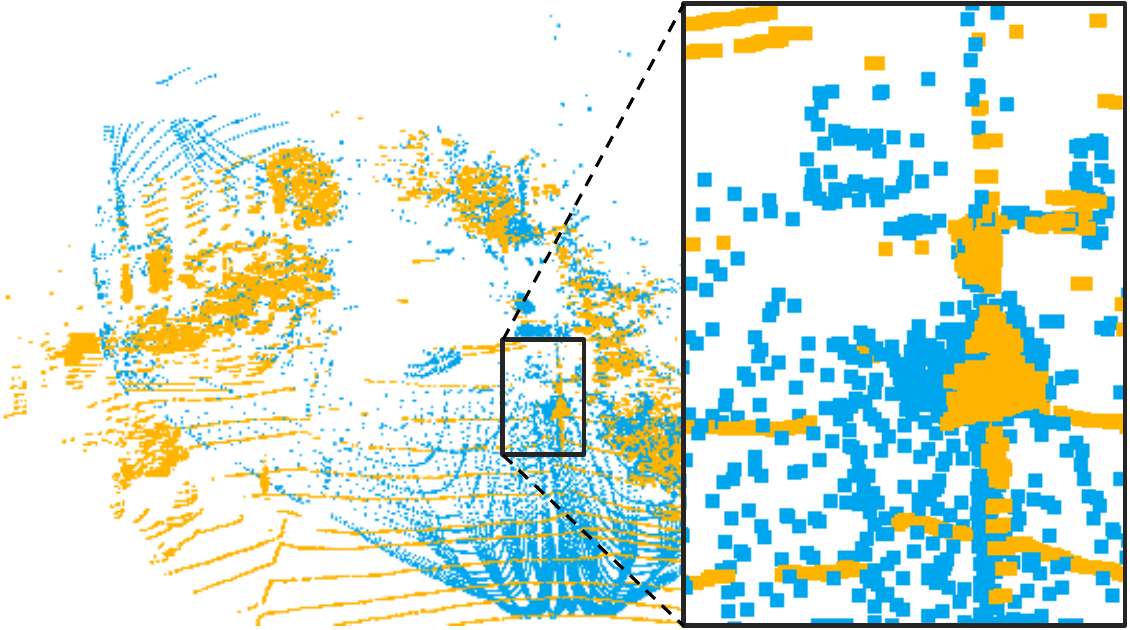}
        \includegraphics[width=\linewidth]{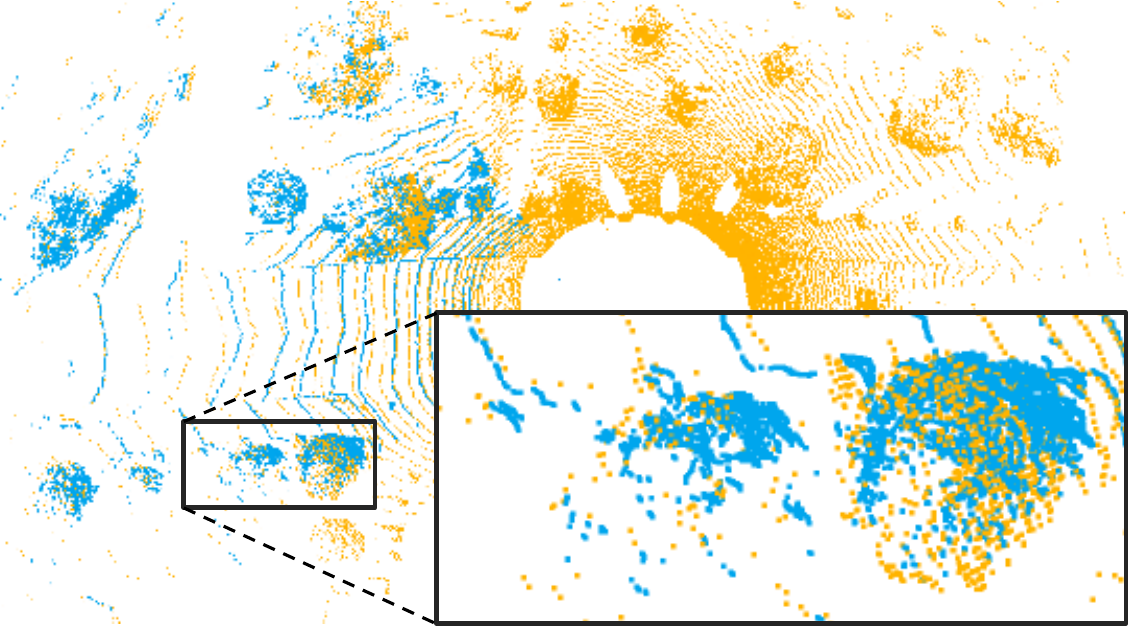}
    \end{minipage}
  }
  \subfloat[GT]{
    \begin{minipage}[b]{\qualW\textwidth}
        \includegraphics[width=\linewidth]{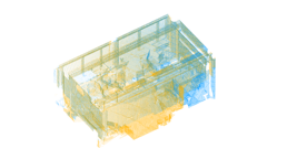}
        \includegraphics[width=\linewidth]{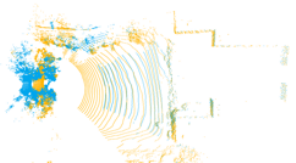}
        \includegraphics[width=\linewidth]{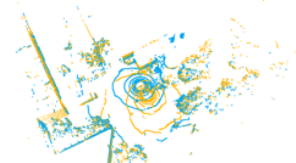}
        \includegraphics[width=\linewidth]{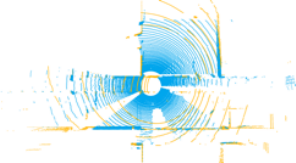}
        \includegraphics[width=\linewidth]{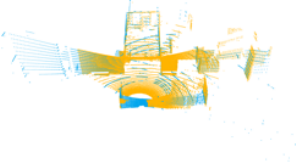}
        \includegraphics[width=\linewidth]{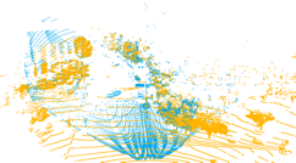}
        \includegraphics[width=\linewidth]{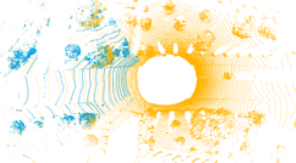}
    \end{minipage}
  }
  \caption{Qualitative comparison of registration results across diverse scenes (source: orange and target: cyan).  All methods were trained on \ThreeDMatch.
  Top three rows: scan pairs from \ScanNetppF, \Oxford, and \MIT.
  Bottom four rows: scan pairs with heterogeneous LiDAR sensor setups from \TIERS~(Vel16$\rightarrow$OS128, OS64$\rightarrow$Vel16) and \KAIST~(Aeva$\rightarrow$Avia, Ouster$\rightarrow$Aeva), respectively, where A$\rightarrow$B means that the source acquired by sensor type A is registered to the target point cloud from sensor type B.}
  \label{fig:qualitative_comparison}  
  \vspace{-3mm}
\end{figure*}

\section{Experimental Results and Discussion}

\subsection{Why existing methods fail to generalize?}\label{sec:analyses}

First, we demonstrate that existing methods struggle in achieving out-of-the-box generalization, leading to performance degradation due to the issues explained in \Cref{sec:our_preliminaries}.
In this experiment, we mainly used renowned learning-based approaches: FCGF~\cite{Choi19iccv-FCGF}, Predator~\cite{Huang21cvpr-PREDATORRegistration}, GeoTransformer~(\textit{GeoT} for brevity)~\cite{Qin23tpami-GeoTransformer}, BUFFER~\cite{Ao23CVPR-BUFFER}, and PARENet~\cite{Yao24iccv-PARENet}.

As shown in \Cref{table:success_rates}, models trained with hyperparameters optimized for one dataset scale~(\eg voxel size and search radius) exhibited substantial performance degradation when applied to datasets with different scales.
More critically, as explained in \Cref{sec:prob_user_defined}, we observed two types of failures when applying methods with mismatched hyperparameters.
First, when using indoor parameters~(small voxel sizes) on outdoor datasets, some methods encountered out-of-memory~(OOM) errors due to the excessively large number of input points remaining after voxelization.
Conversely, when applying outdoor parameters~(large voxel sizes) to \ModelNet and indoor datasets, methods failed due to too few points remaining after voxelization~(marked as N/A in \Cref{fig:kitti_odom}).

Once a properly user-tuned voxel size and search radius were provided~(referred to as \textit{oracle tuning},  \raisebox{-0.3ex}{\includegraphics[width=0.30cm]{pics/symbols_oracle.pdf}}), BUFFER showed remarkable performance improvement owing to its patch-wise input scale normalization characteristics.
In contrast, other approaches still showed relatively lower success rates 
because the networks received point clouds with magnitudes not encountered during training.
This potential limitation is further evidenced by the performance improvement of Predator and GeoT after scale alignment~(denoted by \raisebox{-0.3ex}{\includegraphics[width=0.30cm]{pics/symbols_scale_align.pdf}}
and \raisebox{-0.3ex}{\includegraphics[width=0.30cm]{pics/symbols_scale_up_v2.pdf}}), 
supporting our claim that scale normalization is a key factor in achieving generalizability.

\begin{figure}[t!]
    \centering
    \includegraphics[width=1\columnwidth]{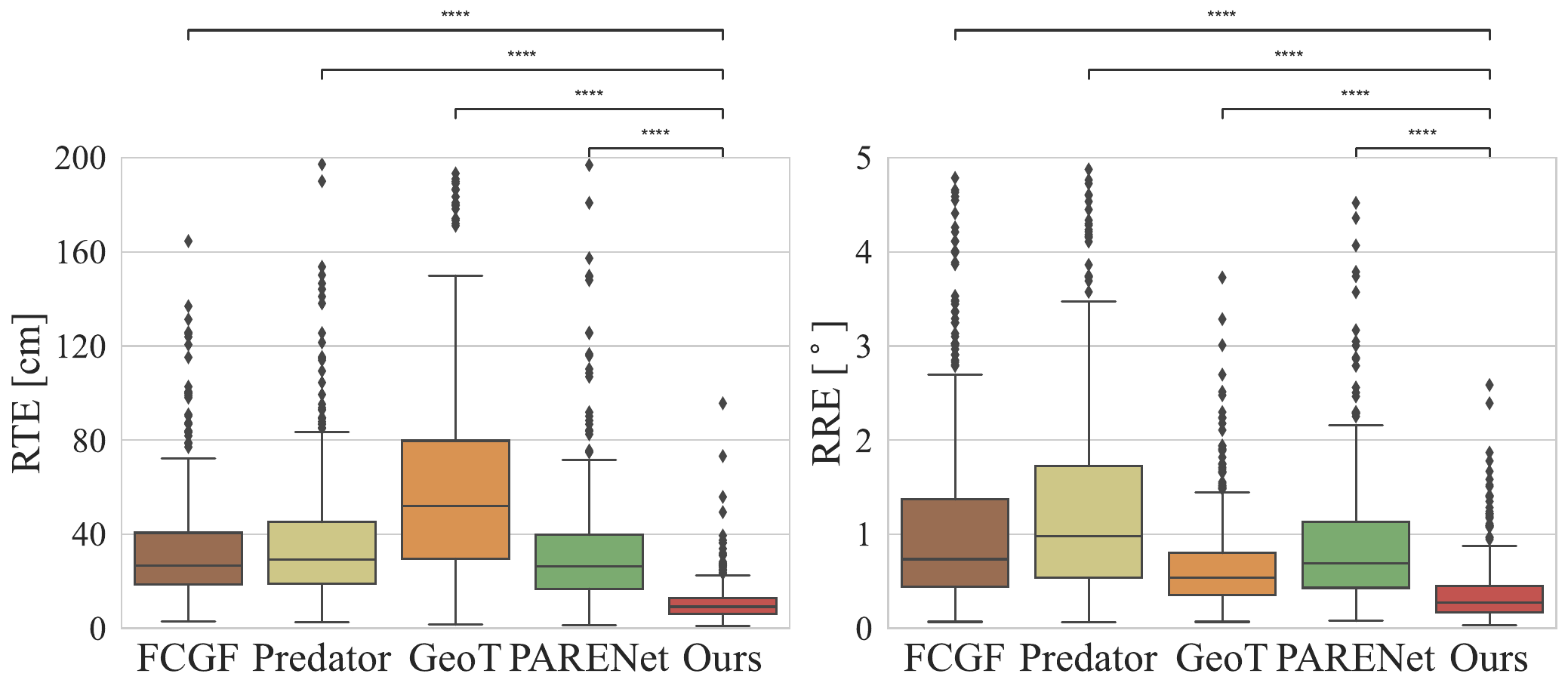}
    \caption{Relative translation error~(RTE) and relative rotation error~(RRE) of our approach to state-of-the-art methods, all trained on {\ThreeDMatch} and tested on \KITTI, with oracle tuning and scale alignment, corresponding to those in \Cref{table:success_rates} under the + \raisebox{-0.4ex}{\includegraphics[width=0.30cm]{pics/symbols_oracle.pdf}} + \raisebox{-0.4ex}{\includegraphics[width=0.30cm]{pics/symbols_scale_align.pdf}} setting.
    The **** annotations indicate measurements with a $p$-value $< 10^{-4}$ after a paired $t$-test.
    Note that even if other state-of-the-art approaches showed promising high success rates in \Cref{table:success_rates}, but in terms of RTE and RRE, our \oursname showed substantially lower values.}
    \label{fig:poor_alignment}
\end{figure}

\subsection{Out-of-the-box zero-shot registration performance}\label{sec:sota_comparison}

Leveraging these insights, our method demonstrates strong zero-shot generalization across diverse datasets without requiring manual hyperparameter tuning.
As presented in \Cref{table:success_rates} and \Cref{fig:qualitative_comparison}, unlike other approaches, 
our approach demonstrates strong generalization across diverse domains~(\eg object-scale, indoor, and outdoor environments) without requiring oracle tuning or scale alignment.
Moreover, even in terms of RTE and RRE, our approach achieves lower errors compared to state-of-the-art methods that use oracle tuning and scale alignment~(\Cref{fig:poor_alignment}).
These results validate that our algorithm generalizes well across domains while preserving accuracy on in-domain scenarios.

To further evaluate the generalization capability of our approach, we analyze the performance across heterogeneous LiDAR sensor configurations with point clouds captured by different sensors within the same environment.
This is a more challenging scenario that requires robustness to varying sensor characteristics such as different scanning patterns, point densities, and field of view; see Figs.~\ref{fig:data_graphs}(f), \ref{fig:data_graphs}(g), and \ref{fig:campus_scale}(c).

\begingroup
\begin{table*}[t!]
    \caption{Registration success rates under heterogeneous sensor configurations. Deep learning-based methods are trained only on \ThreeDMatch~\cite{Zeng17cvpr-3dmatch}. Conventional methods use FPFH~\cite{Rusu09icra-fast3Dkeypoints}. To ensure a fair comparison, RANSAC was employed for pose estimation across all learning-based approaches, with a maximum of 50K iterations.}
    \label{table:hetero_success_rates}
    \setlength{\tabcolsep}{6pt}
    \centering
    {\scriptsize
    \begin{tabular}{l|l|ccc|ccc|c}
        \toprule \midrule
        & Dataset & \multicolumn{3}{c}{\TIERS} & \multicolumn{3}{c}{\KAIST} & \multirow{2}{*}{\begin{tabular}{@{}c@{}}Average \\ rank\end{tabular}} \\
        \cmidrule(lr){3-5} \cmidrule(lr){6-8}
        & Source $\rightarrow$ Target & OS128 $\rightarrow$ OS64 & OS64 $\rightarrow$ Vel16 & Vel16 $\rightarrow$ OS128 & Aeva $\rightarrow$ Avia & Avia $\rightarrow$ Ouster & Ouster $\rightarrow$ Aeva
        & \\ \midrule
        \parbox[t]{5mm}{\multirow{3}{*}
        {\rotatebox[origin=c]{90}{\begin{tabular}{@{}c@{}}Conven- \\tional \end{tabular}}}}
        &  FGR~\cite{Zhou16eccv-FastGlobalRegistration} \ora & 61.57 & 71.49 & 32.57 & 39.27 & 48.79 & 72.10 &  7.67 \\
        & Quatro~\cite{Lim22icra-Quatro} \ora & 75.78 & \thirdc 90.06 & 48.08 & 51.67 & 51.90 & 76.51  & 5.67 \\
        & TEASER++~\cite{Yang20tro-teaser} \ora & 62.11 & 64.76 & 27.96 & 28.93 & 38.28 & 63.29  & 8.50 \\ \midrule
        \parbox[t]{2mm}{\multirow{17}{*}{\rotatebox[origin=c]{90}{\begin{tabular}{@{}c@{}}Deep learning-based\\ (Trained solely on \ThreeDMatch) \end{tabular}}}}
        & FCGF~\cite{Choi19iccv-FCGF} & 0.00 & 0.00 & 0.00 & 0.00 & 0.00 & 0.00 & 13.83 \\
        & \ora & 28.95 & 30.05 & 29.03 & 27.34 & 22.58 & 12.04 & 9.50 \\
        & \osa & 28.95 & 30.05 & 29.03 & 27.34  & 22.58 & 12.04 & 9.50 \\
        & Predator~\cite{Huang21cvpr-PREDATORRegistration} & N/A & N/A  & N/A & N/A & N/A & N/A & 13.50 \\
        & \ora & 1.62 & 0.00 & 0.00 & 0.00 & 0.00 & 0.00 & 13.17 \\
        & \osa & 25.03 & 18.38 & 10.71 &  20.19  & 14.48 & 21.59 & 10.83 \\
        & GeoTransformer~\cite{Qin23tpami-GeoTransformer} & N/A & N/A & N/A & N/A  & N/A  & N/A  & 13.17 \\
        & \ora & 0.27 & 0.22 & 0.05 & 0.00 & 0.00 & 0.00 & 12.67 \\
        & \osa & 85.25 & 83.24 & 55.90 & 76.79 & 73.79 & 56.97 & 5.67 \\
        & BUFFER~\cite{Ao23CVPR-BUFFER} & 7.98 & 3.89 & 2.48 & 0.00 & 0.00 & 0.15 & 11.83 \\
        & \ora & \thirdc 92.83 & 88.54 & \thirdc 66.77 & \thirdc 91.57 & \thirdc 90.52 & \thirdc 97.06 & \thirdc 3.17 \\
        & \osa & \thirdc 92.83 & 88.54 & \thirdc 66.77 & \thirdc 91.57 & \thirdc 90.52 & \thirdc 97.06 & \thirdc 3.17 \\
        & PARENet~\cite{Yao24iccv-PARENet} & N/A & N/A & N/A  & N/A & N/A & N/A  & 12.33 \\
        & \ora & 0.00 & 0.00 & 0.00 & 0.00 & 0.00 & 0.00
        &  12.33 \\
        & \osa & 65.90 & 59.78 & 37.42 & 75.84 & 70.52 & 53.60 & 6.67 \\ \cmidrule(lr){2-9}
        & Our \oursname with only $r_m$ & \secondc 95.40 & \secondc 96.86 & \secondc 77.33 & \secondc 96.04 & \secondc 93.80 & \secondc 97.20 & \secondc 2.00 \\
        & Our \oursname & \firstc \textbf{98.92} & \firstc \textbf{99.46} & \firstc \textbf{84.32} & \firstc \textbf{97.62} & \firstc \textbf{96.03} & \firstc \textbf{98.24} & \firstc \textbf{1.00} \\ \midrule \bottomrule
    \end{tabular}
    }
   	\begin{flushleft}
      {
      \scriptsize
      \hspace{4mm} \oracle: Oracle tuning with manually optimized voxel size and search radius for each dataset. \\
      \hspace{4mm} \scalealign: Scale down alignment to normalize dataset scales. \\
      \hspace{4mm} N/A: Failure due to too few points remaining after voxelization with the voxel size typically used for larger scale scenes. \\ %
      }
	  \end{flushleft}
	  \label{fig:hetero_marks}
    \vspace{-0.7cm}
\end{table*}
\endgroup

\begin{figure}[t!]
    \centering
    \includegraphics[width=1\columnwidth]{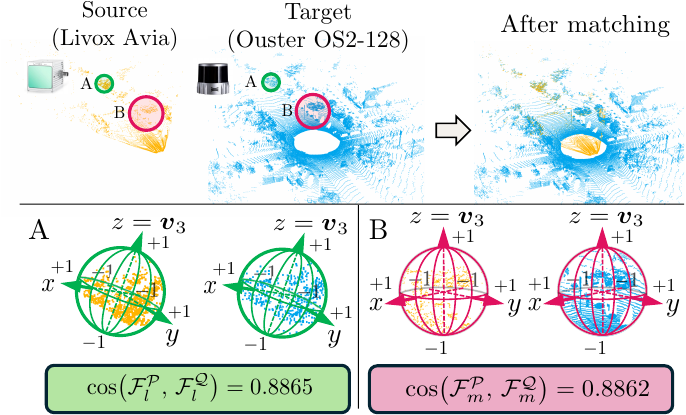}
    \caption{Visualization of patch-level similarity across heterogeneous LiDAR sensors.
Despite large differences in overall point density distributions and fields
of view captured by Livox Avia and Ouster OS2-128,
local patches from different point clouds (A and B) exhibit consistent feature distributions, showing high cosine similarity close to 1.}
    \label{fig:patch_similarity}
\end{figure}

As shown in \Cref{table:hetero_success_rates} and the bottom four rows of \Cref{fig:qualitative_comparison}, our approach demonstrates strong sensor-agnostic generalization capability.
First, most existing deep learning methods completely fail without oracle tuning~(N/A or 0.00\%).  
Even with oracle tuning and scale alignment, state-of-the-art methods show limited performance, highlighting severe brittleness to sensor variations.
In contrast, our approach achieves the highest success rates across all six sensor pairs without any manual tuning, even for challenging cases such as very sparse point clouds to dense clouds~(\ie Vel16 $\rightarrow$ OS128),
showing lower performance degradation compared to other approaches.
Furthermore, we analyzed the rationale at the patch-level. As illustrated in \Cref{fig:patch_similarity}, although density distributions and fields of view differ across sensors, the patch-level features extracted by our method remain consistent.

Therefore, this experimental evidence validates that our geometric bootstrapping, patch-wise normalization, and multi-scale matching strategy effectively handles the variations in point cloud characteristics introduced by different sensor types, making our method sensor-agnostic.

\begin{figure*}[t!]
  \centering
  \includegraphics[width=0.92\textwidth]{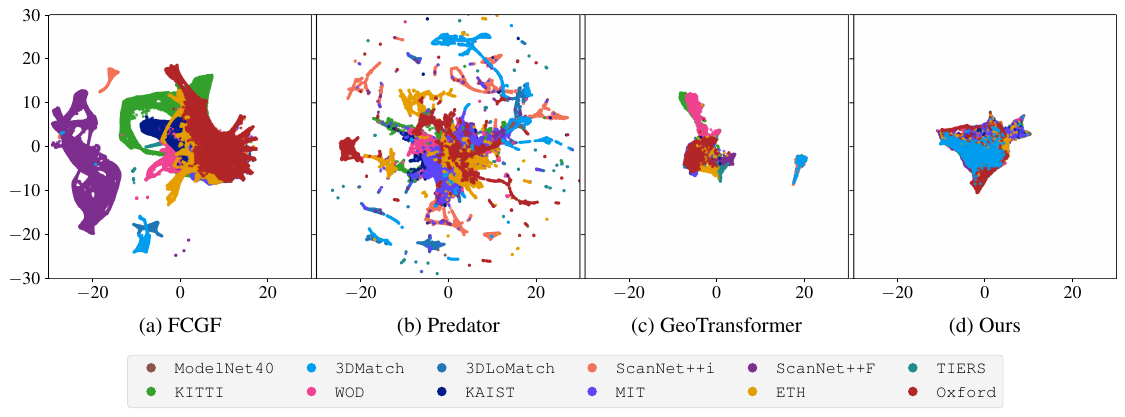}
  \caption{PaCMAP visualization of descriptor distributions across datasets. Unlike state-of-the-art approaches~\cite{Choi19iccv-FCGF,Huang21cvpr-PREDATORRegistration,Qin23tpami-GeoTransformer}, our \oursname shows persistent descriptors that are robust to domain shift, as evidenced by the overlap between \ThreeDMatch~(training domain) and other datasets. That is, the closer and more overlapped with sky blue, the better.}
  \label{fig:PaCMAP}
\end{figure*}

\subsection{In-depth analysis of key components}\label{sec:indepth}

\noindent\textbf{Point normalization for domain-invariant features.}
To understand why our approach achieves superior generalization, we first analyze the feature distributions using PaCMAP visualization~\cite{Wang21jmlr-PaCMAP}, which preserves relative distances between data points better than t-SNE~\cite{Maaten08jmlr-tSNE}.
As presented in \Cref{fig:PaCMAP}, unlike state-of-the-art approaches that exhibit domain-specific clustering,
our approach demonstrates persistent descriptor distributions across domain shift, as evidenced by the significant overlap between \ThreeDMatch~(training domain) and other unseen datasets.
This validates that our point normalization strategy effectively produces domain-invariant features, which is a key factor enabling zero-shot generalization.

\begin{figure}[t!]
 	\centering
  	\includegraphics[width=1.0\columnwidth]{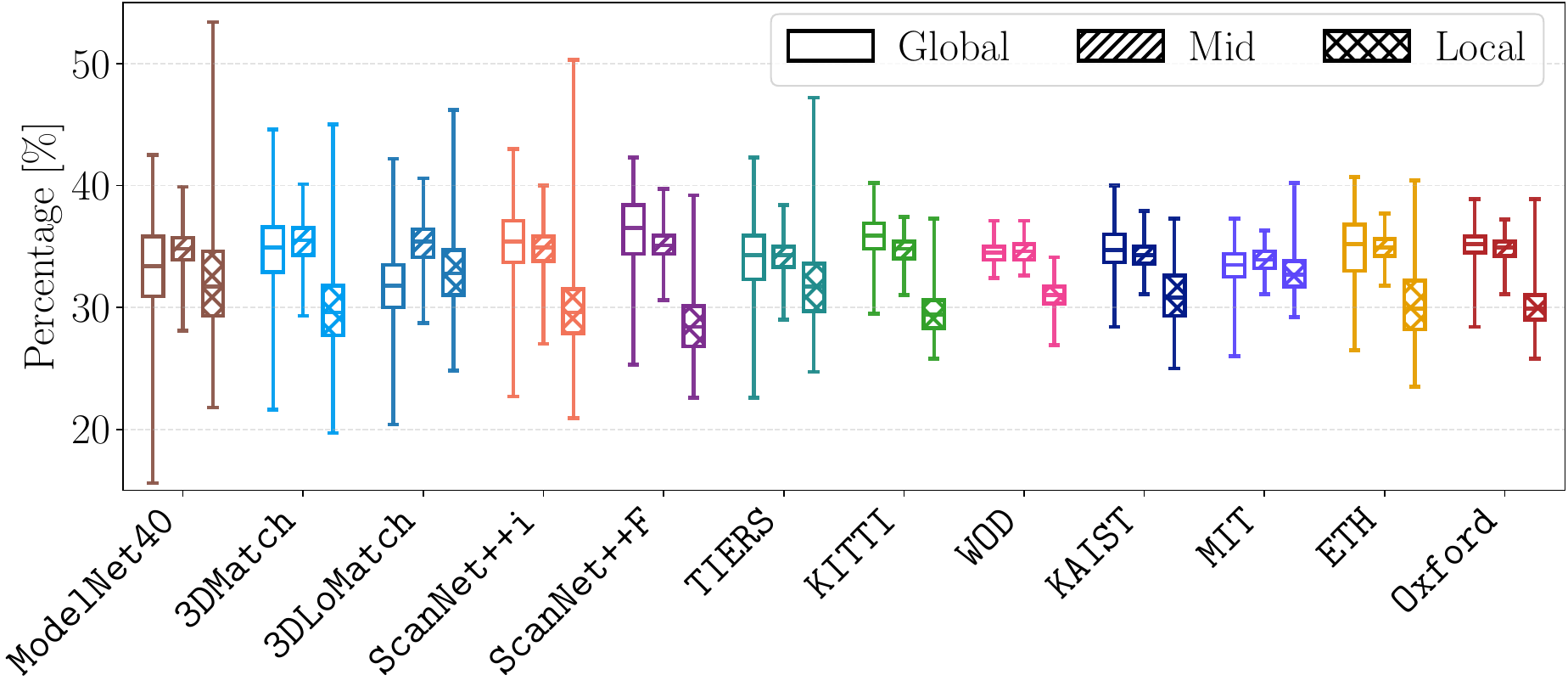}
        \vspace{-5mm}
 	\caption{Distribution of correspondences at different scales (global, middle, local) across datasets, showing the contribution of multi-scale matching.}
    \label{fig:corr_stats}
    \vspace{-2mm}    
\end{figure}

\vspace{2mm}
\noindent\textbf{Multi-scale matching strategy.}
To understand the contribution of our multi-scale matching strategy, we analyze how correspondences at different scales complement each other.
\Cref{fig:corr_stats} shows the distribution of correspondences found at different scales~(global, middle, local) across datasets,
revealing that each scale captures distinct correspondence patterns.
This experimental evidence indicates that certain correspondences can only be obtained when considering multiple scales, as those found at a local scale are not necessarily a subset of those at a global scale.

\begin{table}[t!]
    \centering  
    \setlength{\tabcolsep}{3pt}
    \caption{Ablation study on multi-scale combinations and early exit strategies evaluated on \ThreeDMatch~\cite{Zeng17cvpr-3dmatch}. Top: different scale combinations. Bottom: comparison with RANSAC and \kmsolver~\cite{Lim25icra-KISSMatcher} (referred to as \textit{our solver}) for early exit.}
    {\scriptsize
    \begin{tabular}{ccc|cccc}
        \toprule \midrule
    Local & Middle  & Global  &  {RTE [cm]\,$\downarrow$}  &   {RRE [°]\,$\downarrow$} & {Succ. rate [\%]\,$\uparrow$} & Hz $\uparrow$ \\ \midrule
     \checkmark &   &                      & 6.57 & 2.15 & 84.06 & \firstc \textbf{5.61}  \\
      &   \checkmark    &                  & 5.87 & 1.85 & 93.38 & \thirdc 5.47  \\
      &   & \checkmark                     &  6.06 & 1.91 & 93.57 & \secondc 5.49  \\
      \checkmark &      \checkmark         & & \firstc \textbf{5.73} & {1.81} & 94.31 & 2.35 \\
      \checkmark &      &  \checkmark      & \secondc 5.77 & {1.81} & 94.02 & 2.36 \\
        & \checkmark  &  \checkmark        & \thirdc 5.78 & {1.81} & 94.62 & 2.33 \\
      \checkmark & \checkmark & \checkmark & \thirdc 5.78 & \secondc {1.79} & \firstc \textbf{95.58} & 1.81\\  \midrule
    \multicolumn{3}{c|}{W/ Early Exit}  &  {RTE [cm]\,$\downarrow$}  &   {RRE [°]\,$\downarrow$} & {Succ. rate [\%]\,$\uparrow$} & Hz $\uparrow$ \\ \midrule
    \multicolumn{3}{c|}{W/ RANSAC, $\tau_N= 10$} & 5.87 & 1.85 & 93.38 & \thirdc 5.47 \\
    \multicolumn{3}{c|}{W/ RANSAC, $\tau_N= 50$} & 5.83 & \secondc 1.79 & 94.54 & 5.19 \\
    \multicolumn{3}{c|}{W/ our solver, $\tau_N= 10$} & 5.88 & \firstc \textbf{1.78} & 93.94 & 5.15 \\
    \multicolumn{3}{c|}{W/ our solver, $\tau_N= 50$} & 5.79 & \secondc 1.79 & \secondc 94.95 & 4.72 \\ 
    \midrule \bottomrule
    \end{tabular}
    }
    \vspace{-2mm}
    \label{table:ablation-triscale}
\end{table}

\Cref{table:ablation-triscale} quantifies this observation.
While using only the middle scale achieves 93.38\% success rate at 5.47\,Hz, incorporating all three scales increases the success rate to 95.58\%, confirming that correspondences across scales complement each other.
However, this improvement comes at the cost of reduced speed~(1.81\,Hz), introducing a trade-off between accuracy and computational efficiency.
This allows users to balance performance and speed based on their specific application requirements.

\subsection{\texorpdfstring{\rev{Sensitivity to parameter settings}}{Sensitivity to parameter settings}}\label{sec:param_sensitivity}

\begin{figure}[t!]
    \centering
    \includegraphics[width=\columnwidth]{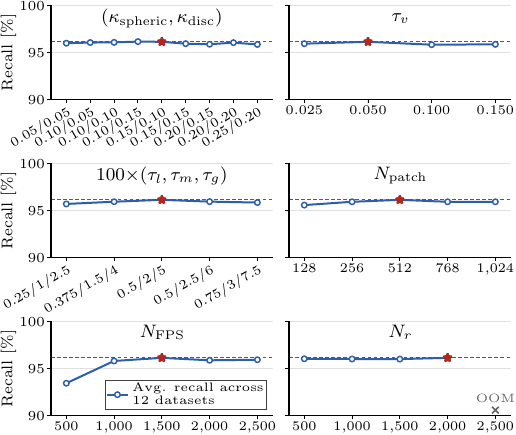}
    \caption{\rev{Benchmark-average recall obtained by varying one parameter group at a time while fixing all other parameters to the default values in \Cref{table:bufferx_params}.
    Red stars indicate the default parameter values used throughout our experiments.
    OOM~(\ie out of memory) denotes configurations that exceeded the available GPU memory.
    Full per-dataset recalls are reported in Tables~S2 and~S3 of the supplementary material.}}
    \label{fig:param_sensitivity_compact}
    \vspace{-2mm}
\end{figure}

\rev{The number of parameters in \Cref{table:bufferx_params} may initially suggest that identifying a reliable configuration requires extensive tuning.
However, our central finding is that the parameter setting used in this work is not a narrow optimum: a broad range of configurations maintains comparable benchmark-wide recall.
To verify this, we vary one parameter group at a time on the 12-dataset benchmark while fixing all parameters not under test to their default values.
All experiments use the same \ThreeDMatch-trained model without retraining or per-dataset parameter tuning.}

\rev{\Cref{fig:param_sensitivity_compact} reveals two main observations.
First, most parameters have only a limited effect on registration performance: varying the scale-density ratios $(\tau_l,\tau_m,\tau_g)$, voxelization coefficients $(\kappa_\text{spheric},\kappa_\text{disc})$, or sphericity threshold $\tau_v$ changes the benchmark-average recall by at most 0.44 percentage points, which is negligible.
Second, among the sampling-related budgets, varying $N_\text{patch}$ has little effect on recall, whereas $N_\text{FPS}$ and $N_r$ are more resource dependent: $N_\text{FPS}=500$ reduces recall by 2.71 points due to insufficient keypoints, while increasing $N_r$ beyond the default can exceed GPU memory on high-resolution inputs.}

\rev{Taken together, these results indicate that the geometry-driven parameters and $N_\text{patch}$ can remain fixed across datasets, whereas $N_\text{FPS}$ and $N_r$ primarily govern the trade-off between registration accuracy and resource usage.}

\subsection{Analysis of \fastversion}\label{sec:fast_buffer_x_exp}

\begin{figure}[t!]
    \centering
    \includegraphics[width=0.45\textwidth]{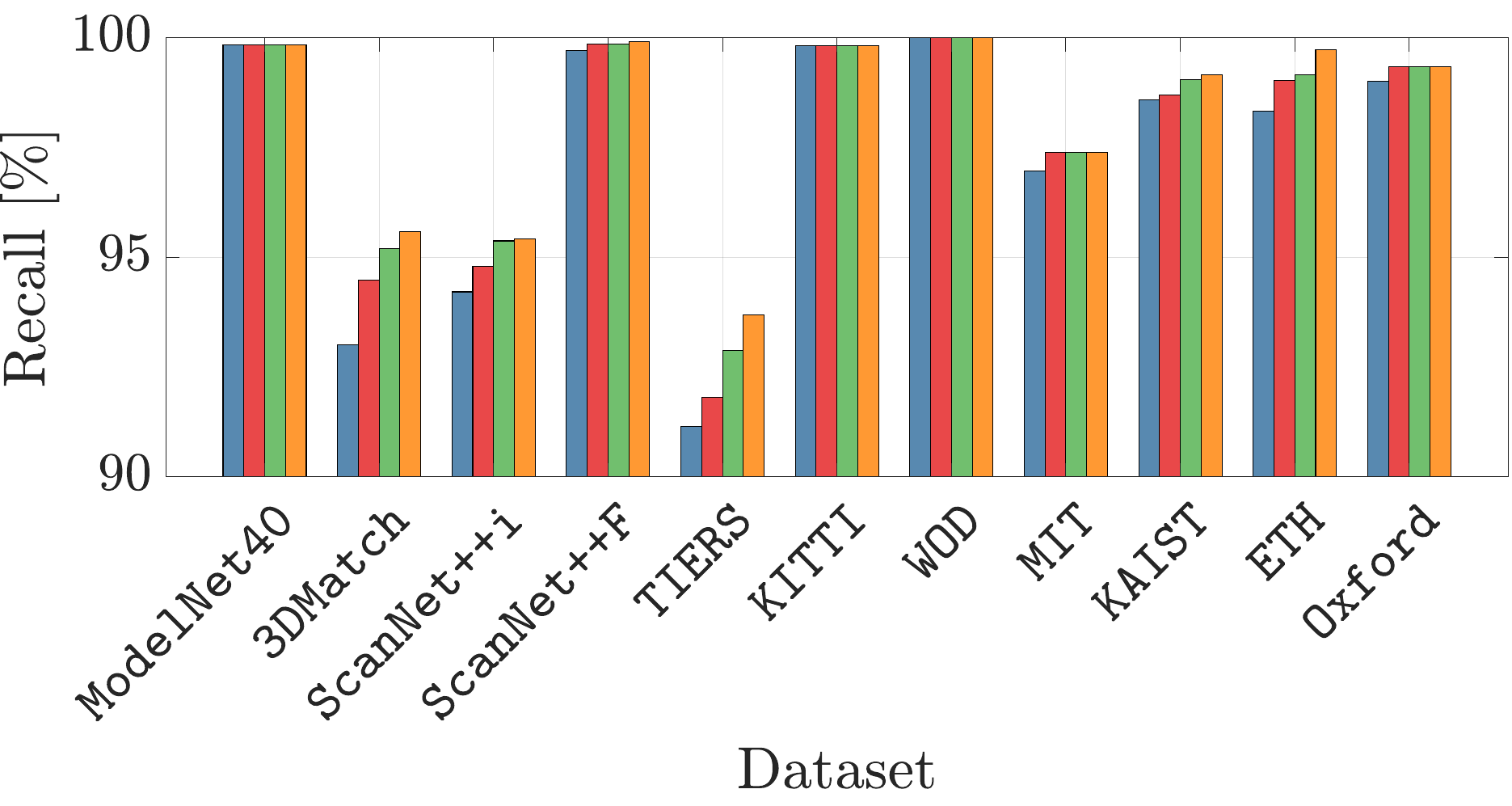}
    \vfill
    \includegraphics[width=0.45\textwidth]{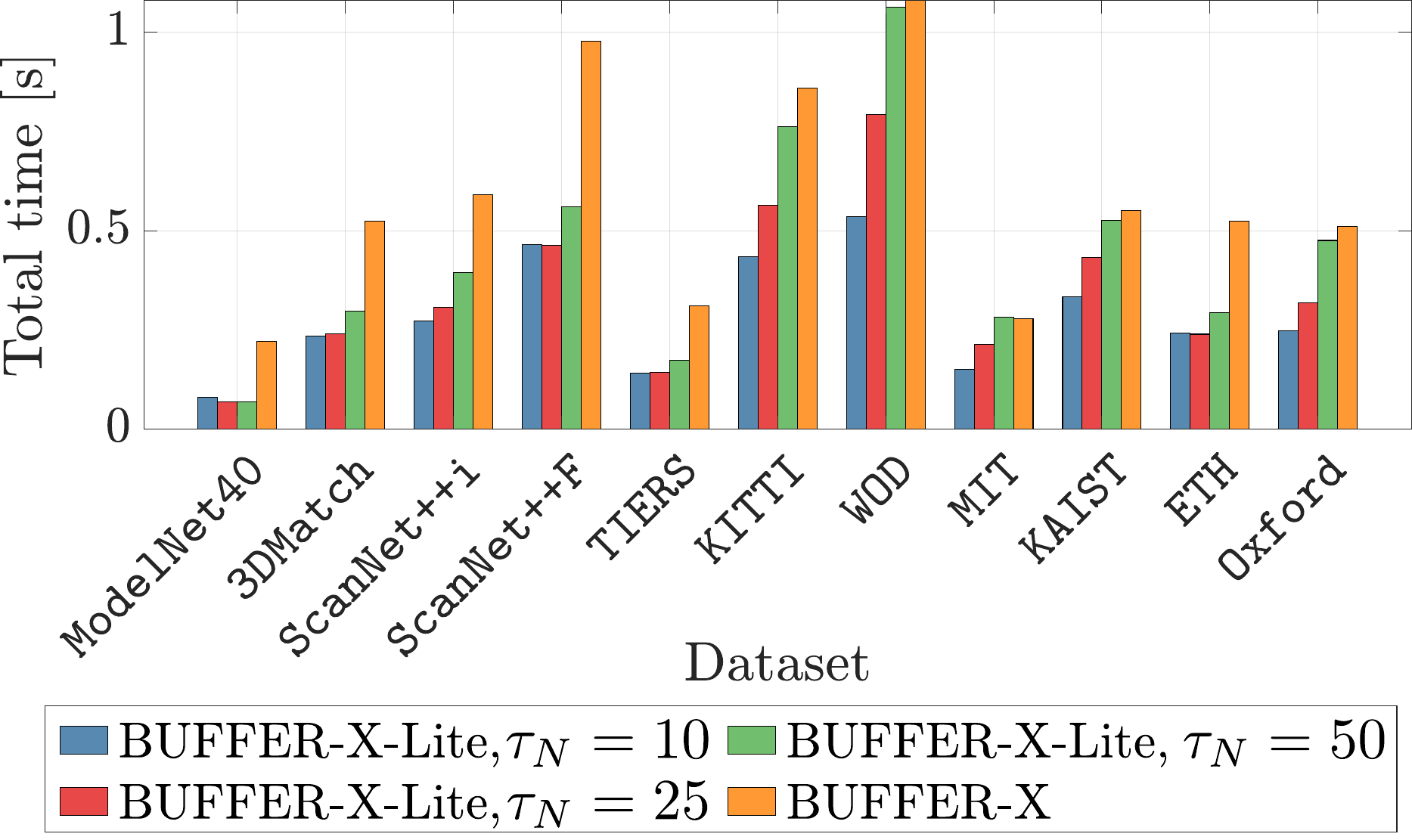}
    \caption{Comparison of success rate and total computation time between \fastversion{} \rev{at different early-exit thresholds and the original \oursname.}}
    \label{fig:comparison_adaptive}
\end{figure}

\begin{figure}[t!]
    \centering
    \includegraphics[width=0.48\textwidth]{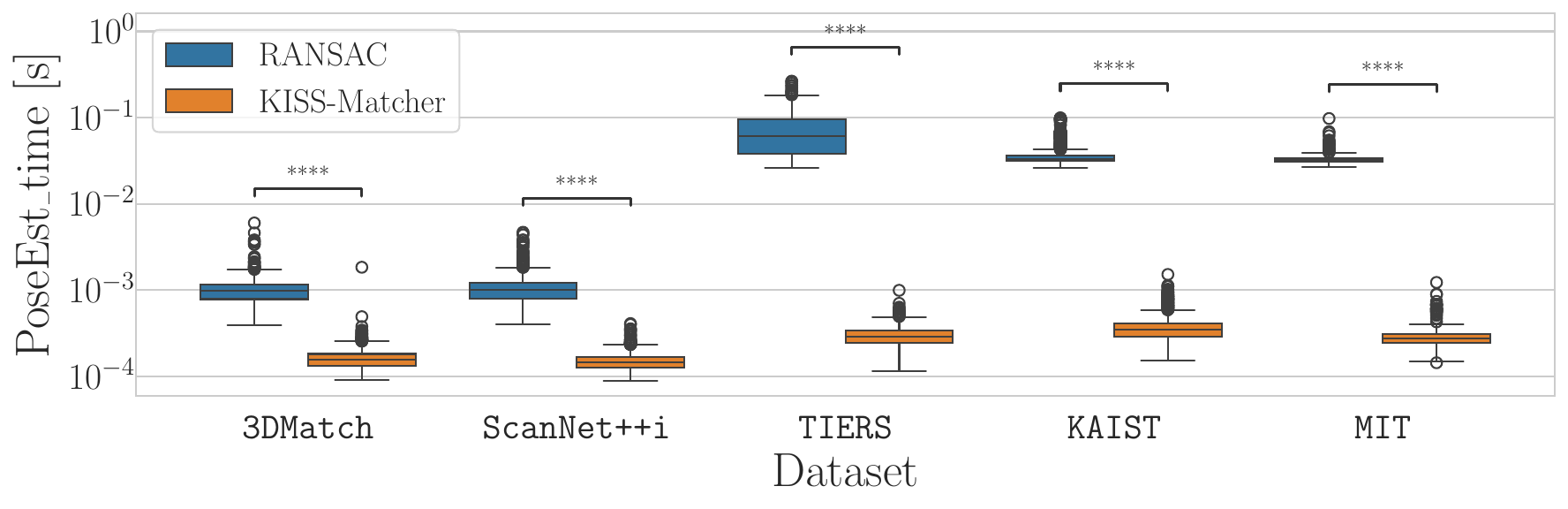}
    \caption{Pose estimation time comparison between RANSAC and \kmsolver given the same correspondences.
    RANSAC, which is still widely used for pose estimation in learning-based approaches, exhibits substantially longer computation time for LiDAR-acquired point clouds~(\ie in \TIERS, \KAIST, and \MIT), whereas \kmsolver~achieves up to 100$\times$ faster performance.}
    \label{fig:pose_est}
    \vspace{-2mm}
\end{figure}

\begin{figure}[t!]
    \centering
    \includegraphics[width=0.48\textwidth]{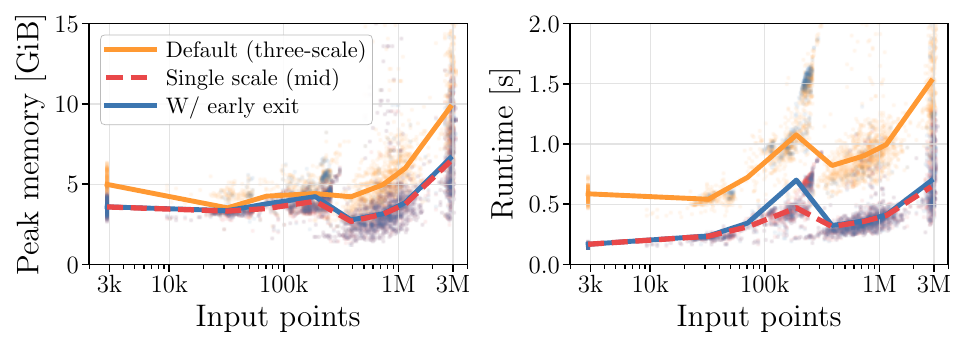}
    \caption{\rev{Peak GPU memory (left) and total runtime (right) versus raw input point count for the default three-scale pipeline, the single-scale variant, and pair-adaptive early exit.
    Dots denote individual registration pairs, and solid lines denote binned medians.
    Note that pooling multiple datasets yields uneven point-count coverage and varying dataset composition across bins, thereby producing local non-monotonicity in the median curves.}}
    \label{fig:memory_scalability}
    \vspace{-2mm}
\end{figure}

\begin{figure}[t!]
    \centering
    \subfloat[RANSAC]{
      \begin{minipage}[b]{0.215\textwidth}
        \includegraphics[width=\linewidth]{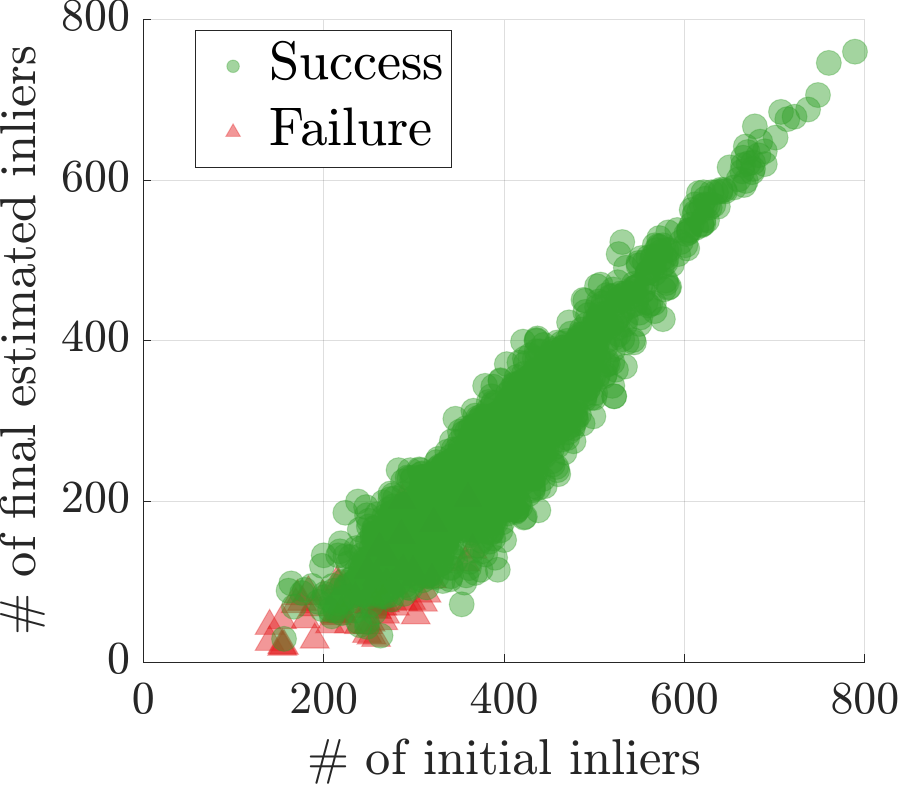}
        \includegraphics[width=\linewidth]{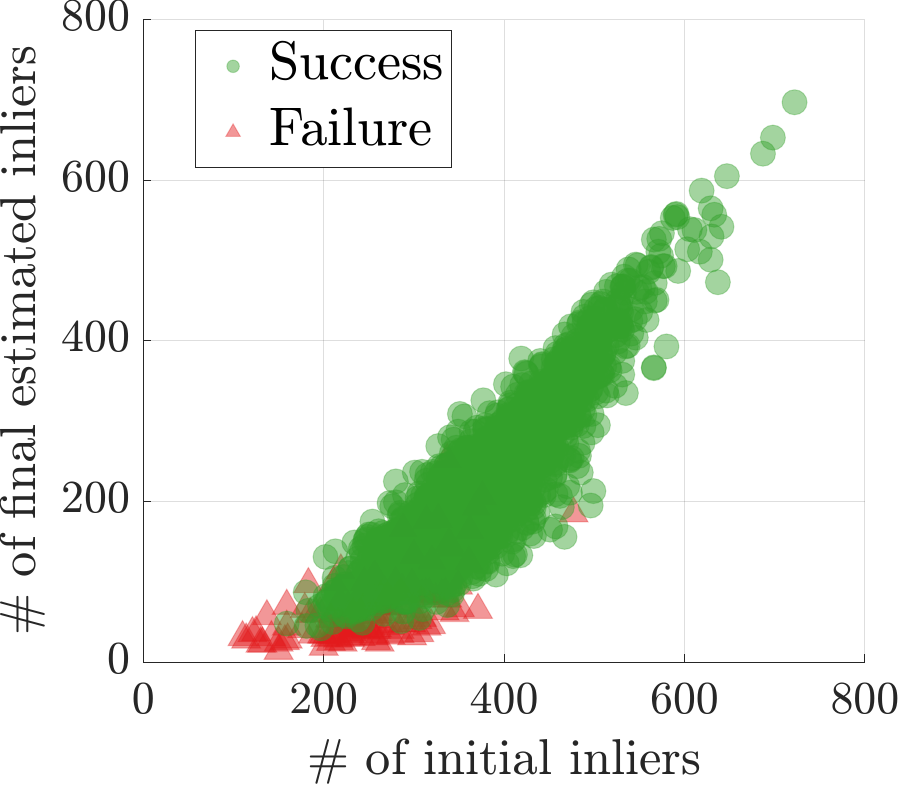}
        \includegraphics[width=\linewidth]{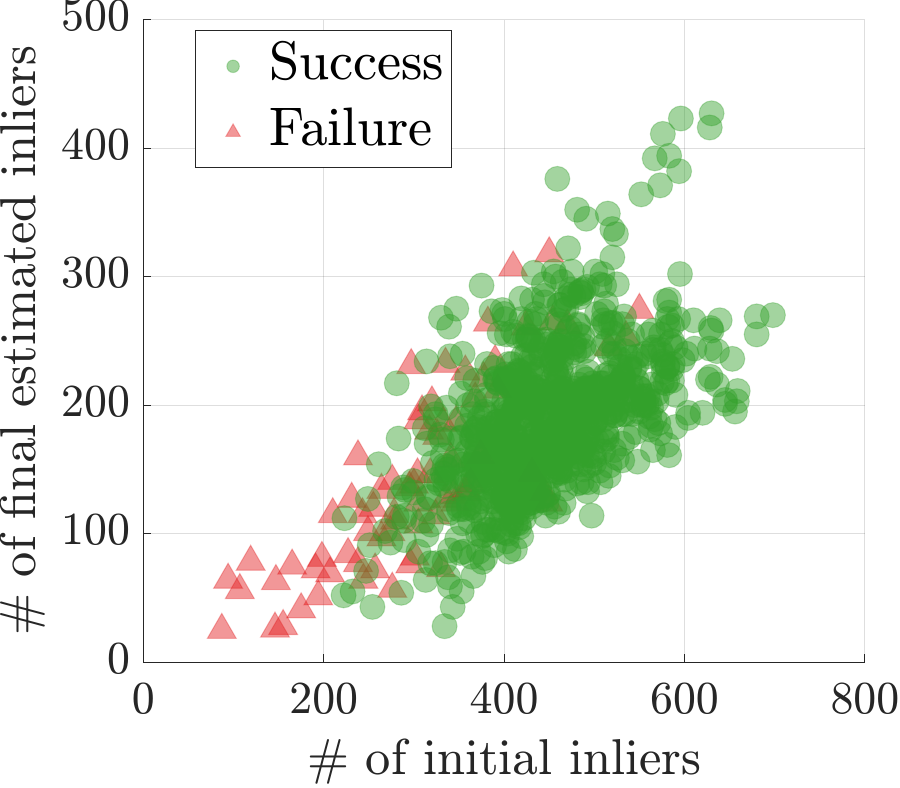}
        \includegraphics[width=\linewidth]{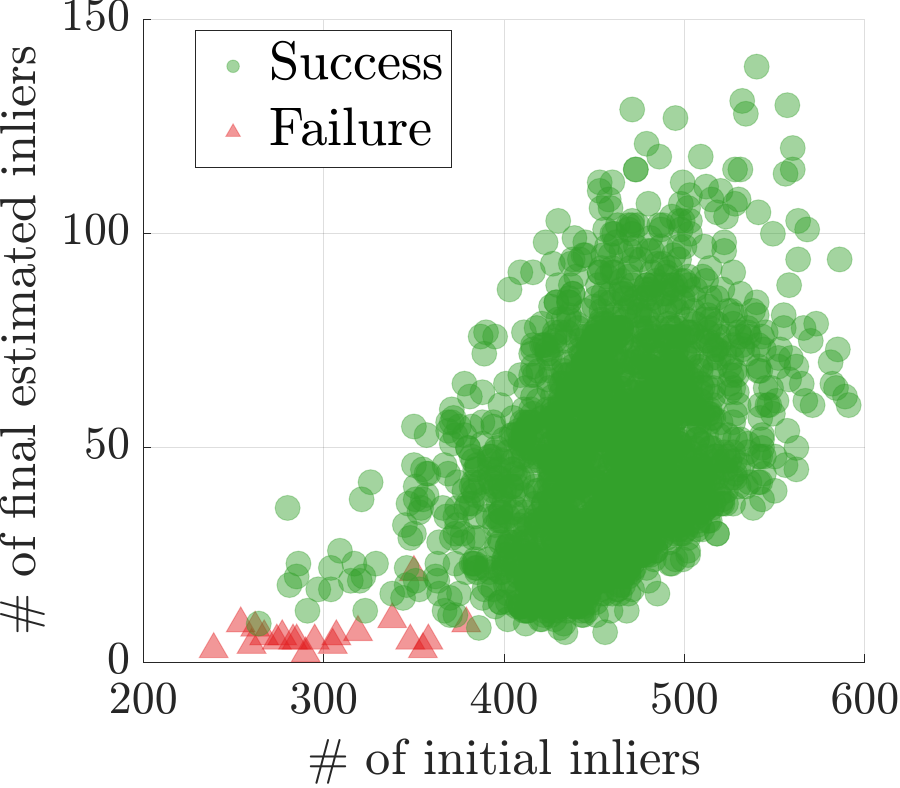}
      \end{minipage}
    }
    \subfloat[\kmsolver]{
      \begin{minipage}[b]{0.215\textwidth}
        \includegraphics[width=\linewidth]{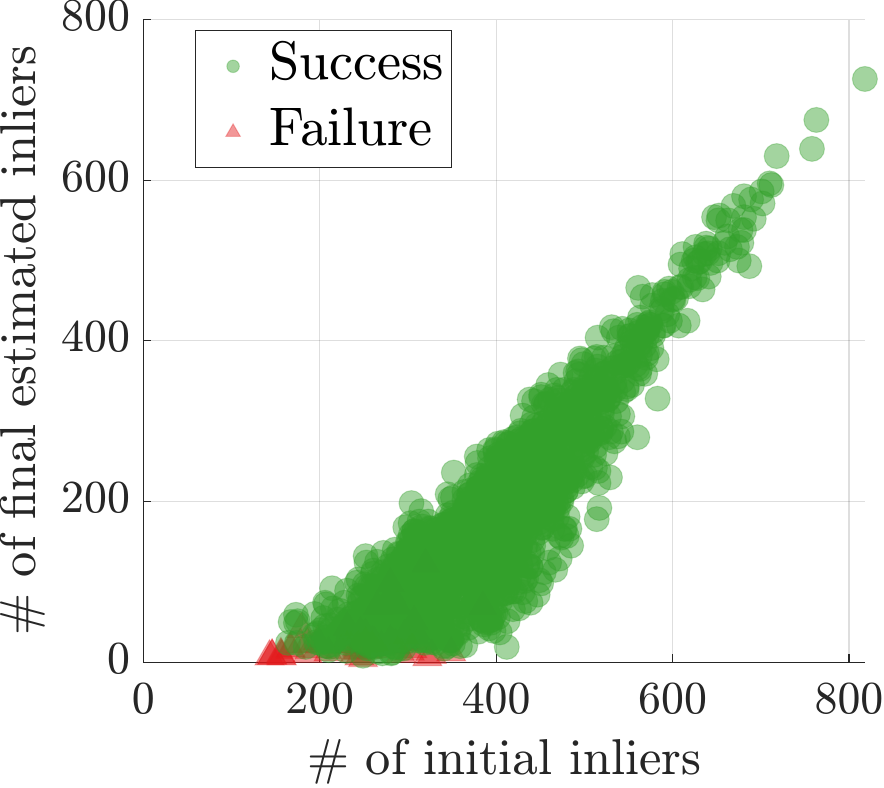}
        \includegraphics[width=\linewidth]{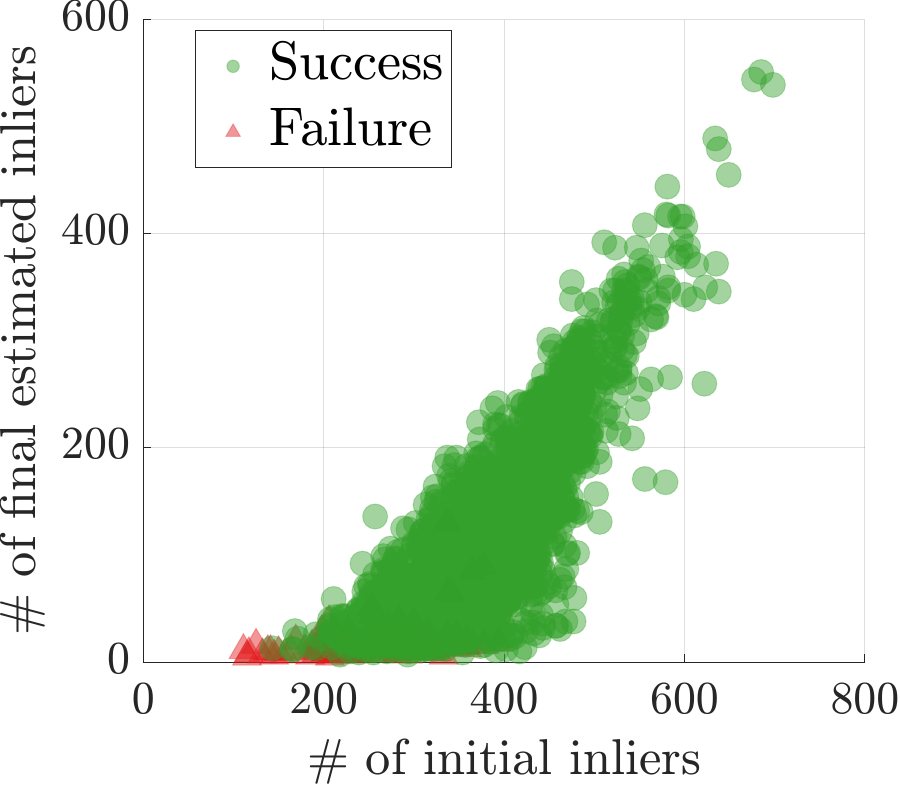}
        \includegraphics[width=\linewidth]{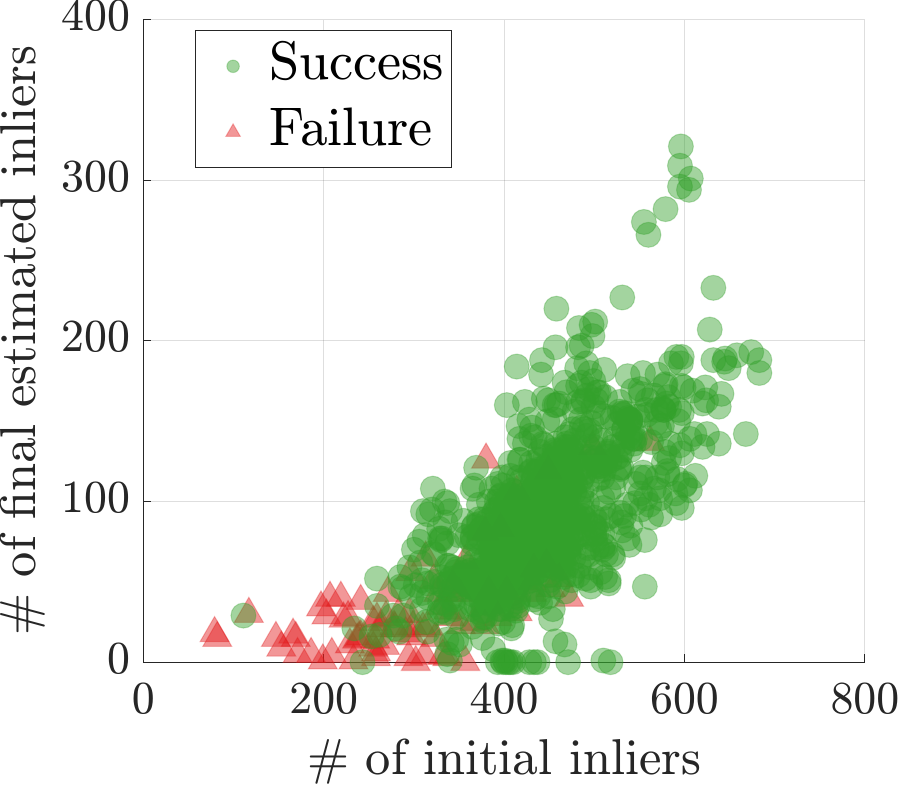}
        \includegraphics[width=\linewidth]{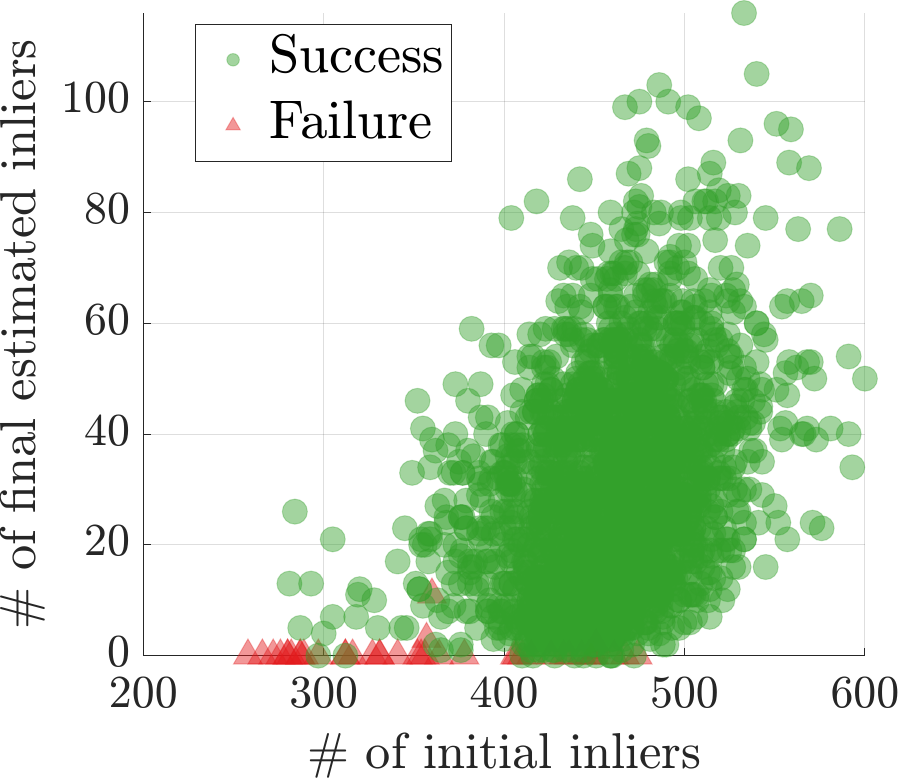}
      \end{minipage}
    }
    \caption[Comparison of mutual correspondences versus inliers for RANSAC and KISS-Matcher solver.]{Comparison of mutual correspondences versus inliers for RANSAC and \kmsolver across \ThreeDMatch, \ScanNetppi, \TIERS, and \KAIST. The number of estimated inliers in failure cases for RANSAC is relatively widespread across lower values, whereas that of \kmsolver is more likely to be near-zero.}
    \label{fig:inliers_comparison}
    \vspace{-2mm}
\end{figure}

\rev{Next, we evaluate \fastversion across the complete registration pipeline, spanning BUFFER-X descriptor generation and feature matching, pair-adaptive early exit, and the correspondence-pruning and pose-estimation modules developed in KISS-Matcher~\cite{Lim25icra-KISSMatcher}, as explained in \Cref{sec:fastbufferx}.}
\rev{As shown in \Cref{fig:comparison_adaptive}, \fastversion with $\tau_N=25$ requires only 56.8\% of the total computation time of the original \oursname on average across all datasets.}
\Cref{fig:pose_est} shows that \kmsolver takes less time than RANSAC, making it particularly suitable for our early exit strategy, which requires multiple pose estimations.

\rev{In addition to total runtime, we report peak GPU memory usage and analyze both quantities across raw input point counts.
As shown in \Cref{fig:memory_scalability}, across the observed point-count ranges, processing fewer scales generally reduces both runtime and memory:
the middle-scale-only variant (denoted \textit{Single scale}) provides a lower-cost reference, while the early-exit variant (denoted \textit{W/ Early exit}) approaches this cost for pairs that can be resolved at the middle scale and processes additional scales only when necessary.
These results suggest that early exit reduces computational and memory costs by terminating after the middle scale for easier pairs while retaining multi-scale processing for more challenging pairs.}

Furthermore, \kmsolver is particularly well-suited for early exit strategy because it can easily detect failure cases by checking the cardinality of final inliers.
As presented in \Cref{fig:inliers_comparison}, given the same number of initial correspondences, \kmsolver separates success and failure cases more clearly using only the number of final inliers.
This enables easier failure detection; as shown in \Cref{table:ablation-triscale}, with the same threshold, the early exit variant with \kmsolver achieves higher success rates.
This is because with RANSAC, the number of inliers in failure cases is relatively widespread across lower values, whereas \kmsolver failures typically have final inlier counts close to zero.

\rev{Together, these results show that \fastversion reduces pipeline-wide computation through both adaptive feature processing and efficient correspondence pruning and pose estimation, while maintaining comparable registration accuracy.}

\subsection{Application: Submap-to-map localization}\label{sec:application}

As an application, \oursname can be exploited for robotic localization in long-term, multi-session environments.
In such settings, a key challenge is localizing within a pre-built map without relying on pose initialization or GPS, especially when the robot resumes operation after significant time gaps~\cite{Lim25arXiv-MultiMapcher}.
This scenario, commonly referred to as the \emph{kidnapped robot} problem, arises in practical deployments such as daily autonomous delivery or multi-day inspection missions.

\begin{figure}[t!]
  \centering
  \includegraphics[width=0.48\textwidth]{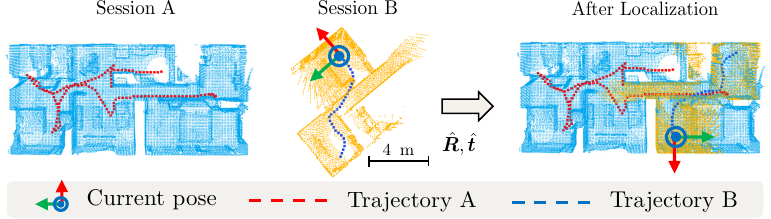}
  \caption{Application of our approach to robotic systems: online submap-to-map localization between two sessions on the Cantwell scene from the Gibson dataset\cite{xia2018-gibsonenv}.}
  \label{fig:multi-session}
\end{figure}

As shown in \Cref{fig:multi-session}, \rev{\oursname enables} robust submap-to-map localization without pose priors.
During online operation, the robot incrementally constructs a local submap from recent LiDAR scans.
\rev{When localization is required, \oursname registers this submap against a pre-built global map and directly estimates the robot's global pose.
Note that the pretrained descriptor model remains fixed during deployment and requires no environment-specific retraining.}
This suggests that the approach has the potential to support multi-session localization in both indoor and outdoor environments.

\subsection{Limitations}\label{sec:limitation}

\begin{figure}[t!]
    \centering
    \includegraphics[width=\columnwidth]{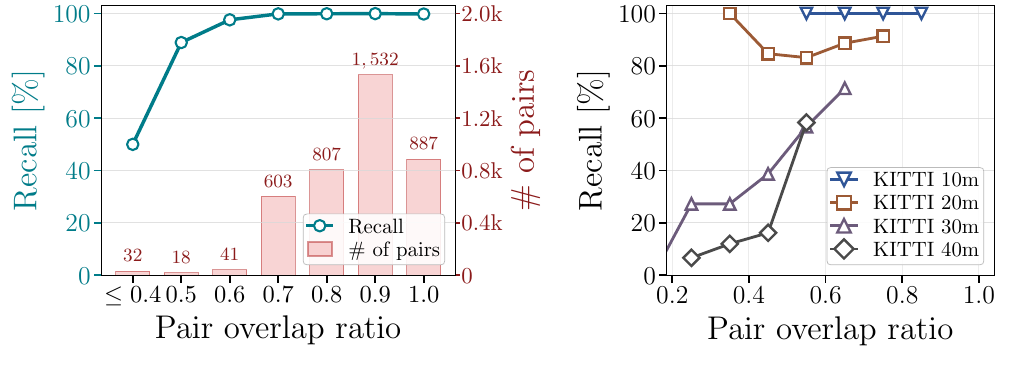}
    \caption{\rev{Overlap-conditioned analysis on outdoor benchmarks.
    Registration recall and pair count of \oursname over all outdoor benchmark pairs grouped by overlap ratio~(left) and
    controlled \KITTI evaluation using a protocol adopted in prior work~\cite{zeng2025unlocking}, where the odometry-based pair-selection distance is increased from 10\,m to 20\,m, 30\,m, and 40\,m without cropping or otherwise modifying the scans (right).}}
    \label{fig:outdoor_low_overlap}
    \vspace{-2mm}
\end{figure}

As seen in \Cref{table:success_rates}, our approach showed a lower success rate on \ThreeDLoMatch, which has only 10--30\% overlap.
\rev{This vulnerability to partial overlap was also observed in outdoor scenes, as presented in \Cref{fig:outdoor_low_overlap}}.
This is because Eq.~\Cref{eq:consensus_max} selects correspondences with the largest cardinality as inliers.
However, in partial overlap scenarios, maximizing the number of correspondences might not yield the actual global optimum~(\ie there might exist $\mathcal{I}^*$ that satisfies $|\mathcal{I}^*| \leq |\mathcal{I}|$ but leads to a better relative pose estimate).
This highlights a trade-off between generalization and robustness to partial overlaps.

\rev{More specifically, this trade-off arises because BUFFER-X deliberately uses a domain-agnostic consensus objective rather than learned overlap or inlier-filtering cues.
For example, Predator estimates overlap scores to bias matching toward shared regions~\cite{Huang21cvpr-PREDATORRegistration} and 
BUFFER uses an inliers generator to search reliable correspondences and improve the inlier ratio on low-overlap pairs~\cite{Ao23CVPR-BUFFER};
however, such learned cues can also weaken out-of-the-box transfer under scale or sensor shifts, as observed in \Cref{sec:analyses,sec:sota_comparison}.
Thus, BUFFER-X improves cross-domain generalization, but it can be less reliable when the largest mutually consistent correspondence set is not the most pose-informative one.
Additional overlap-conditioned evaluations and an analysis of this ambiguity are provided in Secs.~C and~D of the supplementary material.}


\section{Conclusion}

This paper addresses zero-shot generalization in point cloud registration by identifying three fundamental limitations: fixed user-defined parameters, learned keypoint detectors with poor transferability, and absolute coordinates that amplify scale mismatches.
\rev{In the practical setting where a descriptor model is trained once on a source dataset and subsequently deployed without target-domain retraining or parameter tuning, \textit{\oursname} overcomes these limitations} through geometric bootstrapping, distribution-aware sampling, patch-based normalization, and hierarchical multi-scale matching.

We evaluated on a comprehensive benchmark spanning 12 datasets across environmental scales, scanning patterns, acquisition setups, and geographic regions, including heterogeneous LiDAR configurations.
\rev{Beyond the feasibility study reported in our previous work, the expanded experiments provide quantitative evidence for why BUFFER-X generalizes, including analyses of descriptor distributions, complementary correspondence patterns across scales, and parameter sensitivity.}
\rev{We further present \fastversion from a holistic system perspective, integrating BUFFER-X descriptor generation and feature matching with adaptive early exit and the fast and robust correspondence pruning and pose estimation modules developed in KISS-Matcher~\cite{Lim25icra-KISSMatcher}, thereby reducing total computation time while maintaining comparable registration accuracy.}

Future work will explore integrating semantic \rev{and overlap-aware} cues to improve robustness in \rev{highly ambiguous or} extremely low-overlap scenarios while \rev{preserving out-of-the-box deployment on unseen domains}.

\section*{Acknowledgments}
This work was supported by IITP grants funded by the Korea government (MSIT) (RS-2021-II211343: AI Graduate School Program at Seoul National University (5\%), RS-2025-25442338: AI star Fellowship Support Program (45\%), and RS-2025-02303703: Real-world multi-space fusion and 6DoF free-viewpoint immersive visualization for extended reality (20\%)), and by NRF grant funded by the Korea government (MSIT) (No. RS-2024-00461409) (30\%).\par

\bibliographystyle{IEEEtran}
\bibliography{myRefs,extracted_refs}




\makeatletter
\let\BUFFERXoriginalIEEEBIOskipN\@IEEEBIOskipN
\def\@IEEEBIOskipN{3.5\baselineskip}
\makeatother

\begin{IEEEbiography}[{\includegraphics[width=1in,height=1.25in,clip,keepaspectratio]{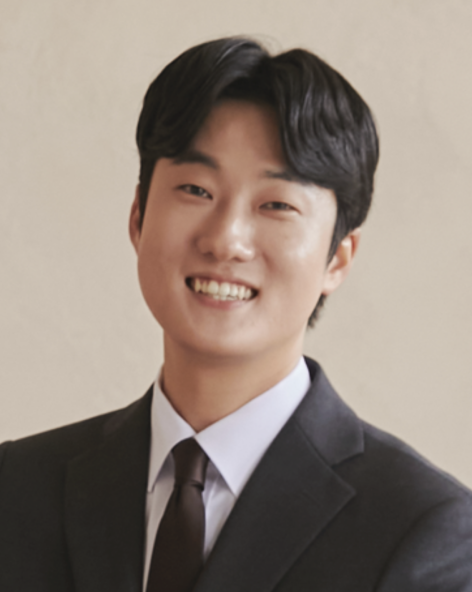}}]{Hyungtae Lim} (Member, IEEE) received the B.S. degree in mechanical engineering, and M.S. and Ph.D. degrees in electrical engineering from the Korea Advanced Institute of Science and Technology (KAIST), Daejeon, Republic of Korea, in 2018, 2020, and 2023, respectively.
    He is currently a postdoctoral associate in the Laboratory for Information \& Decision Systems~(LIDS), Massachusetts Institute of Technology~(MIT), Massachusetts, USA.
    His research interests include SLAM (simultaneous localization and mapping), 3D registration, 3D perception, long-term map management, spatial AI, and deep learning.
\end{IEEEbiography}

\begin{IEEEbiography}[{\includegraphics[width=1in,height=1.25in,clip,keepaspectratio]{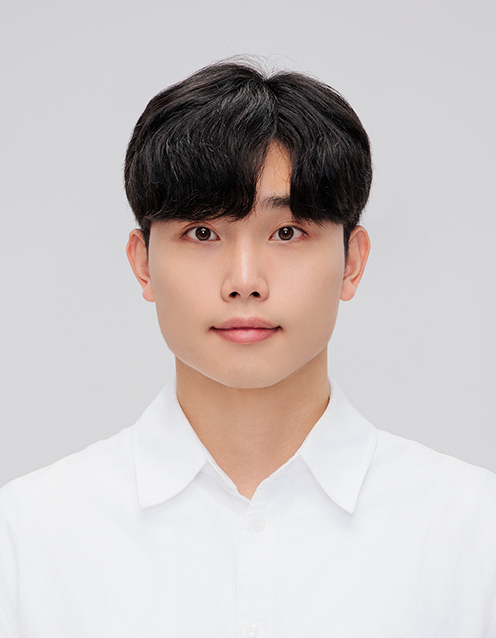}}]{Minkyun Seo} (Graduate Student Member, IEEE)
received the B.S. degrees in Architectural Engineering and Computer Science and Engineering from Seoul National University~(SNU), Seoul, Republic of Korea, in 2025, where he is currently pursuing the Ph.D. degree in Computer Science and Engineering through the integrated Ph.D. program.
He is a member of the Visual and Geometric Intelligence Laboratory, advised by Prof. Jaesik Park. 
His research interests include 3D registration, 3D perception, and 3D reconstruction.
\end{IEEEbiography}

\begin{IEEEbiography}[{\includegraphics[width=1in,height=1.25in,clip,keepaspectratio]{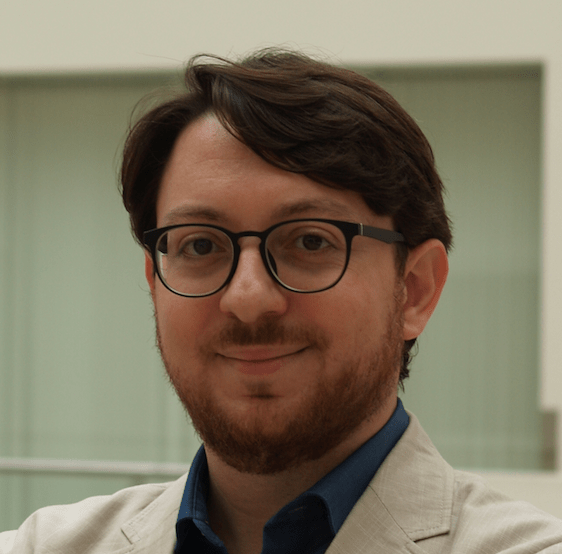}}]{Luca Carlone}
(Senior Member, IEEE) received
the B.S. degree in mechatronics from the Polytechnic University of Turin, Italy, in 2006, the S.M.
degree in mechatronics from the Polytechnic University of Turin, Italy, in 2008, the S.M. degree in
automation engineering from the Polytechnic University of Milan, Italy, in 2008, and the Ph.D. degree in robotics from the Polytechnic University of Turin, in 2012. 
He is currently the Leonardo
career development associate professor with the
Department of Aeronautics and Astronautics, the
Massachusetts Institute of Technology, and a Principal Investigator in
the Laboratory for Information \& Decision Systems~(LIDS). He joined
LIDS as a postdoctoral associate (2015) and later as a research scientist~(2016), after spending two years as a postdoctoral fellow with the Georgia Institute of Technology (2013-2015). 
His research interests include nonlinear estimation, numerical and distributed optimization, and probabilistic inference, applied to sensing, perception, and decision-making in single and multi-robot systems. His work includes seminal results on certifiably correct algorithms for localization and mapping, as well as
approaches for visual-inertial navigation and distributed mapping. 
\end{IEEEbiography}

\begin{IEEEbiography}[{\includegraphics[width=1in,height=1.25in,clip,keepaspectratio]{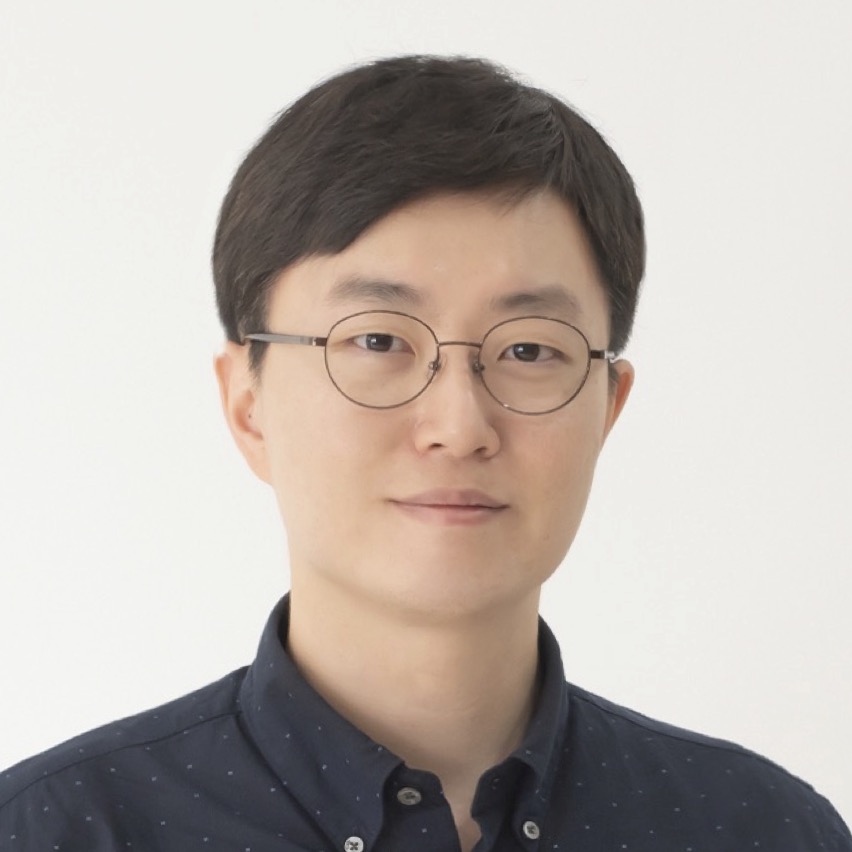}}]{Jaesik Park} (Member, IEEE) received the bachelor’s
degree from Hanyang University in 2009, and the
master’s and Ph.D. degrees from KAIST in 2011
and 2015, respectively. He is an associate professor
with the Department of Computer Science and Engineering, Seoul National University. He worked at
Intel as a research scientist (2015-2019), and he was
a faculty member at POSTECH (2019-2023). His
research interests include image synthesis and 3D
scene understanding.
\end{IEEEbiography}

\makeatletter
\let\@IEEEBIOskipN\BUFFERXoriginalIEEEBIOskipN
\makeatother

\clearpage
\twocolumn[
\begin{center}
  {\LARGE\bfseries Toward Zero-Shot Point Cloud Registration Across Diverse
  Scales, Scenes, and Sensor Setups\par}
  \vspace{0.45cm}
  {\large Hyungtae Lim$^{1,*}$, Minkyun Seo$^{2,*}$,
  Luca Carlone$^{1}$, and Jaesik Park$^{2,\dagger}$\par}
  \vspace{0.22cm}
  {\small
  $^{1}$Laboratory for Information and Decision Systems, Massachusetts
  Institute of Technology, Cambridge, MA, USA\par
  $^{2}$Department of Computer Science and Engineering, Seoul National
  University, Seoul, Republic of Korea\par
  $^{*}$Equal contribution \qquad
  $^{\dagger}$Corresponding author\par}
  \vspace{0.45cm}
  {\large\bfseries Supplementary Materials\par}
\end{center}

]

\raggedbottom

\setcounter{section}{0}
\setcounter{subsection}{0}
\setcounter{equation}{0}
\setcounter{figure}{0}
\setcounter{table}{0}
\setcounter{algocf}{0}
\renewcommand{\thesection}{\Alph{section}}
\renewcommand{\thesubsection}{\Alph{subsection}}
\renewcommand{\theequation}{S\arabic{equation}}
\renewcommand{\thefigure}{S\arabic{figure}}
\renewcommand{\thetable}{S\arabic{table}}
\renewcommand{\thealgocf}{S\arabic{algocf}}
\renewcommand{\theHsection}{supp.\Alph{section}}
\renewcommand{\theHsubsection}{supp.subsection.\arabic{subsection}}
\renewcommand{\theHequation}{supp.equation.\arabic{equation}}
\renewcommand{\theHfigure}{supp.figure.\arabic{figure}}
\renewcommand{\theHtable}{supp.table.\arabic{table}}

\subsection{\texorpdfstring{\rev{Extension beyond prior work}}{Extension beyond prior work}}
\label[suppsection]{app:iccv_extension}

\rev{To make the scope of this extension explicit, \Cref{table:iccv_journal_extension} separates the components inherited from our prior work~\cite{Seo25ICCV-BUFFERX} from the material added in this extended manuscript.
The BUFFER-X backbone is retained to make the presentation self-contained and is not claimed as a new contribution of this extension.
The extension instead shifts the focus from demonstrating whether zero-shot transfer is possible to quantitatively explaining why it works, identifying when it becomes unreliable, and reducing the cost of the complete registration pipeline.}

\begin{table*}[!b]
  \centering
  \caption{\rev{Explicit comparison between our prior work and this extended manuscript. This extension retains the BUFFER-X backbone for a self-contained presentation while adding a broader evaluation scope, quantitative analyses, and an efficiency-oriented system-level integration of the front-end and solver.}}
  \label{table:iccv_journal_extension}
  \setlength{\tabcolsep}{4pt}
  {\scriptsize\color{revisionblue}
  \begin{tabular}{p{0.17\textwidth}p{0.36\textwidth}p{0.39\textwidth}}
    \toprule
    Aspect & Prior work & This extended manuscript \\
    \midrule
    Core BUFFER-X components
      & Geometric bootstrapping, distribution-aware FPS, patch-level coordinate normalization, hierarchical multi-scale matching, and cross-scale consensus maximization.
      & Inherited without being presented as a new contribution of this extension; retained to provide a self-contained description. \\
    \addlinespace
    Primary research focus
      & Establishes the feasibility of zero-shot transfer from a single training dataset to unseen indoor and outdoor scenes.
      & Shifts the focus from \emph{whether} transfer is possible to \emph{why} it works and \emph{under which conditions} it remains reliable. \\
    \addlinespace
    Efficiency extension
      & Not included.
      & Presents BUFFER-X-Lite through a system-level integration of the BUFFER-X front-end, pair-adaptive early exit, and the previously developed KISS-Matcher correspondence-pruning and pose-estimation modules; Algorithm~\ref{alg:buffer_x_lite} provides the complete procedure. \\
    \addlinespace
    Evaluation scope
      & Focuses primarily on homogeneous indoor and outdoor registration.
      & Expands to a 12-dataset benchmark spanning object-scale, indoor, and outdoor scenes, including ModelNet40 and six heterogeneous LiDAR source--target directions: OS128$\rightarrow$OS64, OS64$\rightarrow$Vel16, Vel16$\rightarrow$OS128, Aeva$\rightarrow$Avia, Avia$\rightarrow$Ouster, and Ouster$\rightarrow$Aeva. \\
    \addlinespace
    New quantitative analyses
      & Component-level ablations supporting the initial zero-shot feasibility study.
      & Adds five analysis categories: descriptor-distribution consistency, correspondence complementarity across scales, benchmark-wide parameter sensitivity, pipeline-wide efficiency, and overlap-conditioned failure behavior. \\
    \addlinespace
    New practical material
      & Not included.
      & Adds one submap-to-map localization application, complete BUFFER-X-Lite pseudocode, and an analysis of global-optimum ambiguity under severe partial overlap. \\
    \bottomrule
  \end{tabular}}
\end{table*}

\subsection{\texorpdfstring{\rev{Pseudocode of \fastversion}}{Pseudocode of BUFFER-X-Lite}}
\label[suppsection]{app:buffer_x_lite}

\rev{Algorithm~S1 summarizes the complete \fastversion pipeline described in Sec.~V.
The middle scale is evaluated first because it provides the most favorable accuracy--efficiency trade-off in Table~IV.
If the final inlier set returned by \kmsolver contains at least $\tau_N$ correspondences, the method returns the current pose without computing the local and global descriptors.
Otherwise, the remaining scales are processed and accumulated before cross-scale consensus maximization and final pose estimation.}

\providecommand{\annot}[1]{\textcolor{gray!80}{\% #1}}

\begin{algorithm}[h!]
{\color{revisionblue}\footnotesize
\SetAlgoLined
\DontPrintSemicolon
\textbf{Input:} \ Source cloud $\srccloud$ and target cloud $\tgtcloud$; early-exit threshold $\tau_N$ \\
\textbf{Output:} Relative transformation $\M{T}$ \\
$\mathcal{P}_r \leftarrow \texttt{select\_larger\_cloud}(\srccloud, \tgtcloud)$  \\
$\mathcal{P}_\text{sampled} = \texttt{sample}(\mathcal{P}_r, \delta_v)$ \annot{Sample $\delta_v$\% of cloud points} \\
\annot{Step 1. Geometric bootstrapping} \\
$v = \texttt{calc\_voxel\_size}(\mathcal{P}_\text{sampled}, \tau_v)$ \annot{See Eq.~(3)} \\
$\mathcal{P} \leftarrow f_v(\mathcal{P}), \mathcal{Q} \leftarrow f_v(\mathcal{Q})$ \annot{Downsample the point clouds} \\
$\mathcal{P}_r \leftarrow \texttt{select\_larger\_cloud}(\srccloud, \tgtcloud)$  \\
$\mathcal{R} = \texttt{estimate\_radii}(\texttt{sample}(\mathcal{P}_r, N_r), [\tau_l, \tau_m, \tau_g])$, \\
\qquad where $\mathcal{R}=[r_l,r_m,r_g]$ \annot{See Eq.~(5)} \\
$\mathcal{R}_\text{Lite}=[r_m,r_l,r_g]$ \annot{Evaluate the middle scale first} \\
\annot{Step 2. Sequential patch embedding with pair-adaptive early exit} \\
$\mathcal{D} = \varnothing, \mathcal{T} = \varnothing$ \annot{Containers of correspondences and transformations} \\
\For{$r_\xi$ \normalfont{in} $\mathcal{R}_\text{Lite}$}{
    $\mathcal{P}_\xi = \texttt{farthest\_point\_sampling}(\mathcal{P}, N_\text{FPS})$  \\
    $\mathcal{Q}_\xi = \texttt{farthest\_point\_sampling}(\mathcal{Q}, N_\text{FPS})$ \\
    $\mathcal{F}^\mathcal{P}_\xi, \mathcal{C}^\mathcal{P}_\xi = \texttt{MiniSpinNet}(\mathcal{P}_\xi, \mathcal{P}, r_\xi)$ \\
    $\mathcal{F}^\mathcal{Q}_\xi, \mathcal{C}^\mathcal{Q}_\xi = \texttt{MiniSpinNet}(\mathcal{Q}_\xi, \mathcal{Q}, r_\xi)$ \\
    \annot{Step 3-1. Nearest neighbor-based intra-scale matching}\\
    $\mathcal{A}_\xi = \texttt{mutual\_matching}(\mathcal{F}^\mathcal{P}_\xi, \mathcal{F}^\mathcal{Q}_\xi)$ \\
    $(\widehat{\mathcal{P}}_\xi, \widehat{\mathcal{Q}}_\xi, \widehat{\mathcal{C}}^{\mathcal{P}}_\xi, \widehat{\mathcal{C}}^{\mathcal{Q}}_\xi) = \texttt{filter}((\mathcal{P}_\xi,\mathcal{F}^\mathcal{P}_\xi,\mathcal{C}^\mathcal{P}_\xi),(\mathcal{Q}_\xi,\mathcal{F}^\mathcal{Q}_\xi,\mathcal{C}^\mathcal{Q}_\xi),\mathcal{A}_\xi)$ \\
    \annot{Step 3-2. Pairwise transformation estimation}\\
    $\mathcal{T}_\xi = \texttt{calc\_pairwise\_R\_and\_t}(\widehat{\mathcal{C}}^{\mathcal{P}}_\xi, \widehat{\mathcal{C}}^{\mathcal{Q}}_\xi)$ \\
    $\mathcal{D}.\texttt{append}((\widehat{\mathcal{P}}_\xi,\widehat{\mathcal{Q}}_\xi))$, $\mathcal{T}.\texttt{append}(\mathcal{T}_\xi)$\\
    \annot{Step 3-3. Cross-scale consensus maximization}\\
    $\mathcal{I} = \texttt{consensus\_maximization}(\mathcal{D}, \mathcal{T})$ \annot{See Eq.~(9)} \\
    \If{$r_\xi = r_m$}{
        $\M{T},\mathcal{I}_{\mathrm{KM}}=\texttt{kiss\_matcher}(\mathcal{D}[\mathcal{I}])$ \annot{Fast pose estimation} \\
        \If{$|\mathcal{I}_{\mathrm{KM}}|\geq\tau_N$}{
            \Return $\M{T}$ \annot{Early exit without computing local/global descriptors}
        }
    }
}
\annot{Step 4. Final pose estimation after processing all required scales}\\
$\M{T},\mathcal{I}_{\mathrm{KM}}=\texttt{kiss\_matcher}(\mathcal{D}[\mathcal{I}])$ \\
\Return $\M{T}$ \\
}
\caption{\rev{Pseudocode of the complete \fastversion registration pipeline.}}
\label{alg:buffer_x_lite}
\end{algorithm}


\begin{figure}[h!]
    \centering
    \includegraphics[width=\columnwidth]{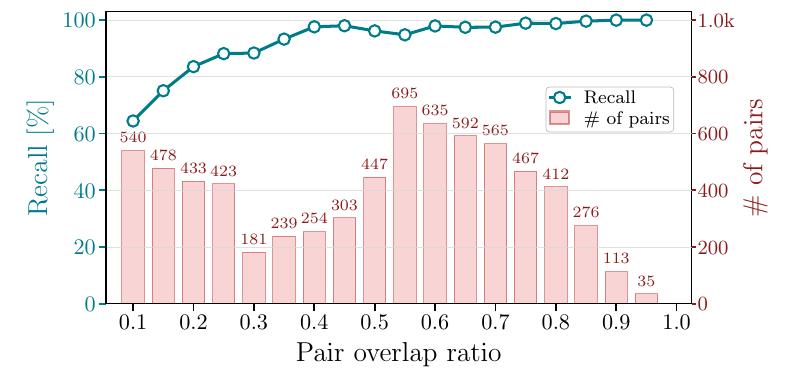}
    \caption{\rev{Registration recall versus measured pair overlap for pooled indoor benchmarks (\ThreeDMatch, \ThreeDLoMatch, \ScanNetppi, and \ScanNetppF). Bars show the number of evaluated pairs in each overlap bin.}}
    \label{fig:low_overlap_recall}
\end{figure}

\subsection{\texorpdfstring{\rev{Overlap-conditioned registration performance}}{Overlap-conditioned registration performance}}
\label[suppsection]{app:trade_off}

\rev{We further analyze low-overlap behavior on indoor benchmarks with measured pair-overlap statistics.
\Cref{fig:low_overlap_recall} pools the RMSE-evaluated \ThreeDMatch/\ThreeDLoMatch pairs and the default \ScanNetppi/\ScanNetppF pairs, then computes recall within fixed overlap intervals.
The line reports recall, while the bars show the number of evaluated pairs in each interval.
The trend shows that registration failures concentrate in the lowest-overlap regime, rather than being specific to a particular benchmark.}

\newcommand{\lowOverlapQualW}{0.270}
\newcommand{\lowOverlapQualImg}[1]{%
  \begin{minipage}[b]{\lowOverlapQualW\textwidth}
    #1
  \end{minipage}%
}
\newcommand{\lowOverlapQualImgCap}[2]{%
  \begin{minipage}[b]{\lowOverlapQualW\textwidth}
    #1\par\vspace{0.5mm}
    {\centering\small\sffamily #2\par}
  \end{minipage}%
}

\subsection{\texorpdfstring{Ambiguity under severe partial overlap}{Ambiguity under severe partial overlap}}
\label[suppsection]{app:ambiguity}

\begin{figure}[t!]
  \centering
  \subfloat[]{%
    \includegraphics[width=0.23\textwidth]{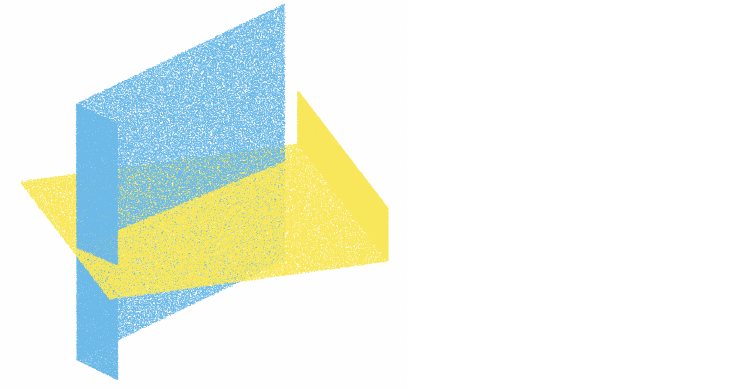}%
    \label{fig:global_optim_a}%
  }\hfill
  \subfloat[]{%
    \includegraphics[width=0.23\textwidth]{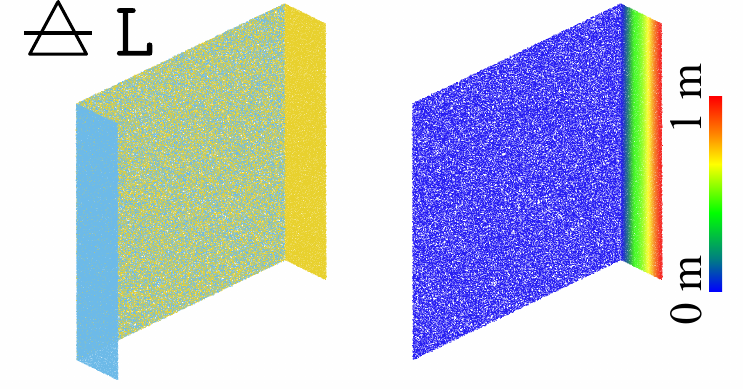}%
    \label{fig:global_optim_b}%
  }\hfill
  \subfloat[]{%
    \includegraphics[width=0.23\textwidth]{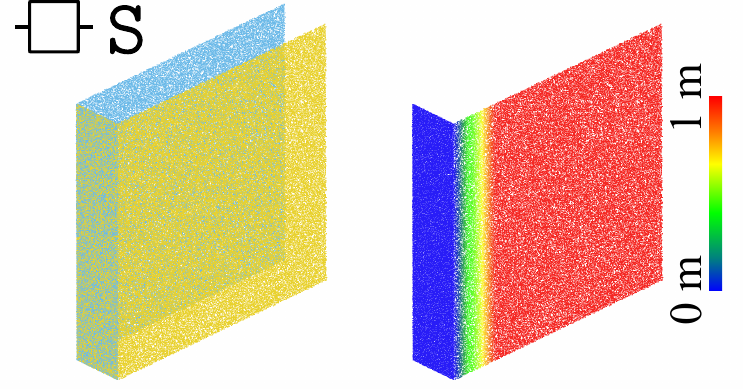}%
    \label{fig:global_optim_c}%
  }\hfill
  \subfloat[]{%
    \includegraphics[width=0.23\textwidth]{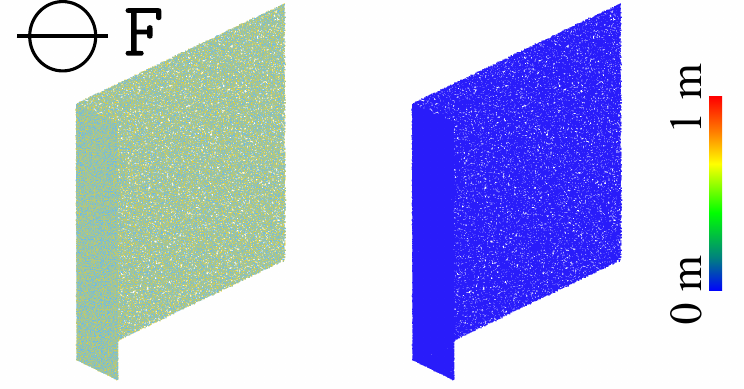}%
    \label{fig:global_optim_d}%
  }

  \vspace{1mm}

  \subfloat[]{%
    \includegraphics[width=0.23\textwidth]{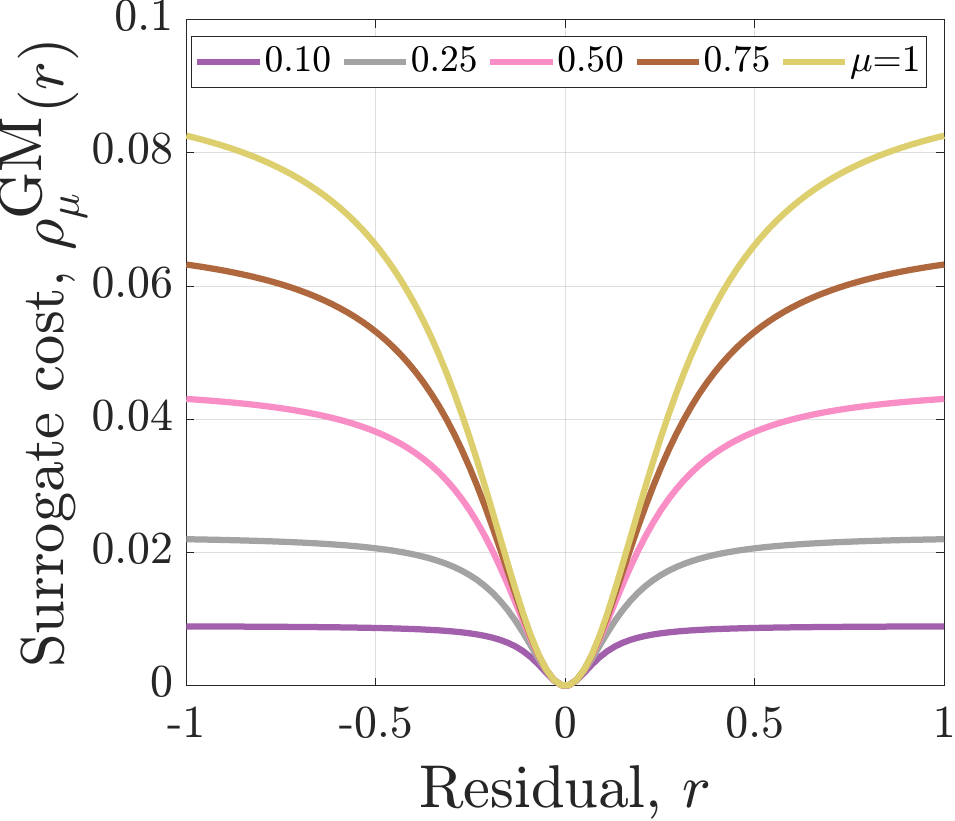}%
    \label{fig:global_optim_e}%
  }\hfill
  \subfloat[]{%
    \includegraphics[width=0.23\textwidth]{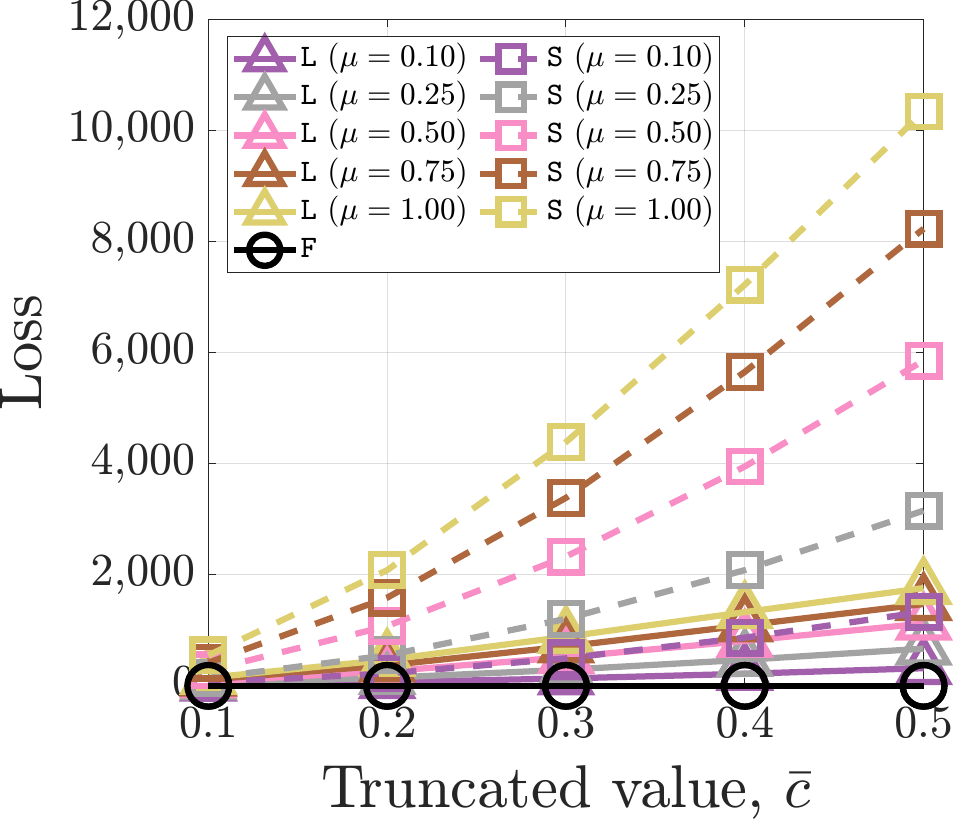}%
    \label{fig:global_optim_f}%
  }\hfill
  \subfloat[]{%
    \includegraphics[width=0.23\textwidth]{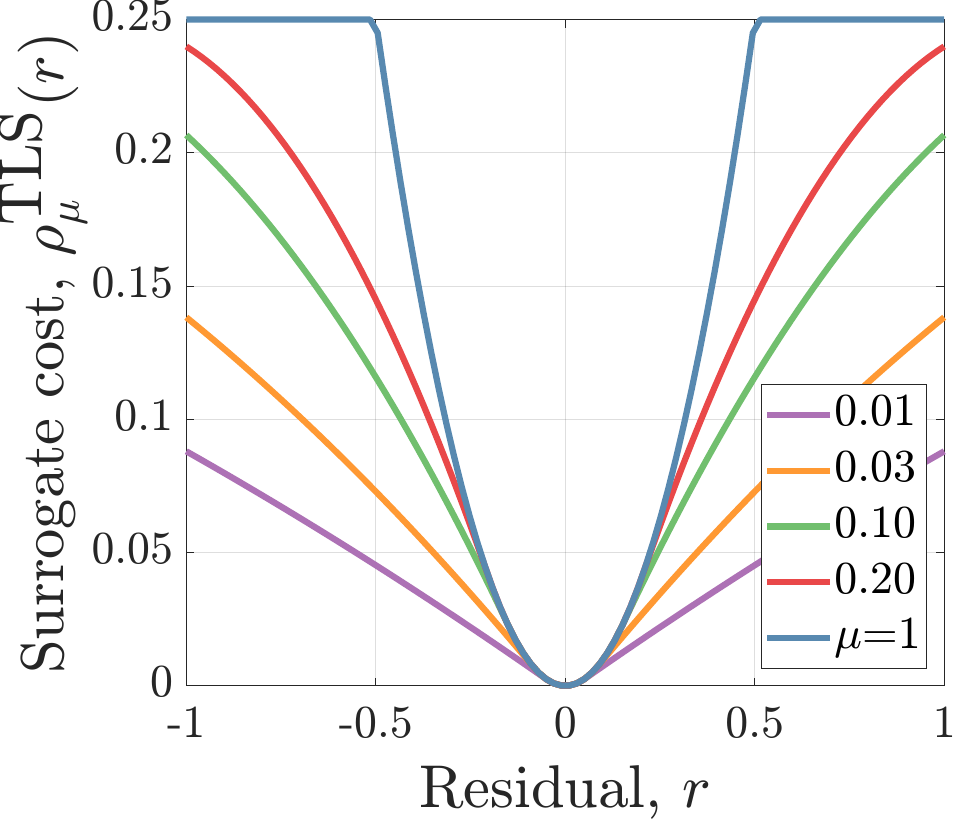}%
    \label{fig:global_optim_g}%
  }\hfill
  \subfloat[]{%
    \includegraphics[width=0.23\textwidth]{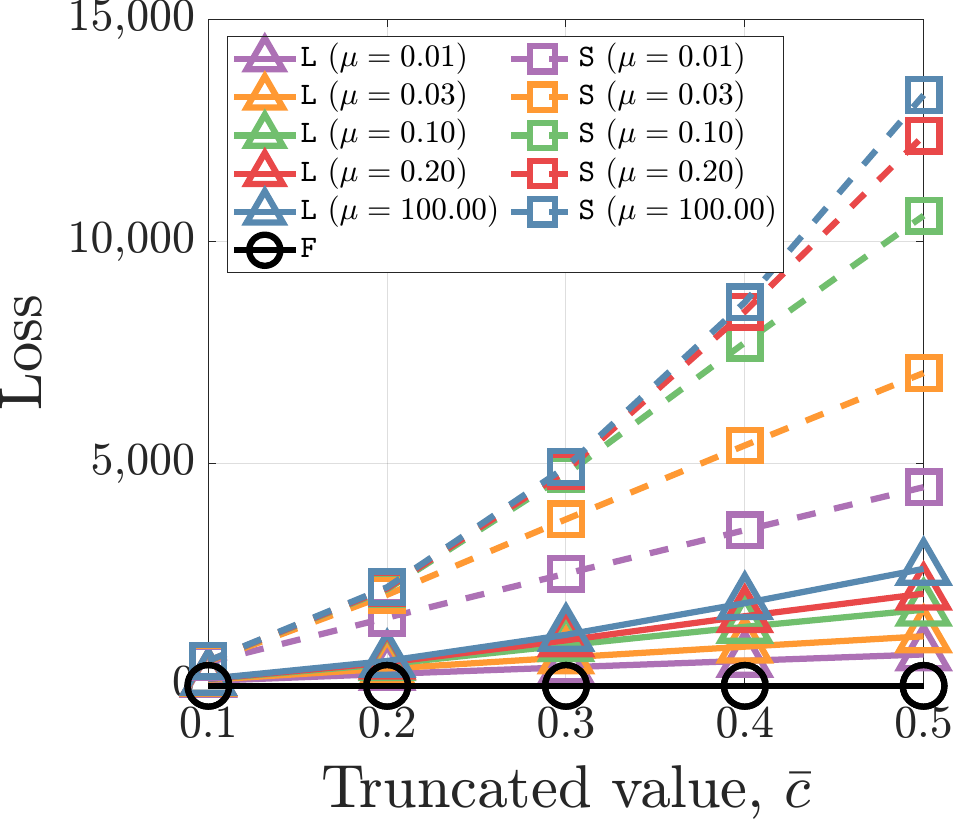}%
    \label{fig:global_optim_h}%
  }
  \caption{Illustration of the \textit{global optimum ambiguity} problem in point cloud registration.
  (a)~Two misaligned L-shaped point clouds and (b)--(d)~possible alignments: overlap along the longer segment (\texttt{L}), overlap along the shorter segment (\texttt{S}), and full overlap (\texttt{F}), respectively.
  (e)~German--McClure (GM) surrogate costs at different control parameters $\mu$ and (f)~the corresponding loss values.
  (g)~Truncated least-squares (TLS) surrogate costs and (h)~their corresponding loss values.
  The fully overlapping case produces the lowest loss for all tested kernel shapes.}
  \label{fig:global_optim}
\end{figure}

We next explain why the consensus-maximization objective can fail under severe partial overlap.
Consider two L-shaped point clouds, as shown in \Cref{fig:global_optim}(a).
Depending on which structures are shared, plausible alignments include overlap along the longer segment (\Cref{fig:global_optim}(b)), overlap along the shorter segment (\Cref{fig:global_optim}(c)), and full overlap (\Cref{fig:global_optim}(d)).
Although full overlap produces the lowest residual when all geometry is observed, either partial alignment may represent the correct pose when only a limited region is shared.
Consequently, the transformation minimizing a geometric cost is not necessarily the ground-truth transformation in a severely partial observation.
We refer to this phenomenon as \textit{global optimum ambiguity}.

This ambiguity cannot be eliminated solely by selecting a different non-convex robust kernel.
We illustrate this using the German--McClure (GM) and truncated least-squares (TLS) surrogate costs employed by graduated non-convexity~\cite{Yang20ral-GNC}.
Let $r$ denote a residual, $\bar{c}$ the truncation parameter, and $\mu$ the control parameter governing the non-convexity.
The GM surrogate is
\begin{equation}
\rho^\text{GM}_\mu(r)=\frac{\mu \bar{c}^2 r^2}{\mu \bar{c}^2+r^2},
\end{equation}
and the TLS surrogate is
\begin{equation}
\resizebox{\columnwidth}{!}{$
\rho^\text{TLS}_\mu(r)=
\begin{cases}
r^2, &
r^2 \in \left[0,\frac{\mu}{\mu+1}\bar{c}^2\right],\\
2\bar{c}|r|\sqrt{\mu(\mu+1)}-\mu(\bar{c}^2+r^2), &
r^2 \in \left[\frac{\mu}{\mu+1}\bar{c}^2,
\frac{\mu+1}{\mu}\bar{c}^2\right],\\
\bar{c}^2, &
r^2 \in \left[\frac{\mu+1}{\mu}\bar{c}^2,+\infty\right).
\end{cases}
$}
\end{equation}

As shown in \Cref{fig:global_optim}(e)--(h), varying $\mu$ changes the shape of each surrogate but does not change the ordering of the candidate alignments: the fully overlapping configuration retains the lowest loss.
Thus, when the true observation is only partially overlapping, robustification alone cannot identify which geometrically plausible alignment corresponds to the physical relative pose.

\begin{figure*}[t!]
  \centering
  \setlength{\arrayrulewidth}{0.3pt}
  \arrayrulecolor[gray]{0.7}
  \begin{tabular}{c c c c}
    \multirow{4}{*}{\makebox[0pt][r]{\raisebox{-2.00cm}[0pt][0pt]{\rotatebox[origin=c]{90}{Failure cases}}\hspace{-0.1cm}}}
    & \lowOverlapQualImg{\includegraphics[width=\linewidth]{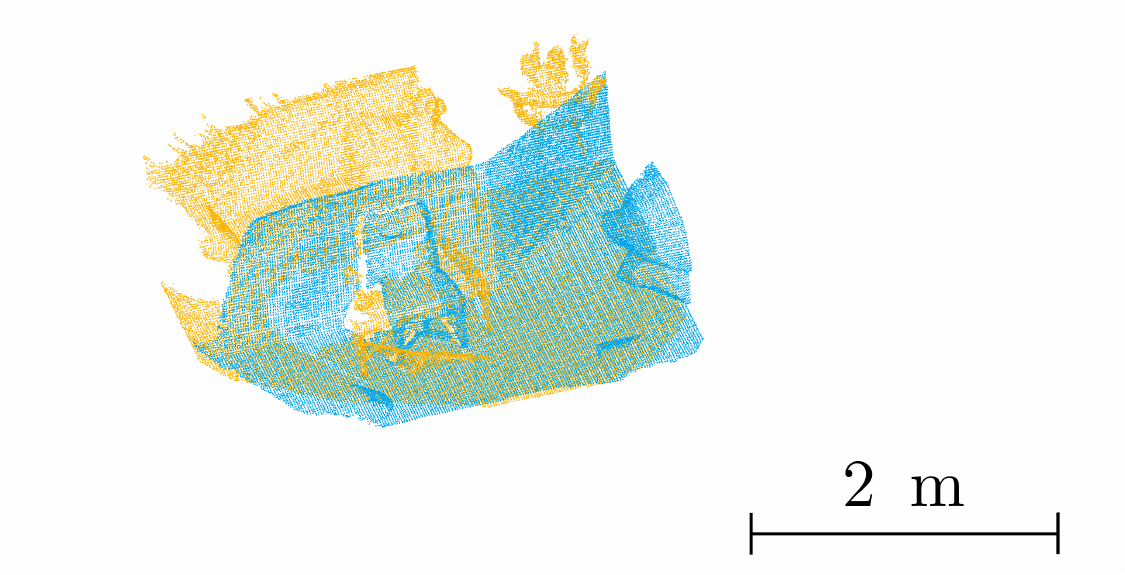}}
    & \lowOverlapQualImg{\includegraphics[width=\linewidth]{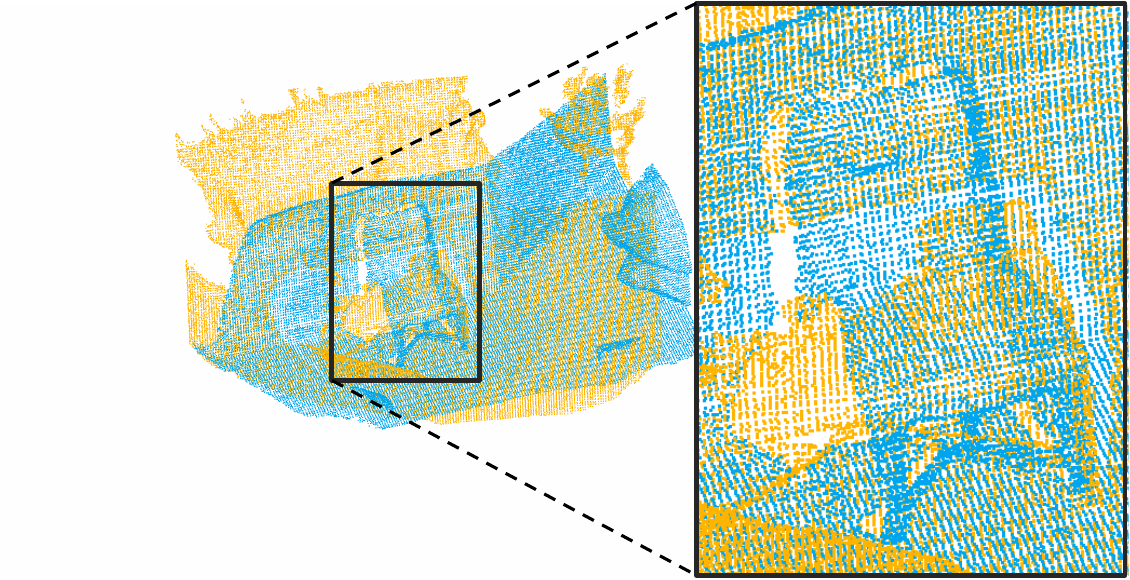}}
    & \lowOverlapQualImg{\includegraphics[width=\linewidth]{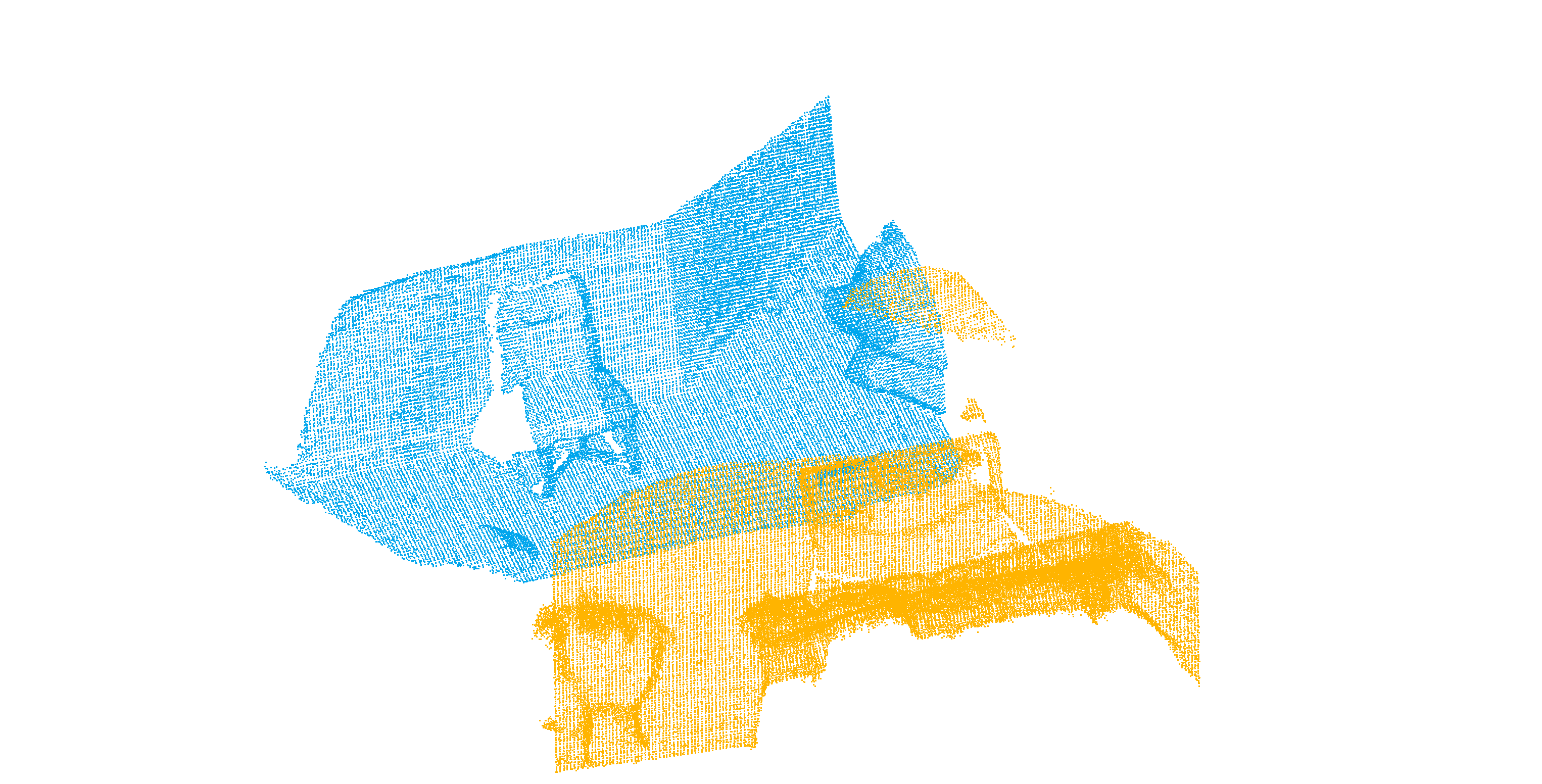}} \\
    & \lowOverlapQualImg{\includegraphics[width=\linewidth]{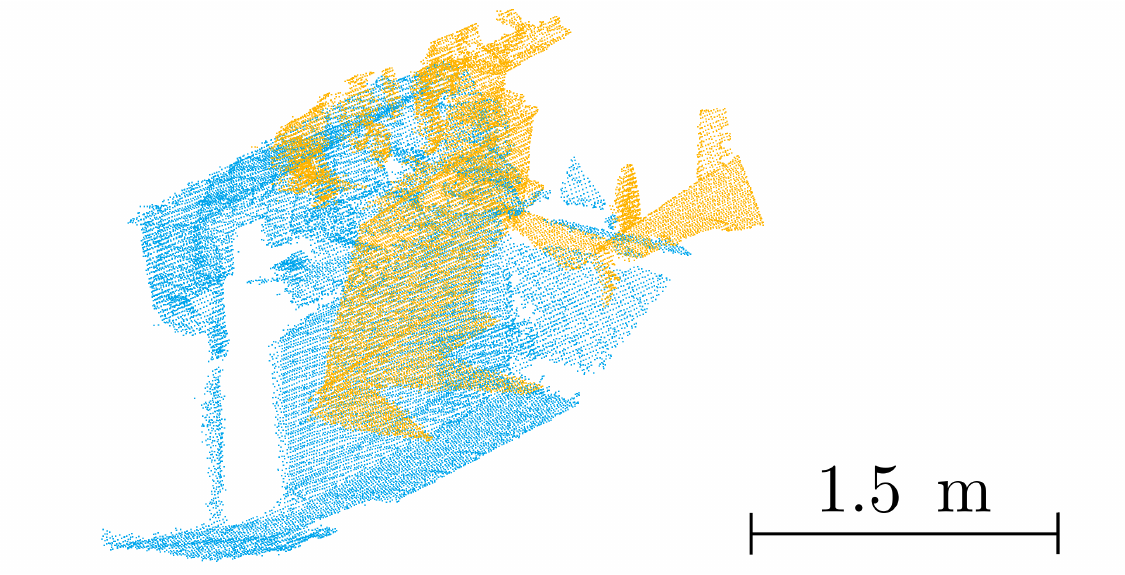}}
    & \lowOverlapQualImg{\includegraphics[width=\linewidth]{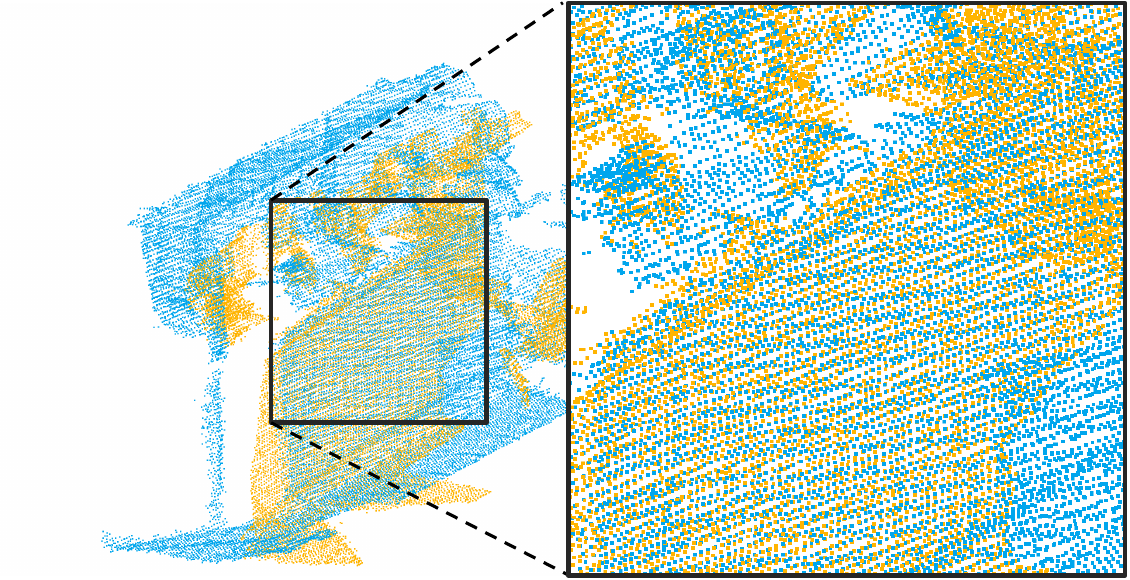}}
    & \lowOverlapQualImg{\includegraphics[width=\linewidth]{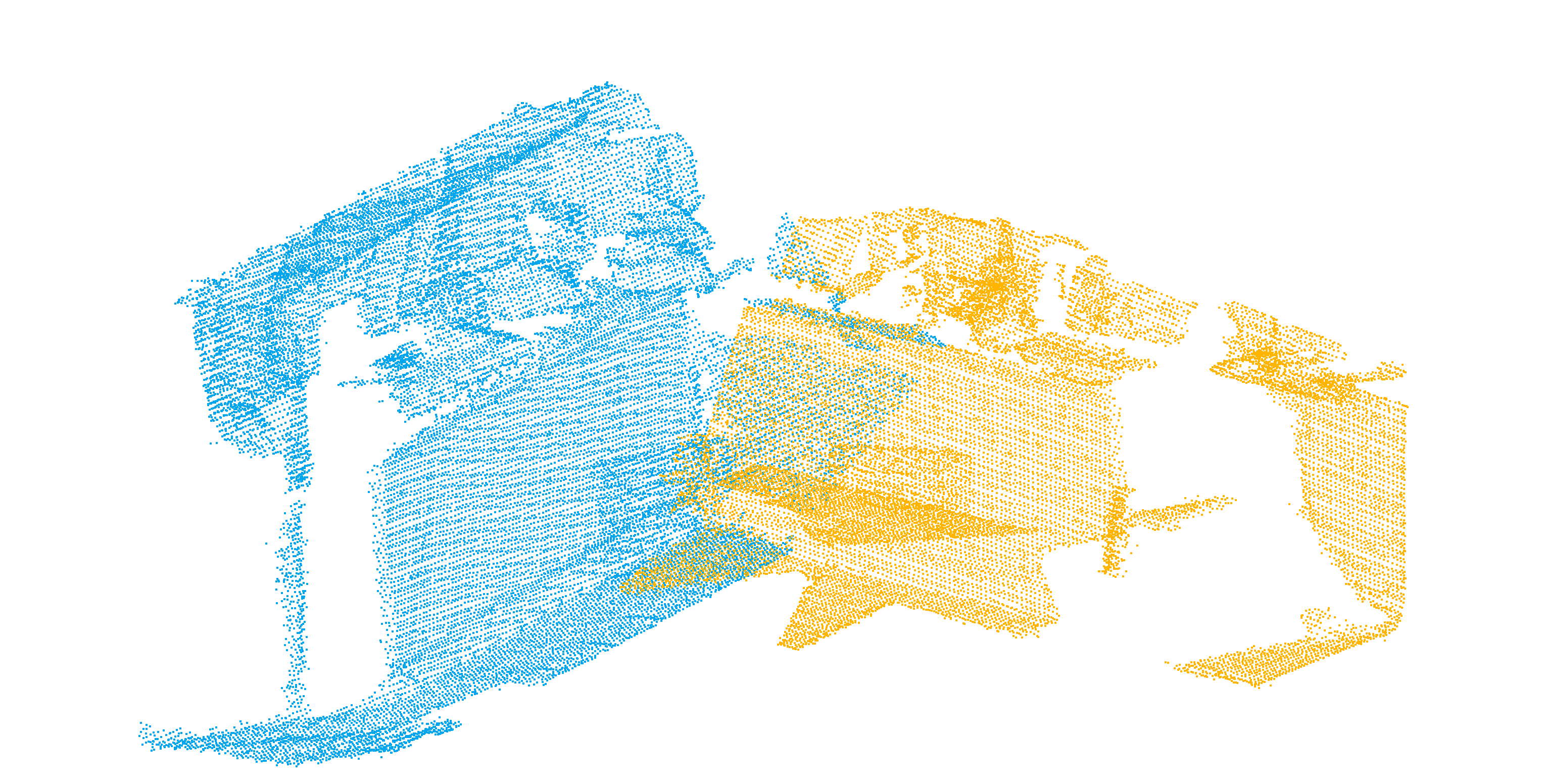}} \\
    & \lowOverlapQualImg{\includegraphics[width=\linewidth]{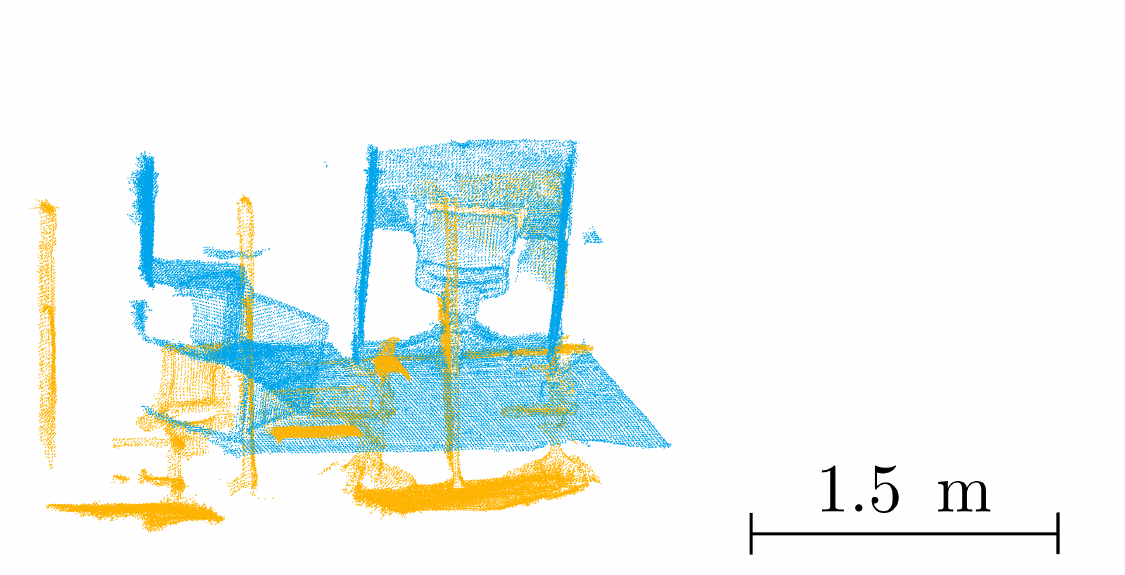}}
    & \lowOverlapQualImg{\includegraphics[width=\linewidth]{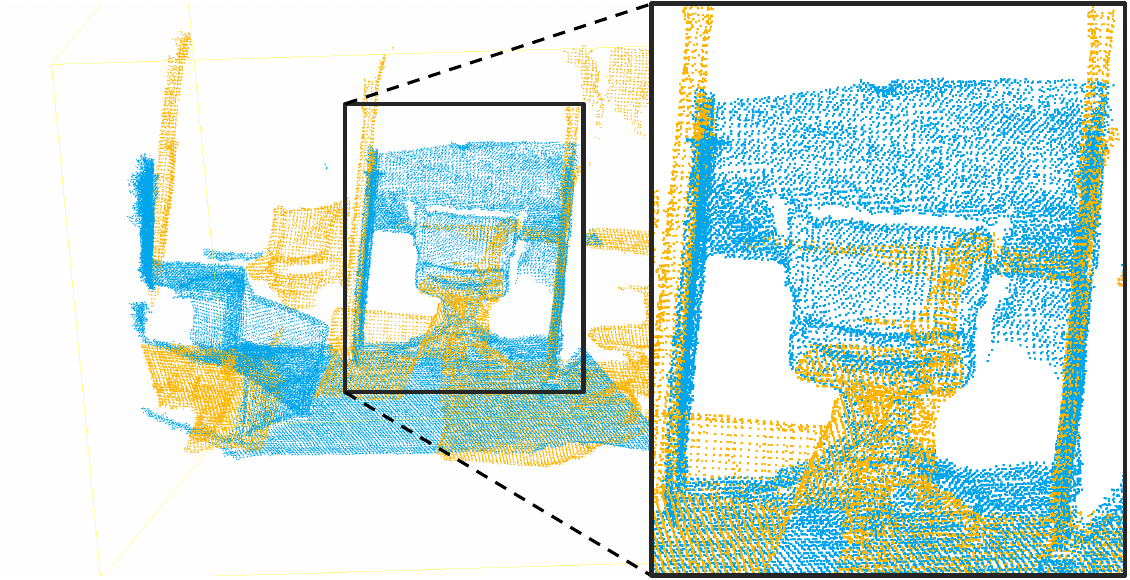}}
    & \lowOverlapQualImg{\includegraphics[width=\linewidth]{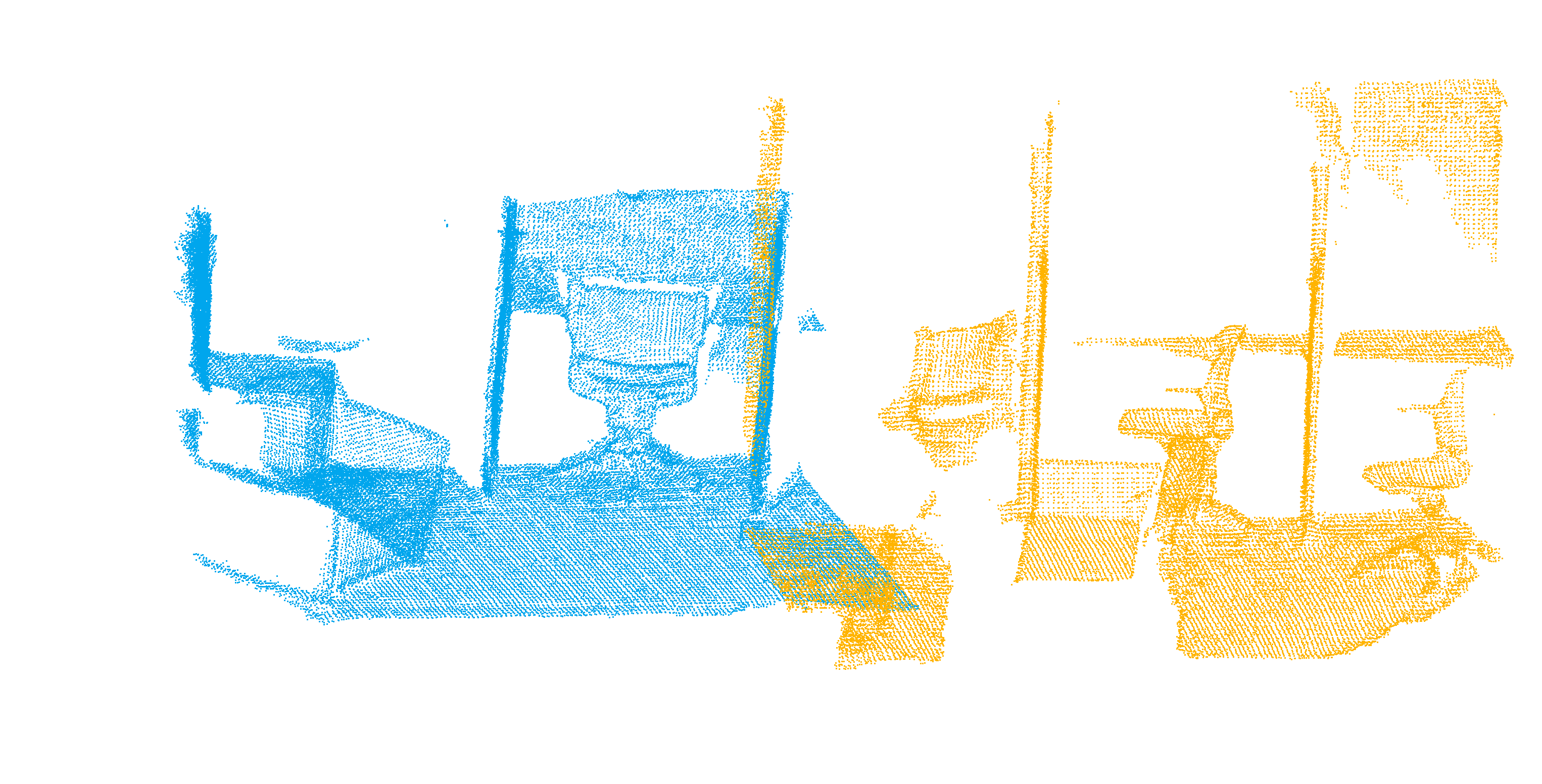}} \\
    & \lowOverlapQualImg{\includegraphics[width=\linewidth]{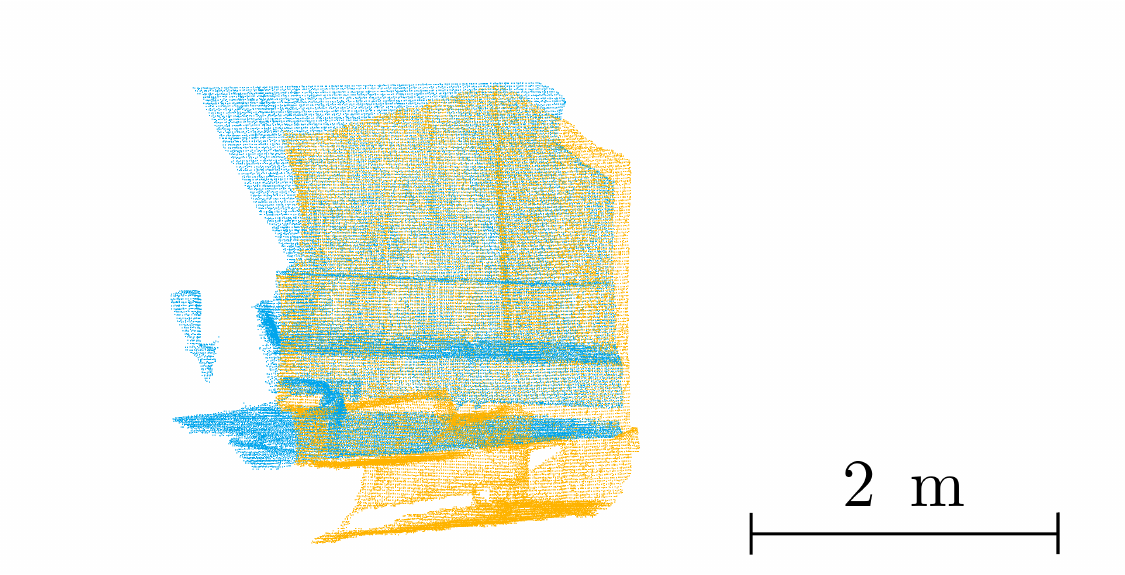}}
    & \lowOverlapQualImg{\includegraphics[width=\linewidth]{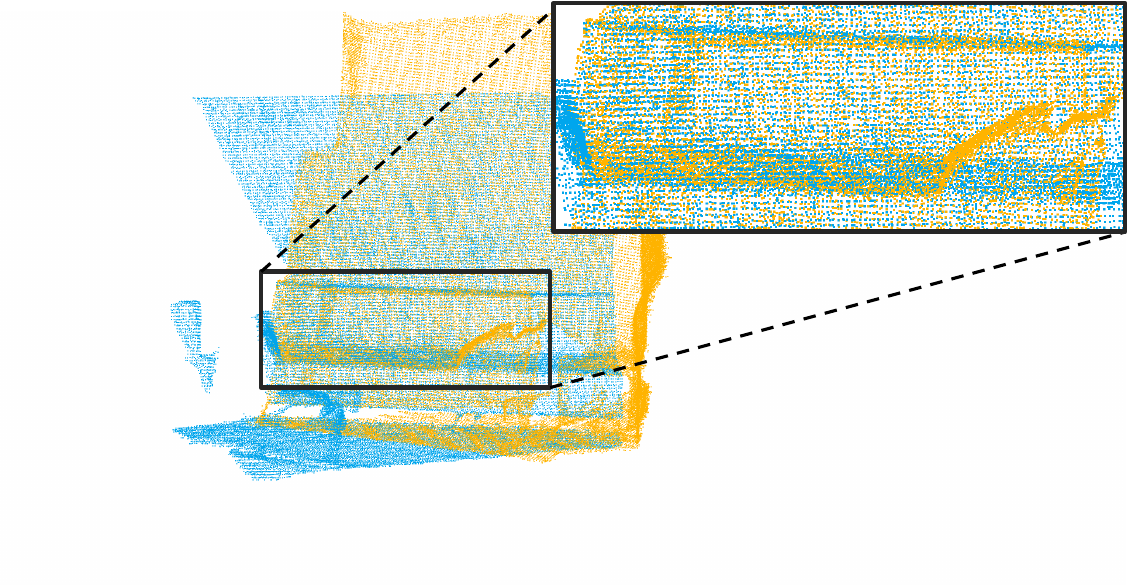}}
    & \lowOverlapQualImg{\includegraphics[width=\linewidth]{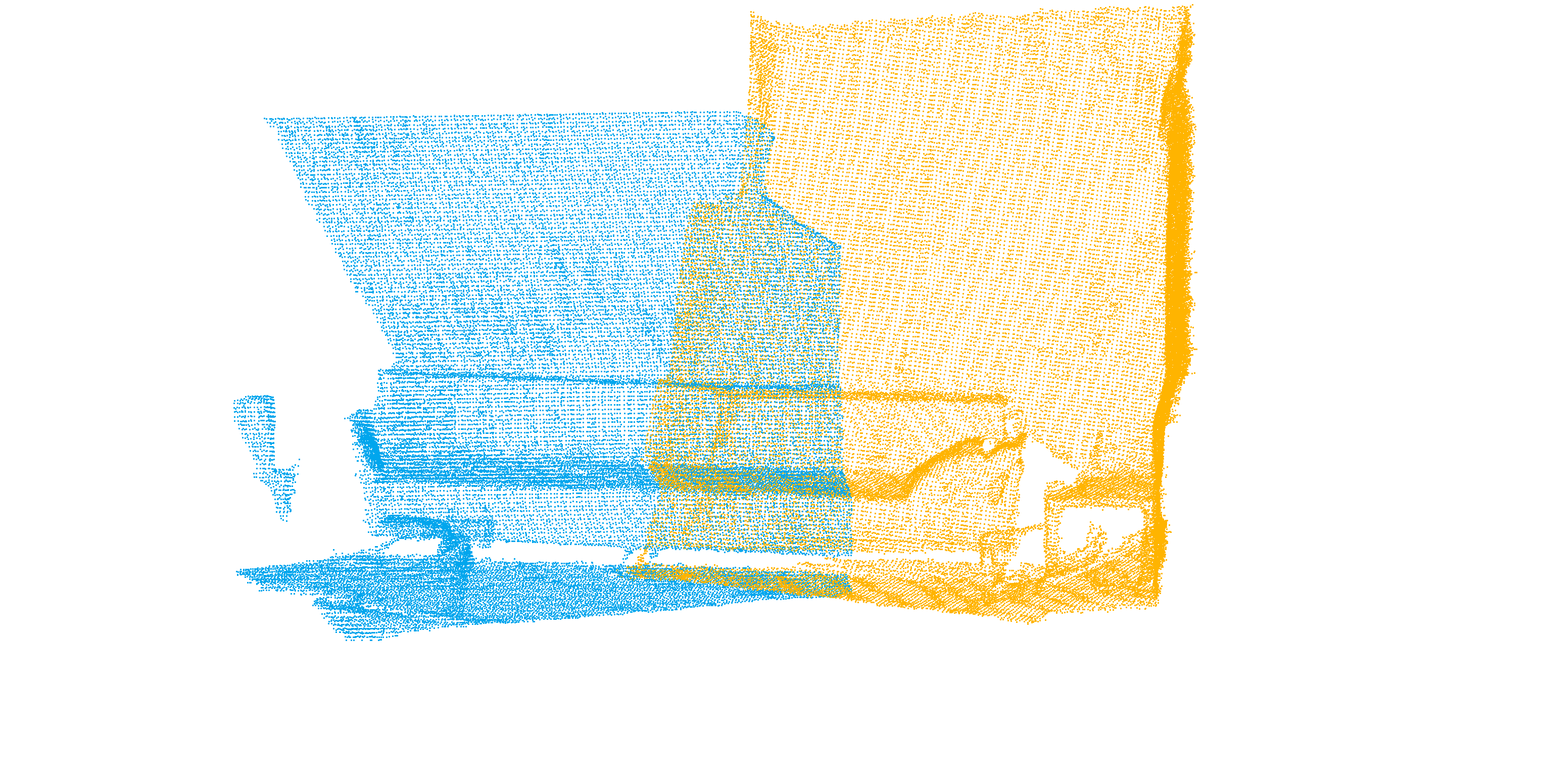}} \\
    \midrule
    \multirow{2}{*}{\makebox[0pt][r]{\raisebox{0pt}[0pt][0pt]{\rotatebox[origin=c]{90}{Success cases}}\hspace{-0.1cm}}}
    & \lowOverlapQualImg{\includegraphics[width=\linewidth]{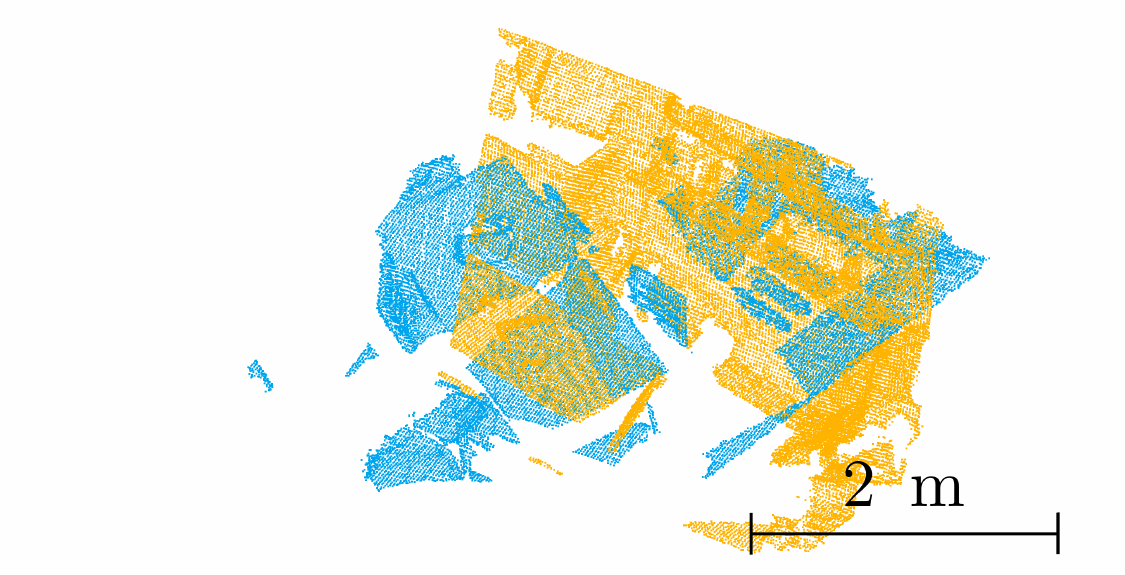}}
    & \lowOverlapQualImg{\includegraphics[width=\linewidth]{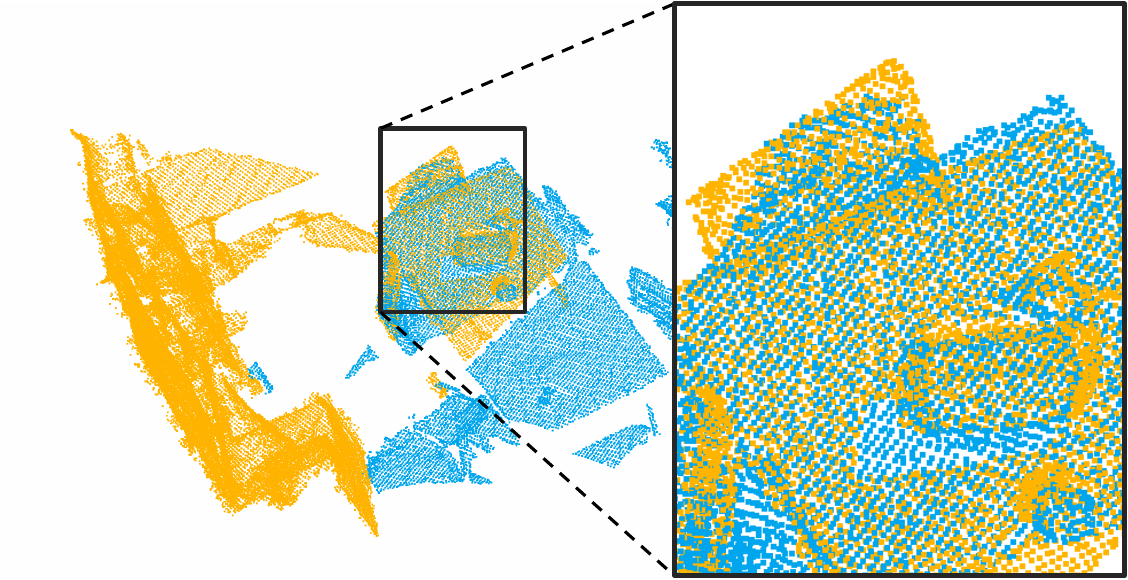}}
    & \lowOverlapQualImg{\includegraphics[width=\linewidth]{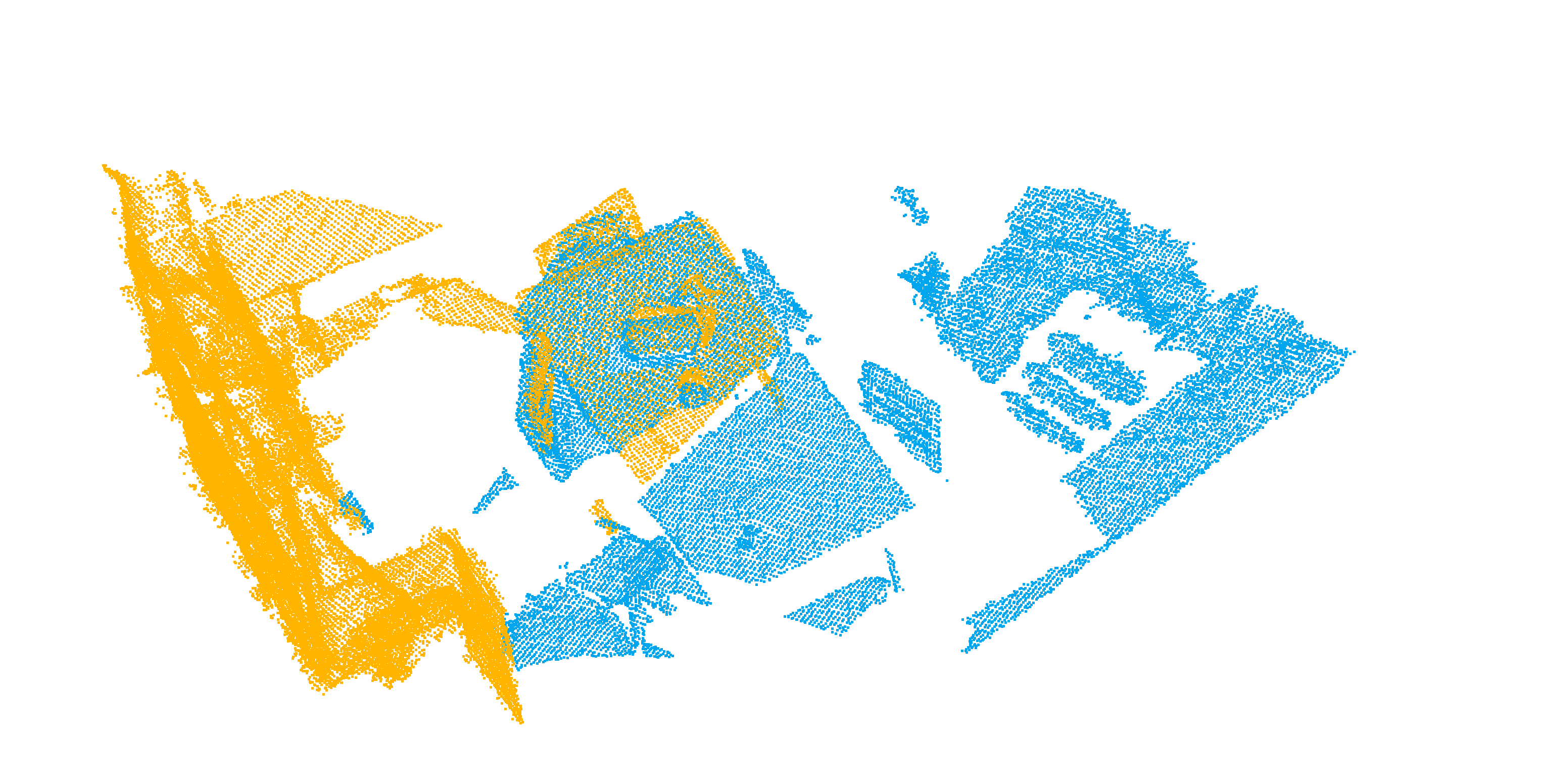}} \\
    & \lowOverlapQualImgCap{\includegraphics[width=\linewidth]{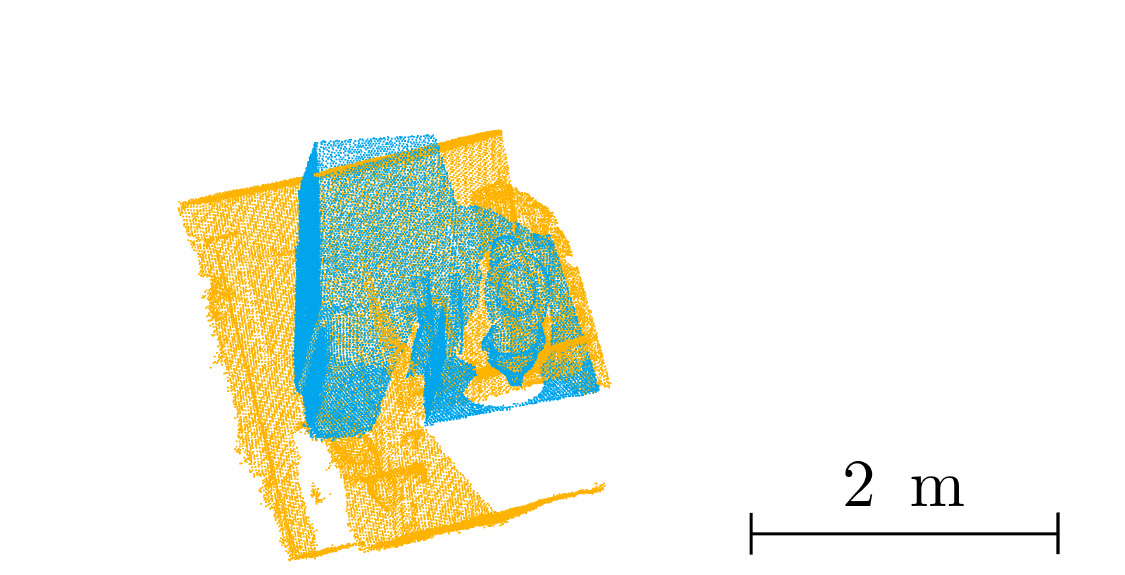}}{(a) Source \& Target}
    & \lowOverlapQualImgCap{\includegraphics[width=\linewidth]{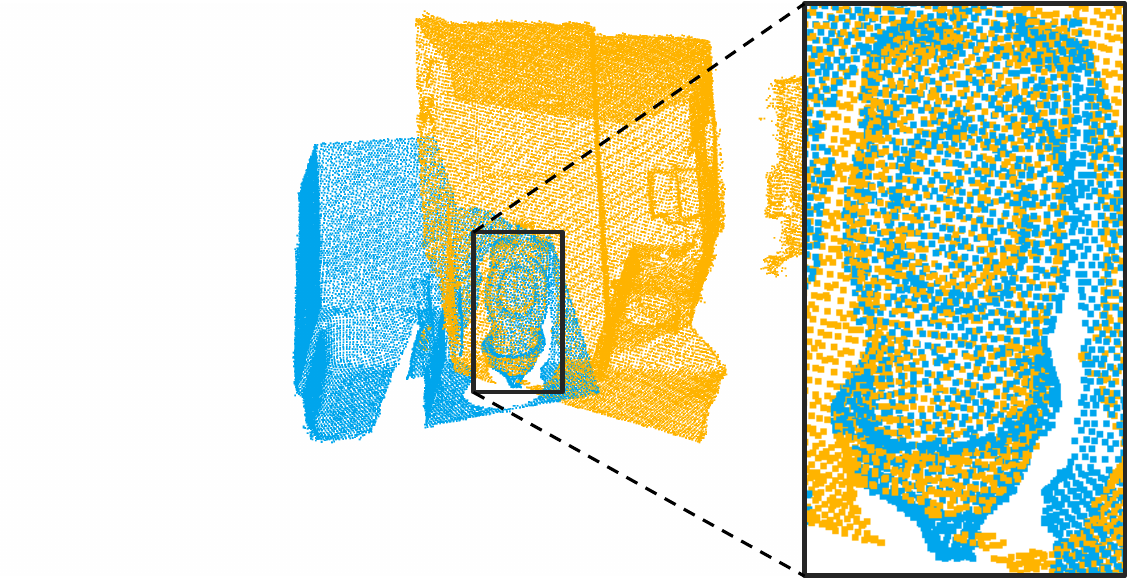}}{(b) Our \oursname}
    & \lowOverlapQualImgCap{\includegraphics[width=\linewidth]{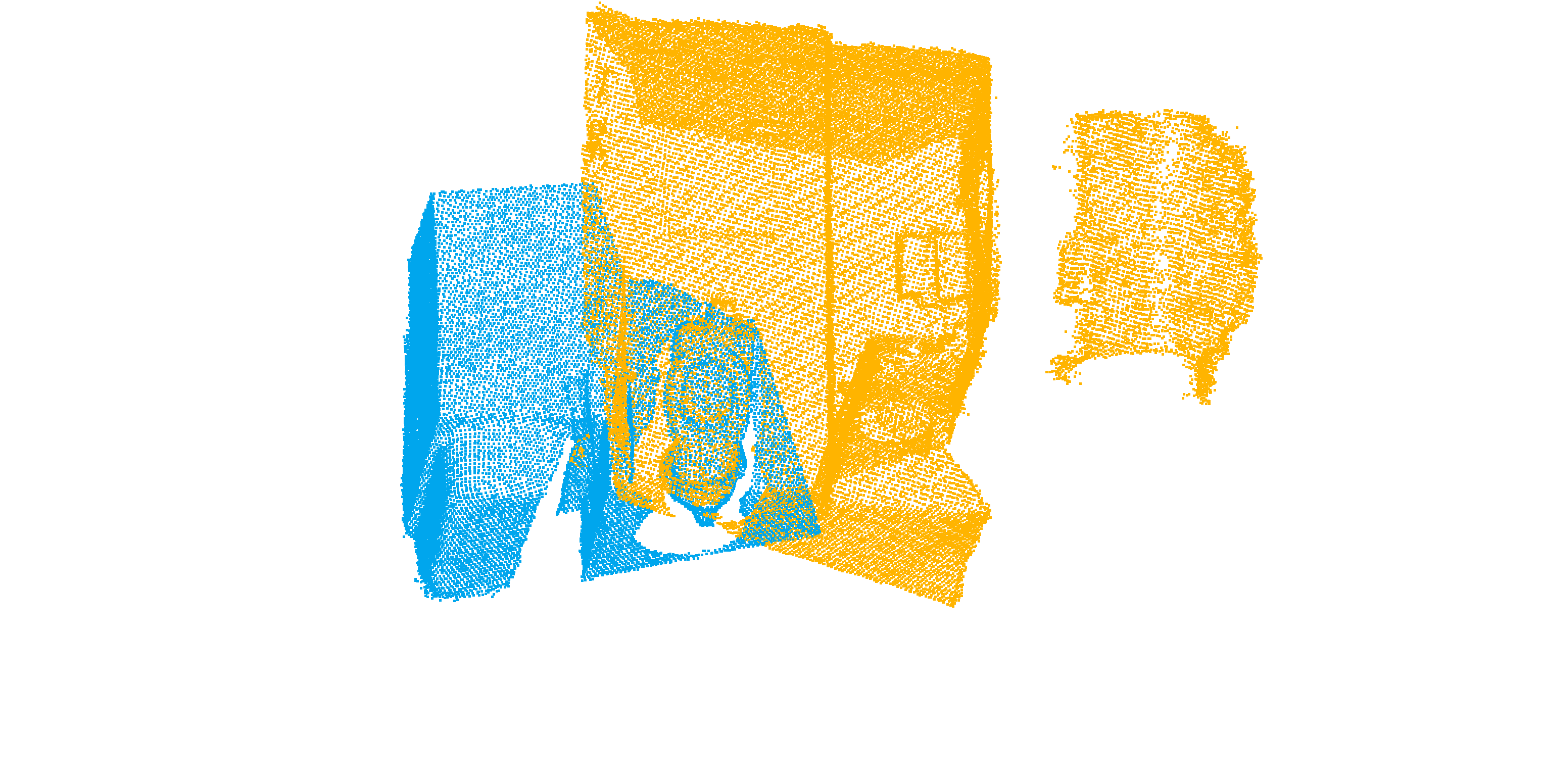}}{(c) GT}
  \end{tabular}
  \setlength{\arrayrulewidth}{0.4pt}
  \arrayrulecolor{black}
  \caption{\rev{Qualitative low-overlap registration examples on \ThreeDLoMatch. The top four rows show the behavior of \oursname in failed cases, where the prediction can favor a geometrically plausible but incorrect high-overlap alignment. The bottom two rows show successful low-overlap cases, where distinctive anchor objects or structures provide sufficient geometric evidence for correct registration.}}
  \label{fig:low_overlap_qualitative}
\end{figure*}

\rev{This interpretation is further supported by the \ThreeDLoMatch registration examples in \Cref{fig:low_overlap_qualitative}.
In the failure cases, \oursname can align a geometrically plausible structure that yields strong local consistency but does not correspond to the ground-truth shared region, which is consistent with the global-optimum ambiguity described above.
In contrast, the successful cases contain distinctive anchor structures that make the correct partial overlap sufficiently informative.
These results support our claim that the low-overlap limitation is tied to the ambiguity and pose-informativeness of the shared geometry, rather than to a dataset-specific failure alone.}

\subsection{\texorpdfstring{\rev{Detailed parameter sensitivity results}}{Detailed parameter sensitivity results}}

\rev{The full per-dataset recall values underlying Fig.~12 are reported in Table~S2.
We additionally provide the outdoor-only analysis of the radius cap $r_\mathrm{max}$ in Table~S3.}

\providecommand{\betterperf}[1]{\textcolor{green!50!black}{#1}}
\providecommand{\lowerperf}[1]{\textcolor{red}{#1}}
\begingroup
\begin{table*}[t!]
    \centering
    \setlength{\tabcolsep}{1pt}
    \caption{\rev{Full per-dataset recalls (\%) for the one-factor parameter sweeps in Fig.~12. Shaded rows denote default settings. Red numbers indicate a decrease from the default parameter setting, while green numbers indicate an increase.}}
    \label{table:sens-general}
    {\scriptsize
    \resizebox{\linewidth}{!}{%
    \begin{tabular}{l|cccccccccccc|c}
        \toprule
        & \multicolumn{1}{c}{Object} & \multicolumn{5}{c}{Indoor} & \multicolumn{6}{c|}{Outdoor} & \multirow{2}{*}{Avg.} \\
        \cmidrule(lr){2-2} \cmidrule(lr){3-7} \cmidrule(lr){8-13}
        Setting & \ModelNet & \ThreeDMatch & \ThreeDLoMatch & \ScanNetppi & \ScanNetppF & \TIERS & \KITTI & \WOD & \KAIST & \MIT & \ETH & \Oxford &  \\ \midrule
        \multicolumn{14}{l}{\textit{$N_\mathrm{FPS}$}} \\
        500 & \lowerperf{99.13} & \lowerperf{93.67} & \lowerperf{69.54} & \lowerperf{94.07} & \lowerperf{99.36} & \lowerperf{92.30} & \lowerperf{99.10} & 100.00 & \lowerperf{93.02} & \lowerperf{93.48} & \lowerperf{89.76} & \lowerperf{97.67} & 93.43~{\tiny \lowerperf{(-2.71\,\%p)}} \\
        1,000 & \lowerperf{99.45} & \lowerperf{94.36} & \lowerperf{72.33} & \betterperf{95.03} & \lowerperf{99.80} & \lowerperf{92.76} & 99.82 & 100.00 & \lowerperf{98.79} & \betterperf{98.70} & \lowerperf{99.02} & 99.67 & 95.81~{\tiny \lowerperf{(-0.33\,\%p)}} \\
        \rowcolor{gray!10} 1,500 & 99.84 & 95.58 & 74.18 & 94.99 & 99.90 & 93.45 & 99.82 & 100.00 & 99.15 & 97.39 & 99.72 & 99.67 & 96.14 \\
        2,000 & \lowerperf{99.45} & \lowerperf{94.79} & \lowerperf{72.68} & \betterperf{95.56} & \betterperf{99.95} & \lowerperf{91.95} & 99.82 & 100.00 & \betterperf{99.20} & \betterperf{97.83} & \betterperf{100.00} & \lowerperf{99.34} & 95.88~{\tiny \lowerperf{(-0.26\,\%p)}} \\
        2,500 & \lowerperf{99.61} & \lowerperf{94.86} & \lowerperf{73.32} & \betterperf{95.42} & 99.90 & \lowerperf{91.15} & 99.82 & 100.00 & \betterperf{99.30} & \betterperf{98.26} & \betterperf{100.00} & 99.67 & 95.94~{\tiny \lowerperf{(-0.20\,\%p)}} \\
        3,000 & \lowerperf{99.68} & \lowerperf{95.31} & \lowerperf{73.48} & \betterperf{95.27} & \betterperf{99.95} & \lowerperf{92.07} & 99.82 & 100.00 & \betterperf{99.45} & \betterperf{98.26} & \betterperf{100.00} & \lowerperf{99.34} & 96.05~{\tiny \lowerperf{(-0.09\,\%p)}} \\
        \midrule
        \multicolumn{14}{l}{\textit{$N_\mathrm{patch}$}} \\
        128 & \lowerperf{99.76} & \lowerperf{93.52} & \lowerperf{70.06} & \betterperf{95.03} & \lowerperf{99.65} & \lowerperf{92.87} & 99.82 & 100.00 & \lowerperf{99.10} & \betterperf{98.26} & \lowerperf{99.58} & \lowerperf{99.34} & 95.58~{\tiny \lowerperf{(-0.56\,\%p)}} \\
        256 & \lowerperf{99.61} & \lowerperf{93.32} & \lowerperf{72.51} & \betterperf{95.71} & \lowerperf{99.80} & \betterperf{93.56} & 99.82 & 100.00 & \betterperf{99.20} & \betterperf{98.26} & \betterperf{100.00} & \lowerperf{99.34} & 95.93~{\tiny \lowerperf{(-0.21\,\%p)}} \\
        \rowcolor{gray!10} 512 & 99.84 & 95.58 & 74.18 & 94.99 & 99.90 & 93.45 & 99.82 & 100.00 & 99.15 & 97.39 & 99.72 & 99.67 & 96.14 \\
        768 & \lowerperf{99.76} & \lowerperf{94.85} & \lowerperf{72.95} & \betterperf{95.32} & 99.90 & \lowerperf{92.53} & 99.82 & 100.00 & \betterperf{99.25} & \betterperf{97.83} & \lowerperf{99.58} & \lowerperf{99.34} & 95.93~{\tiny \lowerperf{(-0.21\,\%p)}} \\
        1,024 & \lowerperf{99.68} & \lowerperf{94.73} & \lowerperf{72.14} & \betterperf{95.47} & 99.90 & \lowerperf{92.76} & 99.82 & 100.00 & \betterperf{99.25} & \betterperf{98.26} & \betterperf{99.86} & \lowerperf{99.34} & 95.93~{\tiny \lowerperf{(-0.21\,\%p)}} \\
        \midrule
        \multicolumn{14}{l}{\textit{Scales}} \\
        $r_l$ & \lowerperf{99.05} & \lowerperf{85.84} & \lowerperf{55.70} & \lowerperf{84.81} & 99.90 & \lowerperf{89.31} & \lowerperf{92.43} & 100.00 & \lowerperf{95.43} & \lowerperf{88.26} & \lowerperf{97.34} & \lowerperf{98.01} & 90.51~{\tiny \lowerperf{(-5.63\,\%p)}} \\
        $r_m$ & 99.84 & \lowerperf{93.38} & \lowerperf{71.69} & \lowerperf{93.10} & \lowerperf{99.60} & \lowerperf{90.80} & 99.82 & 100.00 & \lowerperf{99.05} & \lowerperf{95.65} & \lowerperf{99.30} & \lowerperf{99.34} & 95.13~{\tiny \lowerperf{(-1.01\,\%p)}} \\
        $r_g$ & \lowerperf{99.45} & \lowerperf{92.96} & \lowerperf{69.44} & \betterperf{95.08} & \lowerperf{99.26} & \lowerperf{92.87} & 99.82 & 100.00 & \betterperf{99.25} & \betterperf{98.26} & \lowerperf{98.04} & \lowerperf{99.34} & 95.31~{\tiny \lowerperf{(-0.83\,\%p)}} \\
        $r_l+r_m$ & \lowerperf{99.61} & \lowerperf{94.65} & \lowerperf{70.64} & \lowerperf{94.17} & \betterperf{100.00} & \lowerperf{90.92} & 99.82 & 100.00 & \lowerperf{99.00} & \lowerperf{95.22} & 99.72 & \lowerperf{99.34} & 95.26~{\tiny \lowerperf{(-0.88\,\%p)}} \\
        $r_l+r_g$ & \lowerperf{99.68} & \lowerperf{94.25} & \lowerperf{70.53} & \betterperf{95.18} & \lowerperf{99.75} & \lowerperf{92.41} & 99.82 & 100.00 & \betterperf{99.20} & \betterperf{98.26} & 99.72 & \lowerperf{99.34} & 95.68~{\tiny \lowerperf{(-0.46\,\%p)}} \\
        $r_m+r_g$ & \lowerperf{99.68} & \lowerperf{94.91} & \lowerperf{72.21} & \betterperf{95.18} & \lowerperf{99.70} & \lowerperf{93.22} & 99.82 & 100.00 & 99.15 & \betterperf{99.13} & \betterperf{100.00} & 99.67 & 96.06~{\tiny \lowerperf{(-0.08\,\%p)}} \\
        \rowcolor{gray!10} $r_l+r_m+r_g$ & 99.84 & 95.58 & 74.18 & 94.99 & 99.90 & 93.45 & 99.82 & 100.00 & 99.15 & 97.39 & 99.72 & 99.67 & 96.14 \\
        \midrule
        \multicolumn{14}{l}{\textit{$100\times(\tau_l,\tau_m,\tau_g)$}} \\
        $(0.25,1.0,2.5)$ & \lowerperf{99.45} & \lowerperf{93.86} & \lowerperf{72.48} & \lowerperf{94.74} & 99.90 & \lowerperf{92.53} & 99.82 & 100.00 & \lowerperf{98.79} & \betterperf{97.83} & \betterperf{100.00} & \lowerperf{99.00} & 95.70~{\tiny \lowerperf{(-0.44\,\%p)}} \\
        $(0.375,1.5,4.0)$ & \lowerperf{99.68} & \lowerperf{94.53} & \lowerperf{73.47} & \betterperf{95.08} & \betterperf{99.95} & \lowerperf{92.87} & 99.82 & 100.00 & 99.15 & 97.39 & \betterperf{100.00} & \lowerperf{99.34} & 95.94~{\tiny \lowerperf{(-0.20\,\%p)}} \\
        \rowcolor{gray!10} $(0.5,2.0,5.0)$ & 99.84 & 95.58 & 74.18 & 94.99 & 99.90 & 93.45 & 99.82 & 100.00 & 99.15 & 97.39 & 99.72 & 99.67 & 96.14 \\
        $(0.5,2.5,6.0)$ & \lowerperf{99.68} & \lowerperf{94.02} & \lowerperf{72.55} & \betterperf{95.37} & \lowerperf{99.75} & \lowerperf{93.22} & 99.82 & 100.00 & \betterperf{99.35} & \betterperf{98.26} & 99.72 & 99.67 & 95.95~{\tiny \lowerperf{(-0.19\,\%p)}} \\
        $(0.75,3.0,7.5)$ & \lowerperf{99.61} & \lowerperf{92.38} & \lowerperf{71.69} & \betterperf{95.85} & \lowerperf{99.80} & \betterperf{94.48} & 99.82 & 100.00 & \betterperf{99.35} & \betterperf{98.26} & \betterperf{100.00} & \lowerperf{99.00} & 95.85~{\tiny \lowerperf{(-0.29\,\%p)}} \\
        \midrule
        \multicolumn{14}{l}{\textit{$(\kappa_\mathrm{spheric},\kappa_\mathrm{disc})$}} \\
        $(0.05,0.05)$ & \lowerperf{99.53} & \lowerperf{94.11} & \lowerperf{72.85} & \betterperf{96.62} & \lowerperf{99.55} & \lowerperf{93.33} & 99.82 & 100.00 & 99.15 & 97.39 & \betterperf{99.86} & 99.67 & 95.99~{\tiny \lowerperf{(-0.15\,\%p)}} \\
        $(0.10,0.05)$ & 99.84 & \lowerperf{94.52} & \lowerperf{73.54} & \betterperf{95.95} & \lowerperf{99.80} & \betterperf{93.56} & 99.82 & 100.00 & \betterperf{99.30} & 97.39 & 99.72 & \lowerperf{99.34} & 96.06~{\tiny \lowerperf{(-0.08\,\%p)}} \\
        $(0.10,0.10)$ & \lowerperf{99.37} & \lowerperf{94.64} & \lowerperf{73.05} & \betterperf{96.00} & 99.90 & \lowerperf{93.10} & 99.82 & 100.00 & \betterperf{99.30} & \betterperf{98.26} & \betterperf{99.86} & 99.67 & 96.08~{\tiny \lowerperf{(-0.06\,\%p)}} \\
        $(0.10,0.15)$ & 99.84 & \lowerperf{95.13} & \lowerperf{73.44} & \betterperf{95.85} & \lowerperf{99.80} & \lowerperf{93.10} & 99.82 & 100.00 & \betterperf{99.25} & \betterperf{98.26} & 99.72 & 99.67 & 96.16~{\tiny \betterperf{(+0.02\,\%p)}} \\
        \rowcolor{gray!10} $(0.15,0.10)$ & 99.84 & 95.58 & 74.18 & 94.99 & 99.90 & 93.45 & 99.82 & 100.00 & 99.15 & 97.39 & 99.72 & 99.67 & 96.14 \\
        $(0.15,0.15)$ & \lowerperf{99.61} & \lowerperf{94.28} & \lowerperf{72.98} & \betterperf{95.23} & \lowerperf{99.80} & \lowerperf{92.87} & 99.82 & 100.00 & \betterperf{99.20} & \betterperf{98.26} & \betterperf{99.86} & \lowerperf{99.34} & 95.94~{\tiny \lowerperf{(-0.20\,\%p)}} \\
        $(0.20,0.15)$ & \lowerperf{99.76} & \lowerperf{93.89} & \lowerperf{73.16} & \betterperf{95.37} & 99.90 & \lowerperf{92.64} & 99.82 & 100.00 & \lowerperf{99.10} & \betterperf{97.83} & 99.72 & \lowerperf{99.34} & 95.88~{\tiny \lowerperf{(-0.26\,\%p)}} \\
        $(0.20,0.20)$ & \lowerperf{99.61} & \lowerperf{94.66} & \lowerperf{72.84} & \betterperf{95.85} & 99.90 & \lowerperf{92.76} & 99.82 & 100.00 & \betterperf{99.20} & \betterperf{98.70} & \betterperf{99.86} & \lowerperf{99.34} & 96.05~{\tiny \lowerperf{(-0.09\,\%p)}} \\
        $(0.25,0.20)$ & \lowerperf{99.68} & \lowerperf{94.15} & \lowerperf{71.88} & \betterperf{95.03} & \betterperf{99.95} & \lowerperf{92.64} & 99.82 & 100.00 & \lowerperf{99.10} & \betterperf{98.70} & 99.72 & 99.67 & 95.86~{\tiny \lowerperf{(-0.28\,\%p)}} \\
        \midrule
        \multicolumn{14}{l}{\textit{$\tau_v$}} \\
        0.025 & \lowerperf{99.76} & \lowerperf{93.60} & \lowerperf{73.33} & \betterperf{95.56} & \betterperf{99.95} & \lowerperf{93.10} & 99.82 & 100.00 & 99.15 & \betterperf{97.83} & \betterperf{99.86} & \lowerperf{99.34} & 95.94~{\tiny \lowerperf{(-0.20\,\%p)}} \\
        \rowcolor{gray!10} 0.050 & 99.84 & 95.58 & 74.18 & 94.99 & 99.90 & 93.45 & 99.82 & 100.00 & 99.15 & 97.39 & 99.72 & 99.67 & 96.14 \\
        0.100 & \lowerperf{99.76} & \lowerperf{93.44} & \lowerperf{72.74} & \betterperf{95.56} & 99.90 & \lowerperf{92.41} & 99.82 & 100.00 & \betterperf{99.20} & \betterperf{98.26} & \betterperf{99.86} & \lowerperf{99.00} & 95.83~{\tiny \lowerperf{(-0.31\,\%p)}} \\
        0.150 & \lowerperf{99.61} & \lowerperf{94.74} & \lowerperf{72.54} & \betterperf{95.61} & 99.90 & \lowerperf{92.41} & 99.82 & 100.00 & \betterperf{99.20} & 97.39 & \lowerperf{99.58} & 99.67 & 95.87~{\tiny \lowerperf{(-0.27\,\%p)}} \\
        \midrule
        \multicolumn{14}{l}{\textit{$N_r$}} \\
        500 & 99.84 & \lowerperf{94.43} & \lowerperf{72.90} & \betterperf{95.71} & \lowerperf{99.80} & \lowerperf{92.99} & 99.82 & 100.00 & \betterperf{99.25} & \betterperf{98.70} & 99.72 & \lowerperf{99.34} & 96.04~{\tiny \lowerperf{(-0.10\,\%p)}} \\
        1,000 & \lowerperf{99.61} & \lowerperf{94.60} & \lowerperf{72.96} & \betterperf{95.42} & 99.90 & \lowerperf{92.76} & 99.82 & 100.00 & \betterperf{99.25} & \betterperf{98.70} & \betterperf{99.86} & \lowerperf{99.34} & 96.02~{\tiny \lowerperf{(-0.12\,\%p)}} \\
        1,500 & \lowerperf{99.21} & \lowerperf{94.77} & \lowerperf{73.24} & \betterperf{95.08} & \betterperf{99.95} & \lowerperf{92.87} & 99.82 & 100.00 & \betterperf{99.30} & \betterperf{98.70} & \betterperf{99.86} & \lowerperf{99.34} & 96.01~{\tiny \lowerperf{(-0.13\,\%p)}} \\
        \rowcolor{gray!10} 2,000 & 99.84 & 95.58 & 74.18 & 94.99 & 99.90 & 93.45 & 99.82 & 100.00 & 99.15 & 97.39 & 99.72 & 99.67 & 96.14 \\
        2,500 & \betterperf{99.92} & \lowerperf{93.95} & \lowerperf{72.62} & OOM & \betterperf{99.95} & \lowerperf{92.87} & 99.82 & 100.00 & \lowerperf{99.10} & 97.39 & 99.72 & \lowerperf{99.34} & OOM \\
        3,000 & \lowerperf{99.61} & OOM & OOM & OOM & OOM & \lowerperf{92.18} & 99.82 & 100.00 & 99.15 & 97.39 & 99.72 & 99.67 & OOM \\
        \bottomrule
    \end{tabular}
    }%
    }
\end{table*}
\endgroup

\begingroup
\begin{table}[t!]
    \centering
    \setlength{\tabcolsep}{2pt}
    \caption{\rev{Outdoor-specific sensitivity to the radius cap $r_\mathrm{max}$. This parameter applies to large-scale LiDAR scans; therefore, only outdoor datasets are reported. Shaded rows indicate default settings. Values are registration recall (\%). Red numbers indicate a decrease from the default parameter setting, while green numbers indicate an increase.}}
    \label{table:sens-rmax}
    {\scriptsize
    \begin{tabular}{l|cccccc|c}
        \toprule
        & \multicolumn{6}{c|}{Outdoor} & \multirow{2}{*}{Avg.} \\
        \cmidrule(lr){2-7}
        $r_\mathrm{max}$ & \KITTI & \WOD & \KAIST & \MIT & \ETH & \Oxford &  \\ \midrule
        2.5 m & 99.82 & 100.00 & \lowerperf{97.94} & \lowerperf{96.52} & \betterperf{99.86} & \lowerperf{99.00} & 98.86~{\tiny \lowerperf{(-0.43\,\%p)}} \\
        \rowcolor{gray!10} 5.0 m & 99.82 & 100.00 & 99.15 & 97.39 & 99.72 & 99.67 & 99.29 \\
        7.5 m & 99.82 & 100.00 & \betterperf{99.35} & \betterperf{99.13} & \betterperf{99.86} & 99.67 & 99.64~{\tiny \betterperf{(+0.35\,\%p)}} \\
        10.0 m & 99.82 & 100.00 & \betterperf{99.35} & \betterperf{97.83} & \betterperf{100.00} & 99.67 & 99.44~{\tiny \betterperf{(+0.15\,\%p)}} \\
        \bottomrule
    \end{tabular}
    }
\end{table}
\endgroup

\vfill

\end{document}